\documentclass[Afour,sageh,times,pdftex]{sagej}

\usepackage{pdflscape}
\graphicspath{{./images/}}
\usepackage{amsfonts}
\usepackage{booktabs}
\usepackage{multirow}
\usepackage{amsmath}
\usepackage[colorlinks,bookmarksopen,bookmarksnumbered,citecolor=red,urlcolor=red]{hyperref}
\usepackage[export]{adjustbox}
\usepackage[normalem]{ulem}
\usepackage{tabularx}
\usepackage{array}
\usepackage{subcaption}
\usepackage{caption}
\newcolumntype{M}{>{\arraybackslash}m{0.95\columnwidth}}
\newcolumntype{L}{>{\arraybackslash}m{10em}}
\newcolumntype{K}{>{\centering\arraybackslash}m{3em}}
\newcolumntype{J}{>{\arraybackslash}m{4.2em}}
\newcolumntype{I}{>{\centering\arraybackslash}m{3em}}
\newcolumntype{H}{>{\arraybackslash}m{17.5em}}
\newcolumntype{G}{>{\centering\arraybackslash}m{3.9em}}
\newcolumntype{F}{>{\arraybackslash}m{4em}}
\newcolumntype{E}{>{\arraybackslash}m{1.2em}}
\newcolumntype{C}{>{\centering\arraybackslash}m{5em}}
\newcolumntype{D}{>{\centering\arraybackslash}m{4.5em}}
\newcolumntype{B}{>{\centering\arraybackslash}m{4em}}
\newcolumntype{A}{>{\centering\arraybackslash}m{4.4em}}
\newcolumntype{O}{>{\arraybackslash}m{1.8em}}

\newcommand\BibTeX{{\rmfamily B\kern-.05em \textsc{i\kern-.025em b}\kern-.08em
T\kern-.1667em\lower.7ex\hbox{E}\kern-.125emX}}

\newcommand{\xt}[1]{\textcolor{black}{#1}}

\usepackage[colorinlistoftodos]{todonotes}

\def\volumeyear{2024}

\begin{document} 

\runninghead{Yang, Ji and Lai}
\title{Differentiable Physics-based System Identification for Robotic Manipulation of Elastoplastic Materials}

\author{Xintong Yang\affilnum{1}, Ze Ji\affilnum{1}, Yu-Kun Lai\affilnum{2}}
\affiliation{\affilnum{1}School of Engineering, Cardiff University, Cardiff, UK\\
\affilnum{2}School of Computer Science and Informatics, Cardiff University, Cardiff, UK}
\corrauth{Ze Ji, School of Engineering, Cardiff University, Cardiff, CF24~3AA, UK}

\email{jiz1@cardiff.ac.uk}

\begin{abstract}
  Robotic manipulation of volumetric elastoplastic deformable materials, from foods such as dough to construction materials like clay, is in its infancy, largely due to the difficulty of modelling and perception in a high-dimensional space. Simulating the dynamics of such materials is computationally expensive. It tends to suffer from inaccurately estimated physics parameters of the materials and the environment, impeding high-precision manipulation. Estimating such parameters from raw point clouds captured by optical cameras suffers further from heavy occlusions.
  To address this challenge, this work introduces a novel Differentiable Physics-based System Identification (DPSI) framework that enables a robot arm to infer the physics parameters of elastoplastic materials and the environment using simple manipulation motions and incomplete 3D point clouds, aligning the simulation with the real world. 
  Extensive experiments show that with only a single real-world interaction, the estimated parameters, Young’s modulus, Poisson’s ratio, yield stress and friction coefficients, can accurately simulate visually and physically realistic deformation behaviours induced by unseen and long-horizon manipulation motions. Additionally, the DPSI framework inherently provides physically intuitive interpretations for the parameters in contrast to black-box approaches such as deep neural networks. 
  The project is fully open-sourced via \href{https://github.com/IanYangChina/SI4RP-data}{GitHub}.
\end{abstract}


\maketitle 

\section{Introduction}
\begin{figure*}[t]
    \centering
    \includegraphics[width=\linewidth]{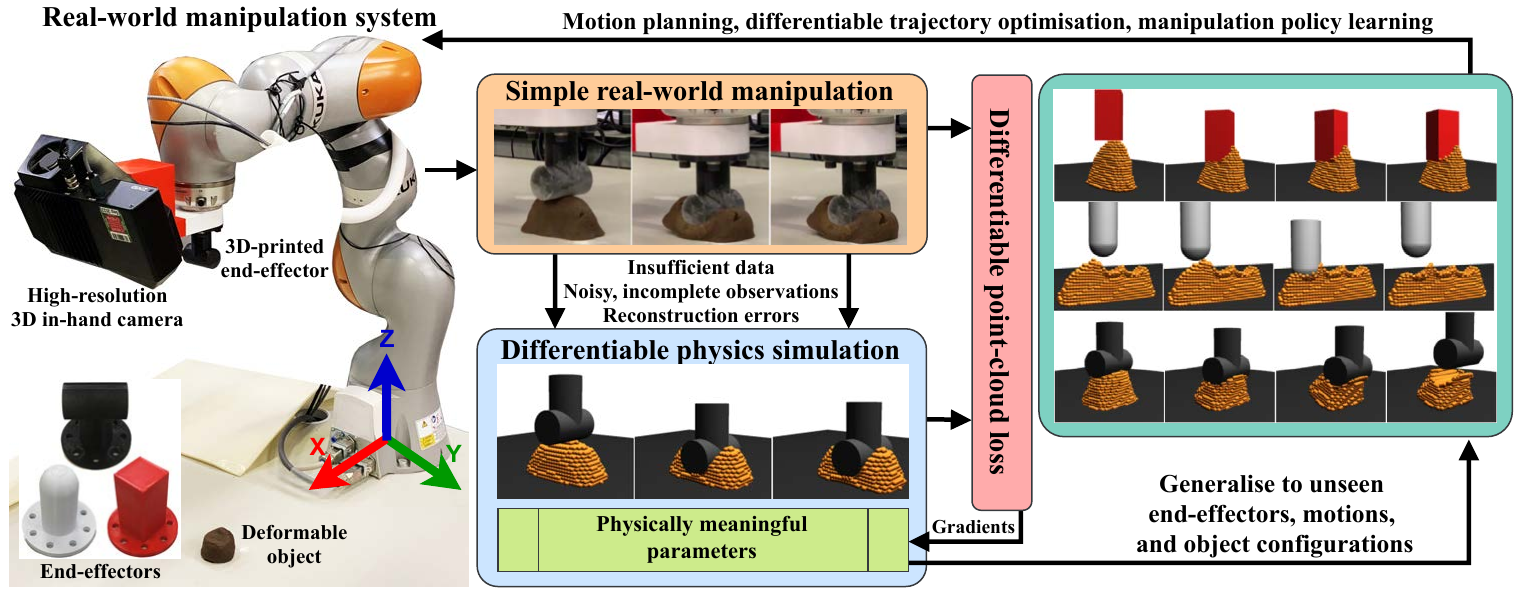}
    \caption{The proposed system identification framework enables a robot to interact with elastoplastic material via simple manipulation motions (orange box) and then identify the physics parameters of the real-world manipulation dynamics. The parameters are found using gradients computed, through differentiable simulation, from a differentiable point-cloud-based similarity function between the real and simulated observations of the manipulated material (blue and red boxes). These parameters then enable accurate simulations that allow the real-world grounding of motion planning, trajectory optimisation or policy learning techniques (cyan box).}
    \label{fig:SI-problem}
\end{figure*}

Despite the recognised importance of robotic manipulation of deformable materials, this topic remains underexplored, particularly when it comes to high-precision manipulation of volumetric elastoplastic materials. A primary challenge in this area arises from the materials' infinite degrees of freedom (DoFs), leading to highly unpredictable deformation dynamics. 

The intrinsic complexity of these dynamics inhibits the direct application of conventional robotic motion planning methods, which typically require explicit physics models for all concerned objects~\citep{latombe2012robot}. 
Learning approaches, such as reinforcement learning (RL), often involve training an agent to learn to interpret its perception and take actions through inefficient trial \& error in a realistic physics-based simulation~\citep{collins2021review,kroemer2021review}, which is both challenging and largely unavailable when it comes to deformable materials. 

In contrast to the well-studied rigid body dynamics in robotics, where motions can be predicted and controlled using well-defined equations of motion and deterministic models~\citep{featherstone2014rigid}, deformable materials do not follow such straightforward patterns. Directly applying these methods is problematic because it is extremely difficult and often infeasible to accurately model and perceive real-world elastoplastic materials and measure the underlying physics parameters that govern their motions and deformations~\citep{arriola2020modeling,yin2021modeling}. \xt{Additional challenges such as time-varying material properties (saturation, dryness, etc.) make the deformation dynamics even more unpredictable.} As a result, achieving high-precision manipulation for such materials with motion planning or data-driven techniques is challenging due to the high computational cost and the lack of techniques to capture the dynamics accurately \xt{and efficiently}. 

To close this gap, this research proposes a Differentiable Physics-based System Identification (DPSI) framework for the robotic manipulation of volumetric elastoplastic materials. Our framework can efficiently estimate key physics parameters governing material deformation dynamics using minimal and simple manipulation motions \xt{(in less than $5$ minutes with a good initial guess)}. The estimated physics parameters enable accurate material simulation for long-horizon predictions of real-world elastoplastic deformation behaviours. 

The workflow of the proposed DPSI framework can be summarised as follows. As shown in Figure~\ref{fig:SI-problem}, a robot equipped with an in-hand 3D camera (Zivid) and three end-effectors is deployed to manipulate the elastoplastic object (e.g., play dough). Before manipulation, the robot takes multi-view point clouds of the object to minimise occlusions. These point clouds are used to create the initial particle system for the material point methods (MPM)-based simulation. The robot then performs a manipulation motion on the object and captures point clouds of the deformed state. Physics simulations of the same manipulation are run with the same initial state and motion, whose resultant particle states are compared to the real-world deformed state through variants of the Chamfer distance (CD) and the earth mover's distance (EMD) loss. The parameters of the simulated physics models are updated after every simulation to minimise the loss. To facilitate fast optimisation, we use a differentiable simulator built on the TaiChi auto-differentiation mechanism, which allows automatic gradient computation from the losses and gradient-based optimisation for the physics parameters~\citep{hu2019taichi,hu2019difftaichi}.

Our approach achieves unprecedented simulation-to-real alignment accuracy, characterised by the integration of the following novel features.

\textbf{\textit{High-fidelity physics}:} Unlike previous methods that either employ non-physics-based models (e.g., neural networks) or highly simplified material geometry representations (e.g., sparse keypoints), we use high-fidelity physics-based simulation powered by the MPM~\citep{jiang2016material}, which simulates materials as Lagrangian particles and keeps track of their positions and velocities. It achieves faster simulation by computing the motions, deformation gradients, and frictional contacts on a background Eulerian grid~\citep{stomakhin2013material,jiang2016material,gao2017adaptive,hu2018moving}. MPM-based simulations provide highly efficient and realistic simulation with high physical plausibility by closely following real-world physics laws such as Newton's laws and elastic and plastic energy conservation models.

\textbf{\textit{Incomplete \& noisy observations}:} Unlike existing works that rely on synthetic videos with complete sequences of perfect observations~\citep{murthy2020gradsim,li2023pacnerf,kaneko2024improving}, our framework uses 3D point clouds to observe real-world object geometries. Capturing the full depth of an object during manipulation is impractical due to occlusions caused by the end-effector or environment. This means that only the point clouds before and after a manipulation motion are practical to obtain and informative enough to observe the full geometry of the deformed object in real-world experiments. In addition, real-world point clouds tend to suffer from inaccurately estimated camera matrices and sensory noises.

\textbf{\textit{Small data, short \& simple motions}:} Using extensive and diverse manipulation motions to collect real-world deformation data is time-consuming and costly. Existing studies demand a significant number of real-world interactions or complete sequences of simulation videos to identify object deformations under various motions, yet still resulting in simulations with insufficient accuracy for real-world applications~\citep{lin2022diffskill,shi2023robocook,li2023pacnerf,kaneko2024improving}. Our goal is to recover physics parameters that enable accurate predictions of long-horizon, unseen and complicated elastoplastic material manipulation dynamics, using minimal simple and short real-world interactions.

\textbf{\textit{Joint parameter estimation}:} We aim to jointly estimate the physics parameters provided by physics models. Besides Newton's laws, we employ the fixed corotated elastic energy model \citep{stomakhin2012energetically}, the von Mises plasticity model \citep{jones2009deformation} and the dynamic friction model in our simulation. These lead to six key parameters: Young's modulus $E$, Poisson's ratio $\nu$, yield stress $\pmb{\sigma}_y$, material density $\rho$, and the friction coefficients of the table $\eta_t$ and end-effectors $\eta_m$ (assuming the three end-effectors share the same coefficient). The first four parameters primarily govern the deformation responses, while the last two handle frictional contacts. These parameters are heavily intertwined in governing the behaviours of the manipulated object and there is no intuitive solution to identify one of them without estimating the effects of other parameters. Therefore, we seek to identify these parameters simultaneously.

\textbf{\textit{Differentiable physics}:} Identifying the physics parameters in their discretised spaces via search or evolutionary algorithms is computationally slow due to the exponentially growing number of possible combinations as the discretisation becomes finer. While gradient-based optimisation methods offer faster convergence toward the minimum, it is infeasible with most physics simulations because many computation steps are not differentiable and these simulators do not support derivative computations. In this work, we explore the feasibility of optimising system parameters using gradients computed by differentiating loss functions through a physics simulator written by DiffTaiChi, a programming language tailored for GPU-accelerated parallel computation and automatic differentiation~\citep{hu2019taichi,hu2019difftaichi}. DiffTaichi generates derivative functions for simulation steps via source code transformation that retains arithmetic intensity and parallelism. It uses a memory-efficient tape to record the order of computation kernels for forward simulation and traverses their derivative functions in the backward order to generate gradients through the computation graph. We build DPSI upon DiffTaiChi and explore the feasibility of directly optimising several physics parameters jointly with gradients computed by differentiating point-cloud-based loss functions through the high-fidelity physics simulator.

Substantial experiments demonstrate that our main contribution, DPSI, can achieve highly accurate simulation-to-reality alignment for elastoplastic materials manipulated by unseen, long horizon and complex motions using minimal simple and short interactions, and noisy and incomplete observations. Results show that when multiple solutions and parameter uncertainty exist, DPSI can provide physically intuitive parameter interpretations that can guide further system identification, model improvement, and motion adaptation. \xt{Statistics on the computation costs also indicate promising practical deployment of the DPSI framework.}

The rest of the article reviews related literature, presents formally our method and experiment results, and discusses limitations and future directions.

\section{Related work}
\subsection{Deformable object manipulation}
Both model-free and model-based approaches have been taken for manipulating deformable materials. Existing model-free methods often lack manipulation precision due to the absence of physics laws that describe the motion and deformation under complex contacts~\citep{mcconachie2020manipulating,cherubini2020model,shi2023robocook,shi2024robocraft,shen2024action}. Physics-model-based methods, while more accurate, struggle with aligning the simulated dynamics with the real world and often rely on simplified geometric representations for higher computational efficiency, sacrificing manipulation precision~\citep{navarro2016automatic,yang2023model,shetab2023lattice}. With or without a model, many of them use simulations that offer a cost-effective way to test manipulation methods and allow the collection of massive data for learning-based approaches \citep{arriola2020modeling,collins2021review,yin2021modeling}. However, inaccurate dynamics predictions and control precision are unacceptable for many tasks such as surgery, assembly and disassembly. Therefore, akin to the way humans efficiently learn about object and environment physics properties, this work proposes to actively identify system parameters that enable high-precision simulation of real-world manipulation dynamics for volumetric elastoplastic objects.

\subsection{Deformable object modelling}
Simulators are essential for advancing robotic manipulation, providing a fast, low-cost alternative to real-world testing~\citep{featherstone2014rigid,collins2021review}. In recent years, arguably the most efficient and accurate simulation for 3D deformable objects is achieved by the material point method (MPM)~\citep{jiang2016material}. Like most well-known methods for simulating continuum matter, such as position-based dynamics (PBD) and smoothed particle hydrodynamics (SPH)~\citep{yin2021modeling}, MPM represents the object as Lagrangian particles and keeps track of their positions and velocities. Unlike pure Lagrangian methods, MPM achieves faster simulation by computing the motions, deformation gradients, and frictional contacts on a background Eulerian grid (akin to the finite element method), governed by elastic and plastic energy functions and Newton's laws in the form of partial differential equations~\citep{stomakhin2013material,jiang2016material,gao2017adaptive,hu2018moving}. MPM has been proven superior to other methods in terms of efficiency and visual effects for objects that undergo large deformations, fractures, and self-collision~\citep{hu2018moving}. Despite its advantages, accurately identifying physics parameters for real-world robotic manipulation of elastoplastic objects remains an open challenge.

\subsection{Deformable object system identification}
Unlike rigid bodies~\citep{miguel2022sysid,heiden2022probabilistic,chen2022real}, identifying the physics properties of deformable objects is more complex than for rigid bodies, primarily due to the difficulty in measuring key parameters like material properties and friction coefficients~\citep{arriola2020modeling,yin2021modeling}. Previous research has mostly focused on linear and planar deformable objects, such as ropes and cloths~\citep{sundaresan2022diffcloud,yang2022learning,caporali2024deformable}. 

Early works that sought to identify 3D volumetric material properties have much simpler assumptions and scenarios. For instance, one of the earlier works uses gradient-based optimisation to retrieve the stiffness of a spring system representing elastic deformable objects~\citep{lloyd2007identification}, while another work uses an exhaustive search method to find the value of Poisson's ratio for an elastic form object~\citep{guler2017estimating}. They focus on single parameter identification for volumetric elastic deformable objects with reduced Dofs that under-represent the geometries and deformation behaviours of real-life objects.

Recent efforts like GradSim~\citep{murthy2020gradsim} have used differentiable physics and rendering to identify five parameters for elastic objects with much higher DoFs from single-view simulation videos, demonstrating the feasibility of differentiable system identification with synthetic videos. Two following works, PAC-NeRF~\citep{li2023pacnerf} and LPO~\citep{kaneko2024improving} propose to jointly reconstruct object geometries (shapes, positions and colours) and physical properties by using a voxel neural radiance field~\citep{sun2022direct} that performs differentiable rendering and allows gradients to be back-propagated from the image space to the Eulerian grid. 

Compared to these works that focused on simulations, our study tackles the more challenging task of system identification for real-world objects through robot interactions. Similarly, we also employ the material point method~\citep{hu2018moving} and DiffTaiChi~\citep{hu2019difftaichi} for differentiable physics simulation and study the system identification task without known object geometries. However, we aim to align simulations with real-world dynamics with minimal robot interactions, using only incomplete, occluded, and noisy point cloud data.

\section{Method}
\label{sec:materials_methods}

\subsection{Overview}

\begin{figure*}[t]
    \centering
    \includegraphics[width=0.8\linewidth]{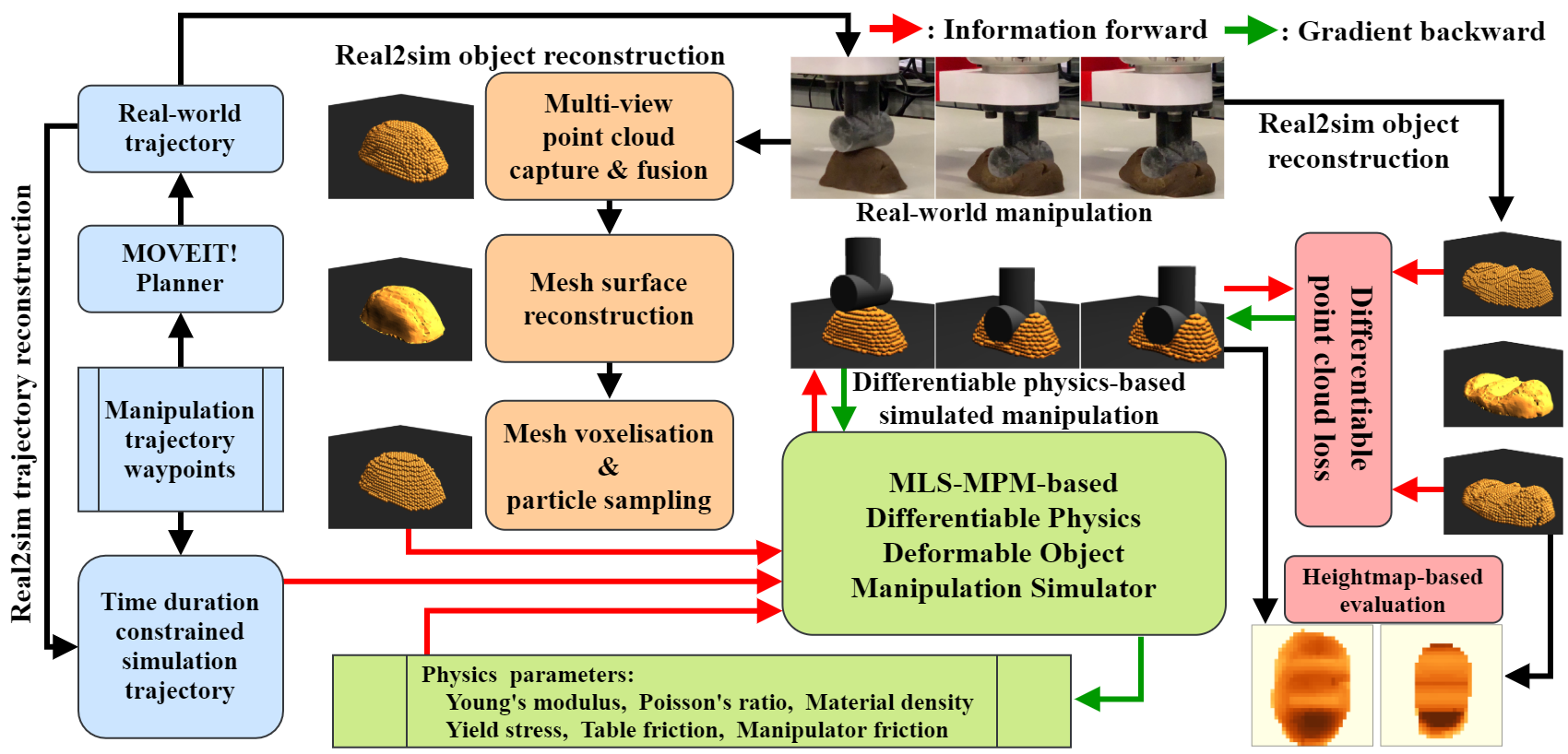}
    \caption{The overall workflow of the proposed differential physics-based system identification (DPSI) framework. Modules in each colour are elaborated in individual subsections of the \textbf{Method} section. Green: differentiable dynamics modelling and the physics parameters. Orange: real-to-sim object reconstruction. Blue: real-to-sim trajectory reconstruction. Pink: optimisation and evaluation loss functions.}
    \label{fig:overview}
\end{figure*}

The proposed differentiable physics-based system identification (DPSI) framework, as shown in Figure~\ref{fig:overview}, can be divided into the following modules: differentiable dynamics modelling, real-to-sim object and trajectory reconstruction, and the loss functions. This section starts with a formal problem description and elaborates on each of these modules.

\subsection{Problem statement}
We define the physics parameter identification problem for real-world deformable materials as follows. Denote $\mathcal{X}^{real}_0$ and $\mathcal{X}^{real}_T$ as the observations of a deformable object before and after a manipulation trajectory $\tau^{real}$ of time duration $T$ is applied, denote $\mathcal{S}^{real}_0$ as the observation of the real-world end-effector before the manipulation, the stochastic real-world forward dynamics can be written as follows:

$$\mathcal{X}^{real}_T \sim p^{real}(\mathcal{X}^{real}_T | \mathcal{X}^{real}_0, \pmb{\theta}^{real}, \tau^{real}, \mathcal{S}^{real}_0)$$

\noindent where $\pmb{\theta}^{real}$ is the set of real-world parameters that govern the behaviours of the dynamical process. As it is unlikely to obtain $p^{real}$ and $\pmb{\theta}^{real}$ explicitly, we use an approximate, deterministic and differentiable dynamics model, $f^{sim}$, to simulate such a dynamical process, written as follows:

$$\mathcal{X}^{sim}_T=f^{sim}(\mathcal{X}^{rec}_0, \pmb{\theta}^{sim}, \tau^{rec}, \mathcal{S}^{rec}_0)$$

\noindent where $\pmb{\theta}^{sim}$ is the set of adjustable parameters that govern the behaviours of the simulated deformation and contact process, $\mathcal{X}^{rec}_0$ is the reconstructed initial observation of the real object, $\mathcal{X}^{sim}_T$ is the resultant observation of the reconstructed object after manipulated by the reconstructed motion $\tau^{rec}$, and $\mathcal{S}^{rec}_0$ is the initial observation of the simulated end-effector. In our work, we assume the end-effectors are rigid bodies, meaning the observations of the real or simulated end-effectors are coordinates of their frames relative to the world frame, i.e., $\mathcal{S}\in\mathbb{R}^3$, which can be readily retrieved from the robot platform or simulator. As we are interested in point cloud observations from the real world and particles in the simulation, all object observations will be 3D point sets, i.e., $\mathcal{X}\in\mathbb{R}^{N\times3}$ where $N$ is the number of points or particles.

With these notations, the optimisation problem could be formulated as the minimisation of some distance function between the real-world and simulated observations after manipulation. As the parameters are optimised with a dataset of interaction experiences with the real objects, $\mathcal{D}$, the minimisation problem can be written as follows: 

\begin{equation}\label{eq:minimisation}
\min_{\pmb{\theta}^{sim}}\ \frac{1}{|\mathcal{D}|}\sum_{\mathcal{D}} d(\mathcal{X}^{real}_T, \mathcal{X}^{sim}_T)
\end{equation}

The rest of this section will discuss the differentiable simulation dynamics $f^{sim}$, the reconstructed object state, $\mathcal{X}^{rec}$, the reconstructed end-effector trajectory, $\tau^{rec}$, and the loss functions, $d(\mathcal{X}^{real}_T, \mathcal{X}^{sim}_T)$.

\subsection{Differentiable dynamics modelling}
\noindent\textbf{\textit{Material point method (MPM)}:} This work employs the mean-least-square material point method (MLS-MPM) to simulate the deformable object manipulation dynamics. 
MPM is a meshless, hybrid computation scheme that enables efficient computation and preserves high physical fidelity for various materials, especially for elastoplastic materials that undergo large deformation, while the MLS-MPM arises from a novel weak form discretisation of the conservation equations and replaces the shape functions in the force computation with MLS approximators, leading to faster and more realistic simulation of sharp separation of particles and two-way coupling with rigid objects that traditional MPM cannot simulate.

We modified the standard procedures in one simulation step of the MLS-MPM as shown below to incorporate end-effector control following~\cite{xian2023fluidlab}. The readers are referred to the original paper for more details about MPM~\citep{stomakhin2013material} and MLS-MPM~\citep{hu2018moving}.
\begin{enumerate}
    \item Compute particle deformation gradient using the MLS approximation equation.
    \item Applied plasticity to recompute deformation gradients and particle stress using the elastic energy model
    \item Particle to grid. Use the affine particle-in-cell transform~\citep{jiang2015affine} to transfer the velocities and masses of the Lagrangian particles to the background Eulerian grid nodes. In our implementation, we assume equal volume and mass for the particles.
    \item Update end-effector positions given control inputs.
    \item Compute grid node momenta and velocities with gravity applied.
    \item Signed-distance field (SDF)-based collision detection with rigid objects (end-effector and boundaries) and frictional contact computation.
    \item Grid to particle. Use the affine particle-in-cell transform to transfer the velocities and affine coefficients from the grid nodes to the particles. Perform SDF-based collision detection and frictional contact computation again to minimise particle penetration.
    \item Update particle positions with the new velocities.
\end{enumerate}

\noindent\textbf{\textit{Elastoplasticity}:} We assume strain elastoplasticity for the studied objects, meaning that the strain-stress relationship of the material is described using the deformation gradient that can be decomposed into elastic and plastic parts
, $\mathbf{F}=\mathbf{F}^E\mathbf{F}^P$, and the first Piola-Kirchoff stress, $\mathbf{P}$, as $\mathbf{P}=\partial\Psi/ \partial\mathbf{F}^E$, where $\Psi(\mathbf{F}^E)$ is the elastic energy density function. In this study, we use the fixed corotated model \citep{stomakhin2012energetically} as the energy function for its robustness under large deformation, although other constitutive models may be used for different needs. The fixed corotated elastic energy density function is

\begin{equation}\label{eq:constitutive}
\Psi(\mathbf{F}^E)=\mu||\mathbf{F}^E-\mathbf{R}^E||^2_F+\frac{\lambda}{2}(J^E-1)^2
\end{equation}

\noindent where $J^E=\text{det}(\mathbf{F}^E)$, and $\lambda$ and $\mu$ are the Lamé parameters, such that

\begin{equation}\label{eq:lame1}
\lambda = \frac{E}{2(1+\nu)}    
\end{equation}

\begin{equation}\label{eq:lame2}
\mu = \frac{E\nu}{(1+\nu)(1-2\nu)}
\end{equation}

\noindent where $E$ is Young's modulus and $\nu$ is Poisson's ratio. Thus, the stress can be calculated as

\begin{equation}\label{eq:pk-stress}
\mathbf{P} = \frac{\partial\Psi}{\partial\mathbf{F}^E} = 2\mu(\mathbf{F}^E-\mathbf{R}^E)(\mathbf{F}^E)^T + \lambda(J^E-1)J^E
\end{equation}

\noindent where $\mathbf{R}^E$ is the rotation matrix that comes from the polar decomposition of the deformation gradient, $\mathbf{F}^E=\mathbf{R}^E\mathbf{S}^E$, where $\mathbf{S}^E$ is a symmetric matrix. In practice, they are computed from the singular value decomposition of $\mathbf{F}^E$ as $\mathbf{F}^E=\mathbf{U}^E\mathbf{\Sigma}^E(\mathbf{V}^E)^T=\mathbf{U}^E(\mathbf{V}^E)^T\mathbf{V}^E\mathbf{\Sigma}^E(\mathbf{V}^E)^T$, and then $\mathbf{R}^E=\mathbf{U}^E(\mathbf{V}^E)^T$ and $\mathbf{S}^E=\mathbf{V}^E\mathbf{\Sigma}^E(\mathbf{V}^E)^T$ \citep{baker2005singular}. 

For the $n$-th simulation step, we compute a trial elastic deformation gradient, $\tilde{\mathbf{F}^{E,n}}$. If no plastic deformation appears, meaning $\mathbf{F}^{P,n}=\mathbf{F}^{P,n-1}$, then we assign $\mathbf{F}^{E,n}=\tilde{\mathbf{F}}^{E,n}$. However, if the yield criterion is violated, $\tilde{\mathbf{F}}^{E,n}$ needs to be modified according to the plasticity model. This process of modifying $\tilde{\mathbf{F}}^{E,n}$ is called the return mapping process, which determines how the object would respond plastically. 

In this work, we follow \cite{gao2017adaptive} to use the von Mises model \citep{jones2009deformation} to compute the return mapping, which takes on the associative plastic flow assumption. The projection process of the trial stress outside of the yield region can be described succinctly as $\tilde{\pmb{\epsilon}}^{E,n}-\pmb{\epsilon}^{E,n}=\delta_{\gamma}\tilde{\pmb{s}}^{E,n}/||\tilde{\pmb{s}}^{E,n}||$, where $\tilde{\pmb{\epsilon}}^{E,n}=\log(\tilde{\mathbf{F}}^{E,n})$ and $\pmb{\epsilon}^{E,n}=\log(\mathbf{F}^{E,n})$ are the trial and modified Hencky strains, $\tilde{\pmb{s}}^{E,n}=\tilde{\pmb{\epsilon}}^{E,n}-\ tr(\tilde{\pmb{\epsilon}}^{E,n})/3\ \pmb{I}$ is the trial deviatoric stress, and $\delta_{\gamma}\tilde{\pmb{s}}^{E,n}/||\tilde{\pmb{s}}^{E,n}||$ is the modifying term based on the solution given in \cite{gao2017adaptive}. In particular, based on the von Mises yield criterion, $\sqrt{3J_2}-\pmb{\sigma}_y\leq 0$, where $\pmb{\sigma}_y$ is the yield stress and $J_2$ is the second deviatoric stress invariant, it was derived that $\delta_{\gamma}=||\tilde{\pmb{s}}^{E,n}||-\pmb{\sigma}_y/2\mu$. With the fact that the singular vectors of the trial elastic deformation gradient do not change, the return mapping operation is done on the eigenvalues of the trial Hencky stress, thus on the eigenvalues of the trial elastic deformation gradient. Then, we can obtain the result of the return mapping by taking the exponential of the modified deviatoric stress: $\Sigma^{E,n}=e^{\pmb{\epsilon}^{E,n}}$ and then the modified deformation gradient: $\mathbf{F}^{E,n}=\tilde{\mathbf{U}}^{E,n}\mathbf{\Sigma}^E(\tilde{\mathbf{V}}^{E,n})^T$.

Note that the Hencky strain formation is only used in plastic response computation. At each simulation step, we compute the trial deformation gradient using the MLS approximation equation, apply plastic response, and then compute the new stress with Equation~\ref{eq:pk-stress}. 


\noindent\textbf{\textit{Frictional contacts}:} Collision detection is done by checking the distance of the particles to the surface of the rigid objects using pre-computed signed distance fields (SDFs). For each rigid object (table, and three end-effectors in our case), we employ the procedure described in \cite{park2019deepsdf} to generate SDFs for watertight meshes. We assume that the frictional contacts only happen in two cases: between the particles and the table, or between the particles and the end-effector. We also assume uniform friction coefficient distribution over the contact surface of the table and the end-effectors. When a particle or grid node and an object are in contact, we follow \cite{stomakhin2013material} to determine the velocity of the particle and grid nodes using dynamic friction with sticky impulse. 

Specifically, for each particle, we calculate the local normal $\overrightarrow{n}$ and the velocity $v_{obj}$ of the rigid object at the particle position. Then, we project the particle velocity $v$ into the object reference frame to have $v_{rel}=v-v_{obj}$. No collision is applied if the particle is separating from the object, i.e., $v_n=v_{rel}\cdot \overrightarrow{n}\geq0$. Define the tangential portion of the relative particle velocity as $v_t = v_{rel}-\overrightarrow{n}v_n$. For sticky impulse, we set the relative particle velocity $v_{rel}'$ after the collision to $0$ if $||v_t||\leq -\mu v_n$, where $\mu$ is the friction coefficient. If the sticky impulse is overcome, dynamics friction is applied, such that $v_{rel}' = v_t (1 + \mu v_n/||v_t||)$. After the collision, the new particle velocity in the world frame is calculated as $v'=v_{rel}'+v_{obj}$. We represent the friction coefficients for the table and the manipulators separately as $\mu_t$ and $\mu_m$.

\noindent\textbf{\textit{Differentiable programming}:} Several programming tools are available for creating differentiable simulations, such as PyTorch \citep{sundaresan2022diffcloud,arnavaz2023differentiable}, DiffTaiChi \citep{hu2019difftaichi}, Jax \citep{schoenholz2020jax} and neural-network-based simulators \citep{heiden2021neuralsim}. We build our simulator based on DiffTaiChi due to its automated differentiation mechanism, GPU-accelerated parallel computation, fast computation kernel evaluation, intuitive Python APIs, rich community support and various promising applications \citep{huang2021plasticinelab,lin2022diffskill,xian2023fluidlab}. 

In particular, for each numerical computation step, DiffTaiChi flattens its computation branches (e.g., boundary and collisions) and replaces mutable local variables with extra local storage variables, producing straight-line codes without mutable variables. It then uses standard source code transform to generate the derivative function based on the adjoint method. To compute gradients with a loss function, DiffTaiChi records the order of computation kernels and the scalar variables in the forward simulation direction, and then it computes the gradients for the concerned variables by evaluating the derivative functions in the reversed simulation direction~\citep{hu2019difftaichi}. We build a simulator based on this programming language to allow automatic gradient computation for the physics parameters.

In summary, the elastic and plastic models describe the deformation behaviours of the object using Young's modulus $E$, Poisson's ratio $\nu$ and yield stress $\sigma_y$, the computation of the particle and grid velocities will involve another parameter, the object density $\rho$, and the frictional contact processes are controlled by respective friction coefficients: table friction coefficient $\mu_t$ and manipulator friction coefficient $\mu_m$. We optimise these six parameters to align the simulation to the real-world dynamics.

\subsection{Real-to-sim object reconstruction}
\noindent\textbf{\textit{Real-world platform}:} To collect real-world data, we set up a deformable object manipulation system where a Kuka IIWA LBR 14 industrial arm \citep{kukalbr} is equipped with a Zivid One+ medium camera \citep{zividone} for perceiving the object, and one of the three end-effectors, namely a rectangular cuboid, a cylinder roller and a bullet-shaped object for collecting interaction experiences with different contact geometries, as shown at the bottom-left of Figure~\ref{fig:SI-problem}. \xt{We use plasticine (non-hardening modelling clay) as the main material for our experiments and cloud slime and soil for generalisation experiments. The clay and cloud slime are purchased from Amazon\footnote{\href{https://amzn.eu/d/fK3HIGa}{Modelling Clay}. \href{https://amzn.eu/d/9BYM1kx}{Cloud Slime Kit}}.}

\noindent\textbf{\textit{Real-world perception}:} We create a multi-view point-cloud capture and fusion process to obtain the real-world observation of the deformable object $\mathcal{X}^{real}_0$ and $\mathcal{X}^{real}_T$. In particular, for each object state, the robot arm moves to six poses around the object for the camera to take a point cloud of the scene. The point clouds are fused and then cropped to contain only the points of the object.

These observations are noisy and incomplete in three senses. Firstly, due to camera calibration error, the fusion result always exhibits a $\sim\pm 3$ mm discrepancy. Secondly, due to joint limits, the robot arm cannot reach poses that allow the camera to capture the bottom part of the object. To simplify the problem, we assume that the angle between the object boundary and the table surface is equal to or greater than $90^{\circ}$, which allows the camera to capture as much as possible the bottom part of the object. We then project all points to its bottom to form a closed surface. Note that the initial configuration of the object can be manually shaped to satisfy this assumption of contact angle but the end configuration of the object after being manipulated is out of manual control, which tends to have more occlusions. Lastly, we only take the observations before and after applying the manipulation motions, without providing the intermediate observations during the manipulation, because it is impractical to do so when too much occlusion occurs during manipulation.

\noindent\textbf{\textit{Reconstruction pipeline}:} To simulate the object as a set of particles, we design a pipeline to reconstruct the object particle system $\mathcal{X}^{rec}$ from the given fusion point cloud, as shown by the orange boxes in Figure~\ref{fig:overview}. Firstly, we down-sample the processed fusion point cloud and create a watertight mesh using the ball-pivoting (BP) algorithm \citep{bernardini1999ball}. The artificial bottom surface points prevent the BP algorithm from creating a non-flat bottom mesh surface. To sample particles, we create a set of voxels that fill in the reconstructed mesh surface uniformly given a resolution ($1080\times1080$). Lastly, we fill particles into the voxels with a given density ($4\times10^6$ particles$/m^3$). We assume that the particles are uniformly distributed within the mesh. 

For the end-effectors, we assume they are rigid bodies. In simulation, we keep track of the coordinate of its frame and a pre-computed SDF for collision detection.
\begin{table*}[t]
\footnotesize
\centering
\begin{tabular}{c|l|H||A|A|A}
\toprule
\multicolumn{3}{l||}{\textbf{Optimisation motions (short and simple)}} & Duration (sec.) & Real Num. waypoints & Sim. Num. waypoints\\
\midrule
\multirow{2}{*}{Lv. 1} & Poking-1 & $-z$ $0.015$ m, $+z$ $0.03$ m & 0.86 & 38 & 87\\
      & Poking-2 & $-z$ $0.02$ m, $+z$ $0.03$ m & 0.93 & 40 & 94\\
\midrule
\multirow{2}{*}{Lv. 2} & Poking-shifting-1 & $-z$ $0.02$ m, $-x$ $0.03$ m, $+z$ $0.03$ m & 1.52 & 68 & 152\\
      & Poking-shifting-2 & $-z$ $0.02$ m, $+x$ $0.03$ m, $+z$ $0.03$ m & 1.50 & 54 & 153\\
\midrule
\multicolumn{3}{l||}{\textbf{Validation motions (unseen, longer, more complex contacts)}} & & & \\
\midrule
\multicolumn{2}{l|}{Flattening (cylinder)} & $-z$ $0.025$ m, $+x$ $0.025$ m, $+z$ $0.025$ m, $-x$ $0.025$ m, $-z$ $0.025$ m, $-x$ $0.025$ m, $+z$ $0.025$ m & 3.74 & 275 & 378\\
\midrule
\multicolumn{2}{l|}{Triple-poking (round)} & $+y$ $0.025$ m, $-z$ $0.025$ m, $+z$ $0.025$ m, $-y$ $0.025$ m, $-z$ $0.025$ m, $+z$ $0.025$ m, $-y$ $0.025$ m, $-z$ $0.025$ m, $+z$ $0.025$ m & 4.55 & 328 & 460\\
\midrule
\multicolumn{2}{l|}{Poking-180-rotating (rectangle)} & $-z$ $0.025$ m, about $+z$ $180^{\circ}$, $+z$ $0.025$ m & 6.23 & 504 & 625\\
\bottomrule
\end{tabular}
\caption{Waypoint designs and statistics of the interaction motions. For collecting optimisation data, there are two motions for each of the two levels of contact complexity. For out-of-distribution validation, one motion per end-effector is designed. Directions of the waypoints are relative to the robot base frame, as shown in Figure~\ref{fig:SI-problem}. Each waypoint of the real trajectories takes an uneven time interval, while each waypoint in simulation takes exactly $0.01$ second.}
\label{tab:motions}
\end{table*}

\subsection{Real-to-sim trajectory reconstruction}
\noindent\textbf{\textit{Manipulation motions}:} As we are interested in recovering real-world dynamics with small data and short, simple motions, we design experiments in two contact complexity levels, for which we create two poking motions and two poking-shifting motions for optimisation and in-distribution validation, and three longer motions for out-of-distribution validation. These motions are summarised in Table~\ref{tab:motions} in terms of consecutive trajectory segments in the world frame. 

The first contact level focuses on identifying the parameters that primarily govern the deformation of the object (Young's modulus, Poisson's ratio, yield stress and material density) using two poking motions, which press the object down by a certain distance. The second level further includes the friction coefficients, using two poking-shifting motions, which press down the object and make horizontal shifting movements.
For out-of-distribution validation, three longer motions with more drastic contact processes are created. Each motion uses a different end-effector. The triple-poking motion is designed for the round end-effector to validate the long-term deformation prediction with small fraction influences. The flattening motion is designed to validate the long-term deformation and frictional contact prediction under large linear movements, using the cylinder end-effector. The pressing-180-rotating motion is designed to validate the long-term deformation and frictional contact prediction under large rotational movements using the rectangle end-effector.

All motions start from a configuration where the end-effector tip is positioned at the top centre of the object. ROS \citep{quigley2009ros} and the MoveIt!-based \citep{gorner2019moveit} OMPL planner \citep{sucan2012open} are used to plan real-world motion trajectories $\tau^{real}$. For calculating each motion plan, we pass a series of waypoints to the MoveIt! planner by discretising each segment of the motion with an interval of $0.002$ m or $5^{\circ}$. 

A challenging phenomenon that occurs during our data collection process is that the object tends to stick to the end-effectors after contact is made. If we allow this to happen and assume that the object always drops down eventually, the optimisation process will be extremely difficult due to a large uncertain dynamics process of the object dropping from the air. For example, the object may bounce out of the workspace or even off the table. This further exacerbates the difficulty of optimising with only the start and end observations of the object. However, it is beyond the capability of current perception hardware and thus a future research direction.
To simplify the problem, we take a simple workaround by covering the end-effector with a thin layer of flour before each manipulation motion is executed. \xt{We sink the effector into the flour and flick it gently to ensure that it is only covered by a thin layer of flour.} This greatly prevents the sticking phenomenon from happening. Note that the end-effector friction coefficient we are optimising for is then the one covered by flour instead of the original value.

\noindent\textbf{\textit{Time duration constrained reconstruction}:} To simulate end-effector motions, the real-world trajectory generated by MoveIt! is inconvenient as it has uneven time differences between consecutive waypoints and the simulation can only handle a constant step size $dt$. Thus, for a motion segment between a pair of waypoints, we reconstruct its simulation counterpart to have constant velocity by dividing the travelled distance by the real-world time duration which is provided by the motion planner, and then we discretise the segment with a constant $dt$. In this study, we set $dt = 0.01$ second for better simulation stability. A larger $dt$ will result in a too-high compounding error during simulation stepping, while a too-small $dt$ will demand too much computation. The statistics of the real and reconstructed motions are summarised in Table~\ref{tab:motions}. It can be seen that the validation motions are much longer than the optimisation ones. The MoveIt! trajectories and the reconstructed ones are saved as .npy files and will be open to the public upon acceptance.

\subsection{Loss functions}
We use four loss functions to calculate the difference between the simulated and real object states. As we are dealing with points and particles, it is natural to select point-based metrics. Therefore, we employ the two most common distance metrics for point sets, namely, the Chamfer distance (CD) and the earth mover's distance (EMD). Given two point sets $\mathcal{X}_0$ and $\mathcal{X}_1$, the CD and EMD are defined as follows.

\begin{equation*}
\begin{split}
d_{CD}(\mathcal{X}_0, \mathcal{X}_1) = &\frac{1}{|\mathcal{X}_0|}\sum_{\pmb{x}\in\mathcal{X}_0}\min_{\pmb{y}\in\mathcal{X}_1}||\pmb{x}-\pmb{y}||_2 +\\ 
&\frac{1}{|\mathcal{X}_1|}\sum_{\pmb{y}\in\mathcal{X}_1}\min_{\pmb{x}\in\mathcal{X}_0}||\pmb{x}-\pmb{y}||_2
\end{split}
\end{equation*}

$$
d_{EMD}(\mathcal{X}_0, \mathcal{X}_1) = \frac{1}{|\mathcal{X}_0|}\sum_{\pmb{x}\in\mathcal{X}_0}\min_{\phi^{EMD}:\mathcal{X}_0\rightarrow\mathcal{X}_1}||\pmb{x}-\phi(\pmb{x})||_2
$$

\noindent \xt{where $\pmb{x}$ and $\pmb{y}$ denote the 3D coordinates of the points}, $\phi^{EMD}$ denotes a one-to-one injective mapping that only exists when $|\mathcal{X}_0|\leq|\mathcal{X}_1|$. In practice, a linear assignment algorithm is used to calculate $\phi^{EMD}$. We also find that the three average operations reduce the magnitudes of the gradients of the physics parameters to an order of $4$ to $5$, which is undesired for optimisation; therefore, we use the distances without averaging the values over the point sets.

To calculate the loss between the real and simulated end configurations, one can use the original fusion point cloud $\mathcal{X}_T^{real}$ or the reconstructed particle systems $\mathcal{X}_T^{rec}$ as the target observations. Thus, there are two variants for each loss: $d_{PCD-CD}(\mathcal{X}_T^{real}, \mathcal{X}_T^{sim})$, $d_{PRT-CD}(\mathcal{X}_T^{rec}, \mathcal{X}_T^{sim})$, $d_{PCD-EMD}(\mathcal{X}_T^{real}, \mathcal{X}_T^{sim})$, $d_{PRT-EMD}(\mathcal{X}_T^{rec}, \mathcal{X}_T^{sim})$. When reconstructing $\mathcal{X}_T^{rec}$, we decrease the particle density for filling the voxels slowly until $|\mathcal{X}_T^{rec}|\leq|\mathcal{X}_T^{sim}|$ is satisfied. When computing losses involving the real-world point clouds, we downsample it with a voxel radius of $0.005$ so that the number of points in it is about $\frac{1}{6}$ of the number of particles in the reconstructed system. This is because a point cloud only contains points on the surface, while the reconstructed particle system is non-hollow.

Finally, although the CD and EMD losses are very common in calculating point set distance, they are not intuitive to visualise and understand. \xt{Also, as revealed by the experiment results shown in Figures~\ref{fig:loss-param-curves-lv1} and ~\ref{fig:loss-param-curves-lv2}, the CD and EMD losses focus on different spatial aspects of two point sets, which make them biased for result analysis across all loss functions. To make results easier to analyse and compare in a more unbiased way}, we further calculate the heightmaps of the fusion point cloud and the simulated particles, denoted as $\mathcal{I}_T^{real}$ and $\mathcal{I}_T^{sim}$. Specifically, we discretise a $0.11\times0.11 \text{m}^2$  square area centred at the centre of the object into a $32\times32$ grid. For each cell of the grid, we find the point or particle in the x-y range with the highest $z$ value and assign that value as the height of that cell. To compare the distance between two heightmaps, we simply take their summed pixel difference as follows.

$$
d_{Heightmap}=\sum_i\sum_j||\mathcal{I}_{T, ij}^{real}-\mathcal{I}_{T, ij}^{sim}||_2
$$

In practice, we use the fusion point cloud without downsampling to compute $\mathcal{I}_T^{real}$. However, a particle density that is too low will result in void cells when generating $\mathcal{I}_T^{sim}$, and we found that a density value of $4\times10^6$ particles per cubic metre is good enough for generating $32\times32$ heightmaps. The heightmap distance is only used for evaluation purposes because we found it leads to very poor optimisation results when used as a loss function.

\section{Results}
This section presents the design of the experiment and the result analyses including the optimisation and generalisation performances of the DPSI framework in identifying the physical characteristics of elastoplastic matter through simple robot manipulation.

\subsection{Experiment design}
The performance of the DPSI framework is examined in three steps: loss landscape analysis, in-distribution performance analysis and out-of-distribution generalisation analysis. The first step visualises the four loss functions (PCD-CD, PRT-CD, PCD-EMD, and PRT-EMD) to help understand how the physics parameters are related to the loss value distributions. 
The second step investigates the in-distribution performance of the proposed DPSI framework, i.e., whether it can produce realistic simulations for unseen object configurations with the same motions used in parameter identification. The loss landscape and in-distribution performance analyses are carried out at two levels of contact complexity. At each level, the performances are compared across four loss functions and five optimisation datasets (see data collection description below). 
The last step explores the out-of-distribution generalisability of the DPSI by inspecting the simulation accuracy of three unseen longer motions using identified parameters from the contact complexity level 2 experiment.

\noindent\textbf{\textit{Data collection}:} For each optimisation motion (i.e., Poking-1, Poking-2, Poking-shifting-1, Poking-shifting-2), we collect datapoints by executing the motions on the objects with different end-effectors. Each datapoint contains the multi-viewpoint point clouds of the object before and after a manipulation motion. For each of the two complexity levels, five optimisation datasets are created to examine the data hungriness of the proposed DPSI framework. 

With level (Lv.) 1 as an example, the first dataset is created with both motions (Poking-1 and Poking-2). Each motion is performed twice using three different end-effectors. Therefore, $6$ datapoints are created for each motion, and a total of $12$ datapoints are created for both motions. The second dataset for Lv. 1 is similar, except that only $1$ datapoint is created from each motion-effector pair, resulting in $3$ datapoints for each motion, and a total of $6$ datapoints for both motions. The other three datasets for Lv. 1 are more straightforward, with only one datapoint collected for each dataset. The three datapoints are created using the three end-effectors individually, performing the second motion from Lv. 1 (i.e. Poking-2).
We name these datasets as \textit{12-mix, 6-mix, 1-rectangle, 1-round, and 1-cylinder}, respectively. Data collection for Lv. 2 is similar, except that the two motions of Poking-1 and Poking-2 are replaced by Poking-shifting-1 and Poking-shifting-2, respectively. The datasets will be open-sourced upon the acceptance of the article.

For in-distribution validation, with each contact complexity level, we collect two extra datapoints from the second motion (i.e., Poking-2 and Poking-shifting-2) with all three end-effectors, resulting in $12$ validation datapoints. For out-of-distribution validation, where we collect two datapoints with each of the three long-horizon motions, resulting in $6$ datapoints.

For each datapoint, the real-world object is initially roughly shaped into a convex shape. We then acquire its point cloud by fusing the captured multi-view data. Next, we calculate the coordinates of the top centre of the object, where the end-effector tip will be moved to, and the motion will always be executed from there. We perform the same point cloud capturing and fusion process after a motion is executed. Each datapoint takes no more than $2$ minutes to collect, with most of the time spent on capturing point clouds.

\noindent\textbf{\textit{Optimisation \& validation}:} At each contact level, for each pair of loss function and dataset, gradient descent is carried out with the Adam algorithm \citep{adam2015} with three random seeds for $100$ iterations (gradient updates). For each datapoint, the simulation loads the initial particle configuration, simulates the motion and produces the resultant particle state, which is used to compute losses and gradients. Each optimisation iteration goes over all datapoints within a dataset and takes the average gradients to update the parameters. In-distribution validation is done after every gradient update with the validation dataset, simulating all datapoints and calculating the losses. 

Table~\ref{tab:lr} displays the step sizes and the value ranges of the parameters of interest for gradient updates. To determine the appropriate step sizes, we randomly sample $20$ sets of parameter values and compute the statistical means and standard deviations of the gradients for each parameter based on the \textit{12-mix} dataset. We then randomly initialise the parameters, run the Adam optimiser for $100$ gradient updates with uniformly randomly generated gradients using the statistical means and deviations with different step sizes, and select the step size that allows the randomised parameters to converge in about $70$ gradient updates. The value ranges of the parameters are selected according to either the definitions of the physics models~\citep{jones2009deformation,stomakhin2013material} or values reported by various research on manual parameter identification experiments with similar real-world materials, such as soil and clay~\citep{StructX,waheed2023evaluation}. However, a too-high value of Young's modulus $E$ was found to cause numerical instability in the simulation with the particles exploding away. Thus, an empirical upper bound for $E$ was determined by slowly increasing it until the simulation became unstable.

\begin{table}[h]
\footnotesize
\centering
\begin{tabular}{JFEFFEE}
\toprule
Parameter     & $E$ (kPa) & $\nu$ & $\rho$ (kg/m$^3$) & $\pmb{\sigma}_y$ (kPa) & $\eta_t$ & $\eta_m$\\
\midrule
Step size & $4000$ & $0.01$ & $10$ & $500$ & $0.01$ & $0.01$ \\
Min value     & $10000$  & $0.01$ & $1000$ & $1000$ & $0.01$ & $0.01$\\
Max value     & $300000$ & $0.48$  & $2000$ &$20000$ &$2.0$ &$2.0$ \\
\bottomrule
\end{tabular}
\caption{step sizes and the ranges of values for the parameters: Young's modulus $E$, Poisson's ratio $\nu$, yield stress $\pmb{\sigma}_y$, material density $\rho$, and the friction coefficients of the table $\eta_t$ and end-effector $\eta_m$.}
\label{tab:lr}
\end{table}

\subsection{Loss landscape analysis}
\begin{figure*}
    \centering
    \includegraphics[width=0.4\linewidth]{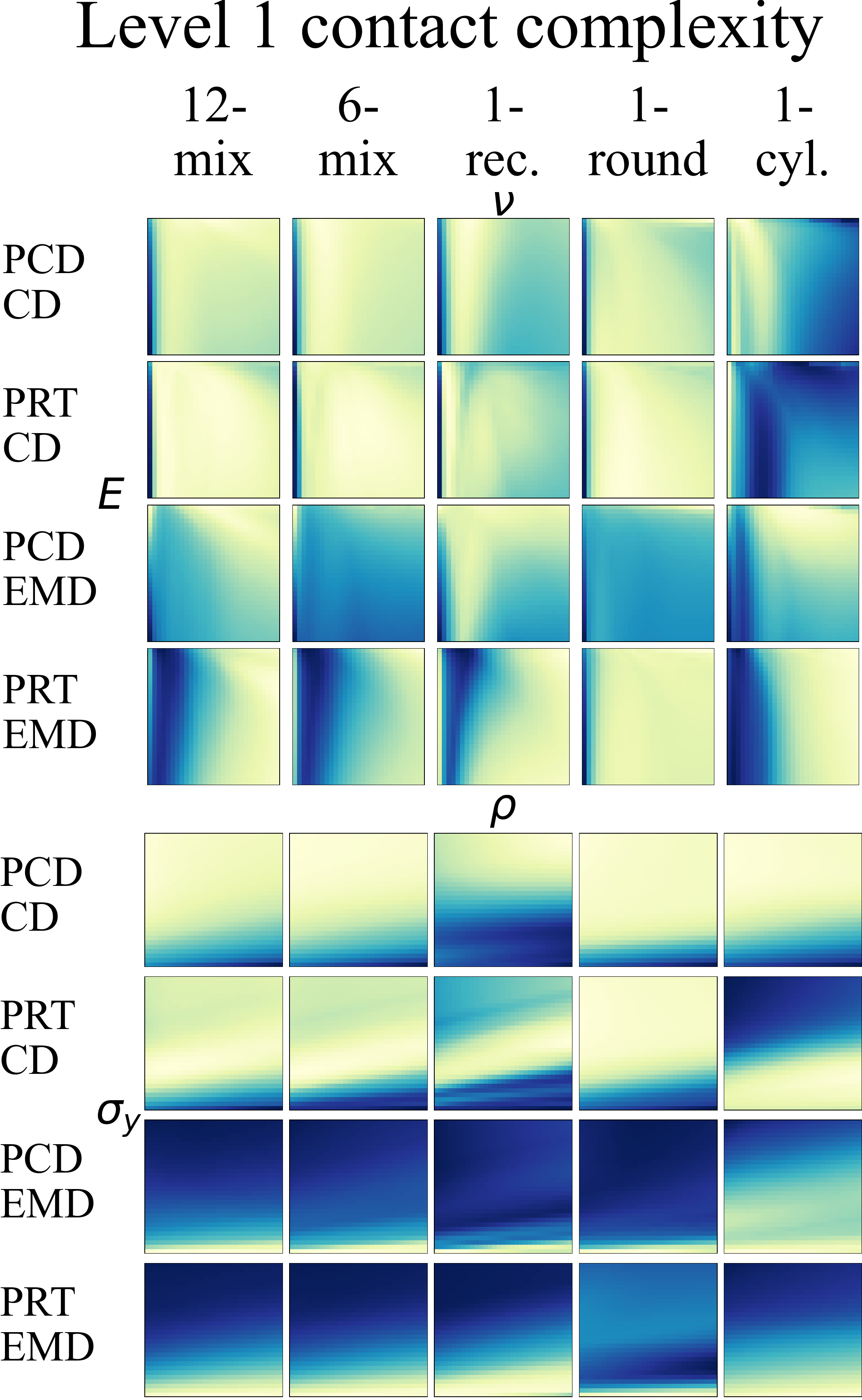}\hfil\includegraphics[width=0.4\linewidth]{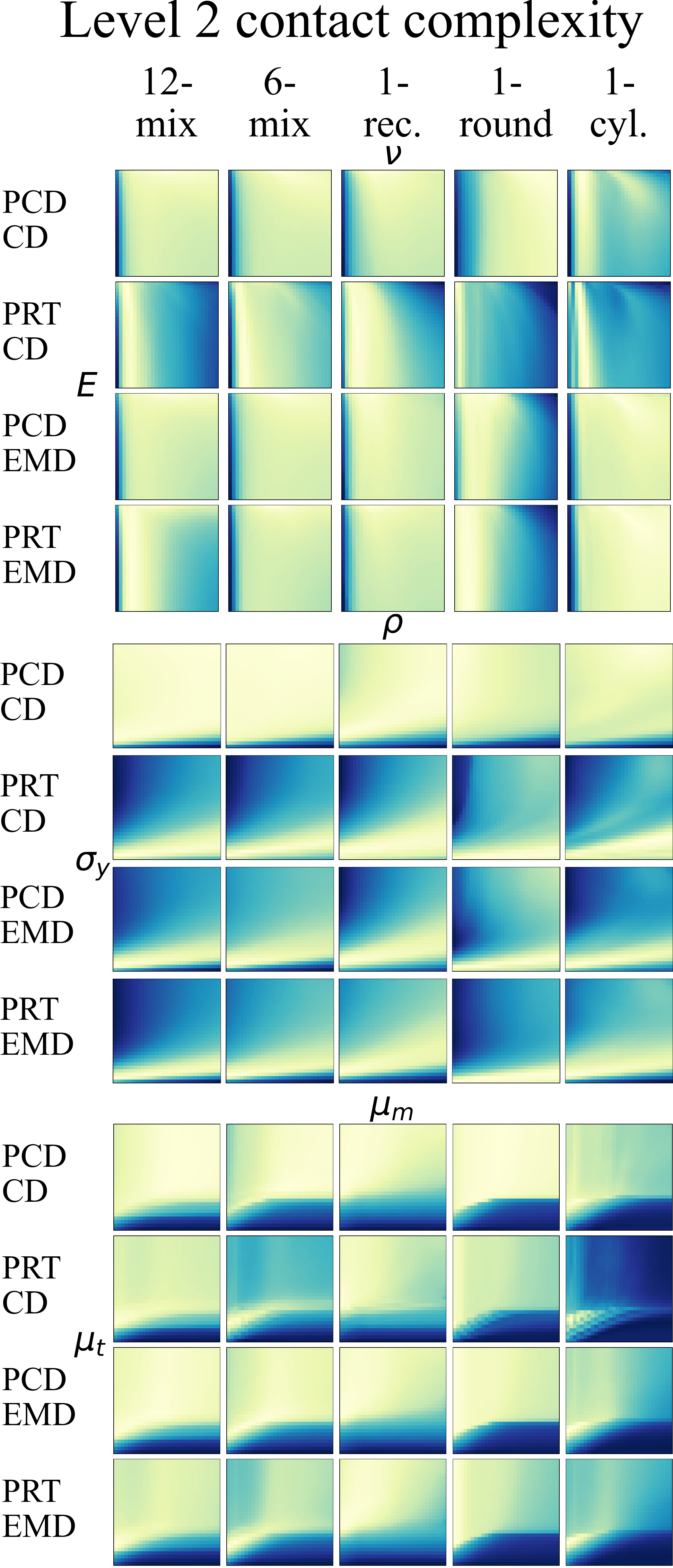}
    \caption{\small Loss landscapes (centralised to have zero means) at level-1 (left) and level-2 (right) contact complexity over pairs of physics parameters. Darker colours represent higher loss values. Parameters: Young's modulus ($E$) against Poisson ratio ($\nu$), yield stress ($\pmb{\sigma}_y$) against material density ($\rho$) for both levels, and table friction coefficient ($\mu_t$) against manipulator friction coefficient ($\mu_m$) for level 2. Other parameter values are fixed while computing the losses over one pair of parameters. Each column presents the losses computed with a different dataset. Each row presents the values of a different loss function. PCD: point cloud. PRT: particle system. CD: Chamfer distance. EMD: earth mover's distance.\label{fig:loss-landscapes}}
\end{figure*}

The loss landscapes, including point-cloud and particle chamfer distance (PCD CD, PRT CD), point-cloud and particle earth mover's distance (PCD EMD, PRT EMD), are computed over three pairs of physics parameters: $(E, \nu)$, $(\pmb{\sigma}_y, \rho)$ and $(\mu_g, \mu_e)$. For each pair, we set other parameters to fixed values and compute the losses with the parameters of interest discretised into $30$ intervals. Figure~\ref{fig:loss-landscapes} displays the loss landscapes computed at the two contact complexity levels and five datasets. To facilitate distribution-level comparison, the loss landscapes are \xt{centralised to have zero means (subtracting the mean of the matrix)}. The following observations can be drawn.

Firstly, we compare the landscapes in Figure~\ref{fig:loss-landscapes} vertically to examine the sensitivity of the loss functions against different parameters. At contact complexity level 1, it shows that the CD and EMD losses exhibit quite similar value changing directions along the $E$, $\nu$ and $\rho$ axes, while quite the opposite directions along the $\pmb{\sigma}_y$ axis. This can be observed from the vertically reversed colour distributions between the first and last two rows of landscapes against $\pmb{\sigma}_y$ and $\rho$. On the other hand, the losses at level 2 contact complexity exhibit similar distributional patterns along all parameter axes. In addition, at level-2 contact complexity, the PRT CD loss is more sensitive (exhibiting more drastic value changes) along the $\nu$ axis than other losses (see the second row of the level-2 landscapes against $E$ and $\nu$). In contrast, the values of the PRT CD and both EMD losses are more sensitive along the $\pmb{\sigma}_y$ axis (the last three rows of the level-2 landscapes against $\pmb{\sigma}_y$ and $\rho$). The level-2 landscapes against the friction coefficients ($\mu_t$ and $\mu_m$) show that all losses distribute very similarly over these two parameters. Overall, the EMD losses seem to be more sensitive to all the parameters, as their loss landscapes tend to be less flat.

These observations indicate that, with the adopted von Mises plasticity model, whose plastic deformations are governed by the yield stress ($\pmb{\sigma}_y$) of the material, the CD and EMD losses may focus on different aspects of the point sets, as they tend to disagree with each other distributionally along the $\pmb{\sigma}_y$ axis. Also, we may expect the EMD losses to perform better in parameter identification because they exhibit higher sensitivity against most parameters.

Secondly, by comparing the landscapes horizontally, Figure~\ref{fig:loss-landscapes} shows the distributional patterns of most (not all) loss landscapes are quite similar across the five datasets. This means that the loss values, and thus the optimisation processes, are not sensitive to the number of datapoints used in computation. This observation allows us to expect the recovery of physics parameters with small data, even with a single interaction experience.

Thirdly, many of the loss distributions display large areas of flat regions, where the loss values are very similar. This indicates many optimisation saddle points and that multiple parameter value combinations may serve as plausible solutions, describing the same physical characteristics. Whether this is an issue from an inaccurate physics model or a general fact of the real-world dynamics remains to be determined. 

\subsection{In-distribution performances}
This subsection investigates the in-distribution performance of the parameter identification task through gradient descent at two levels of contact complexity. For each level, experiments are conducted with the four loss functions and five datasets. To evaluate the performances thoroughly, the four loss functions and a heightmap-based distance function are used to evaluate the differences between the real and simulated manipulation results using the in-distribution validation dataset.
This enables the observation of the influences on other distance functions from minimising each objective. In particular, this subsection investigates the following questions.

\begin{itemize}
    \item Can the loss functions be minimised (does optimisation converge)?
    \item Do the loss functions agree with each other in such system identification tasks?
    \item If local minima appear and multiple solutions exist, do they produce visually distinct manipulation results?
    \item How does the number of datapoints affect optimisation?
    \item Does DPSI produce parameter values that are physically realistic and interpretable? \xt{Are they close to the parameter values reported in relevant literature and studies?}
\end{itemize}

\begin{figure*}
    \centering
    \includegraphics[width=0.6\linewidth]{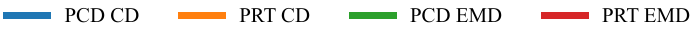}\\
    \includegraphics[width=0.8\linewidth]{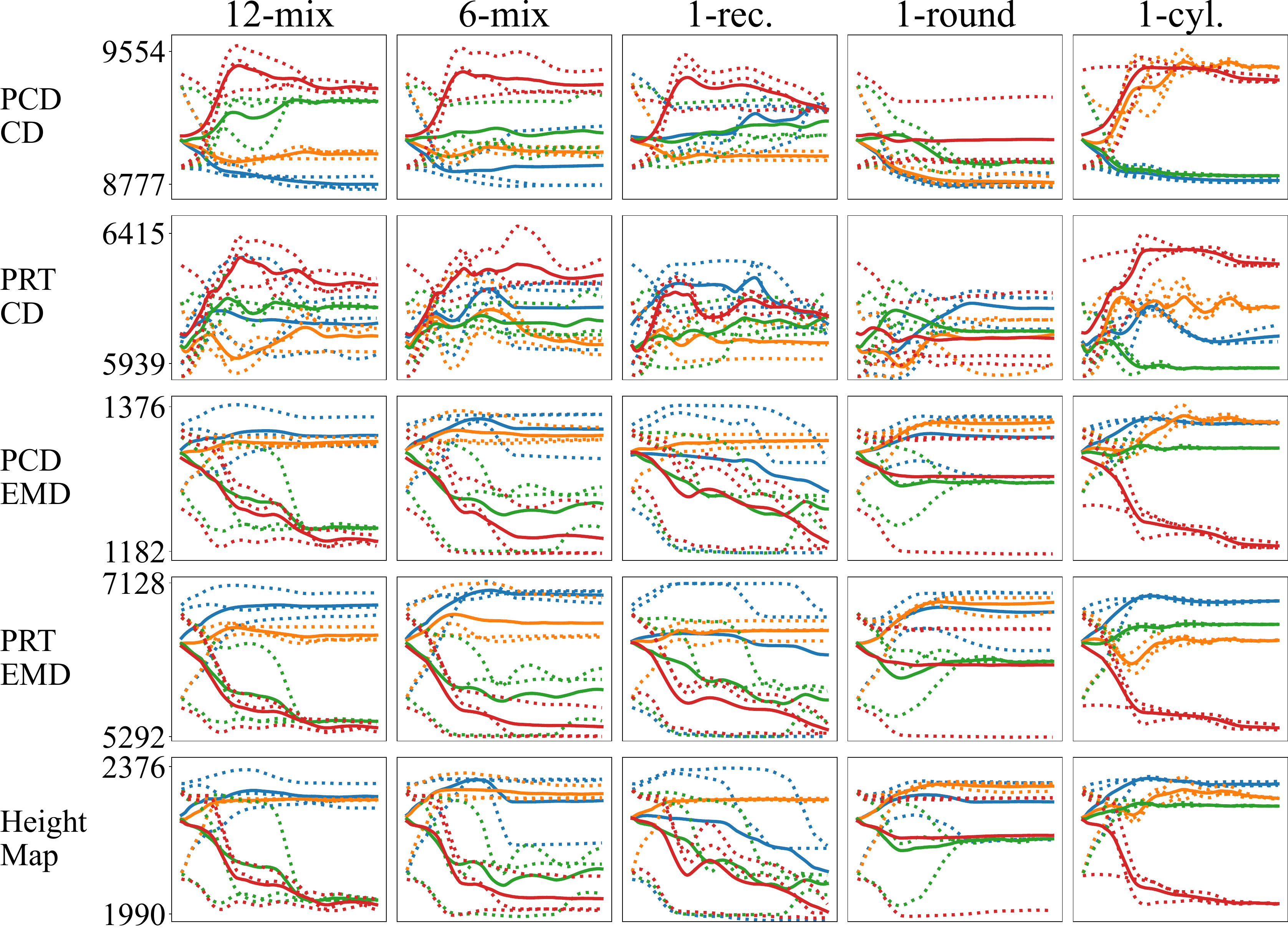}\\
    \includegraphics[width=0.8\linewidth]{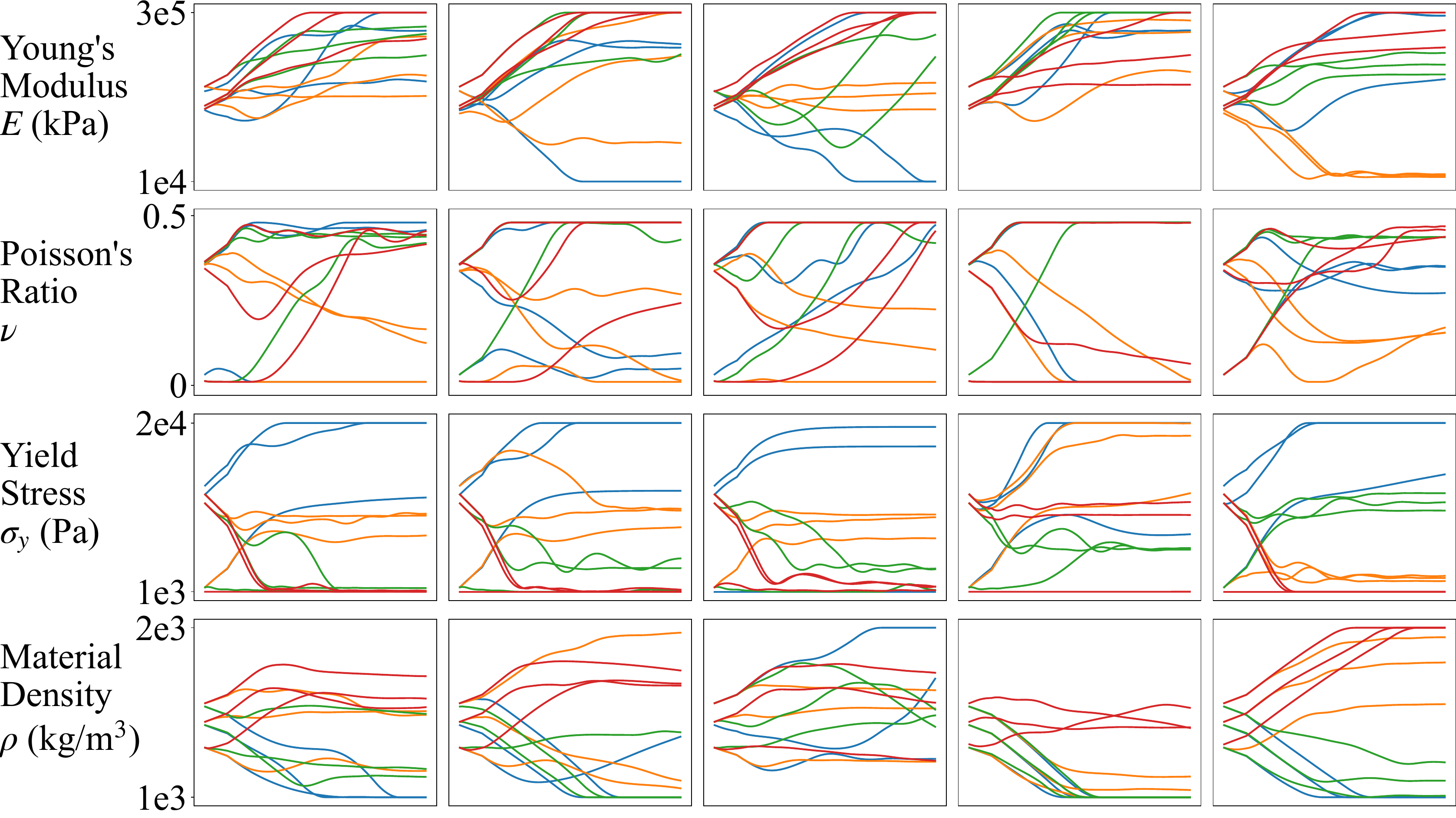}
    \caption{\small Validation losses (top five rows) and parameter values (last four rows) over $100$ gradient updates at level-1 contact complexity. Each column presents the results of optimising with a different dataset. Each row shows the changes of an evaluation metric or a parameter, denoted on the left. In each figure, different colours indicate the results of minimising a different loss function, as labelled by the legend on the top. For the top five rows, each colour has three dotted lines corresponding to the results of three random seeds and a solid line corresponding to their means. For the last four rows, each colour has three solid lines corresponding to the results of three random seeds.}
    \label{fig:loss-param-curves-lv1}
\end{figure*}

\begin{figure*}
    \centering
    \includegraphics[width=0.6\linewidth]{note_images/legend.pdf}\\
    \includegraphics[width=0.8\linewidth]{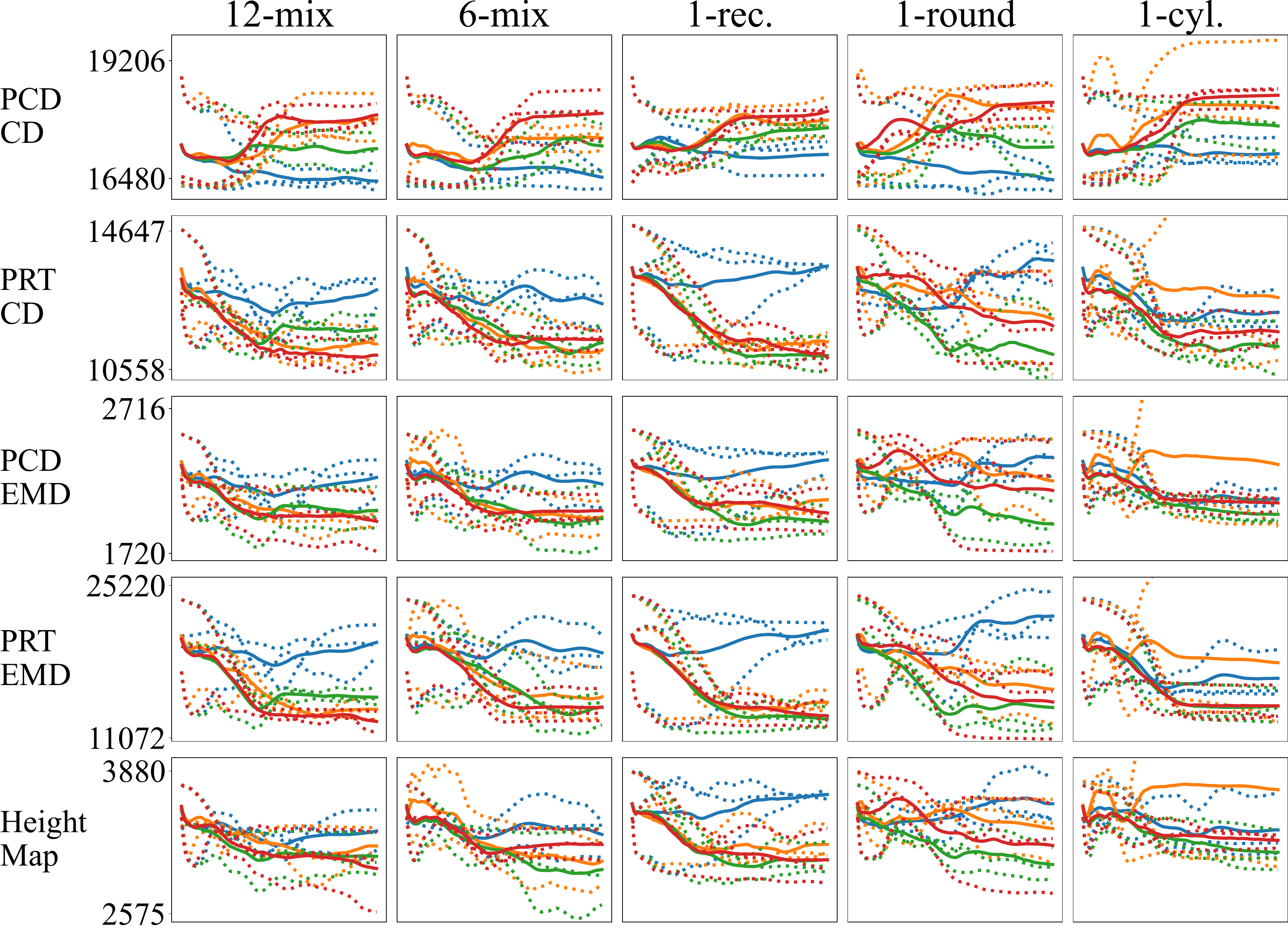}\\
    \includegraphics[width=0.8\linewidth]{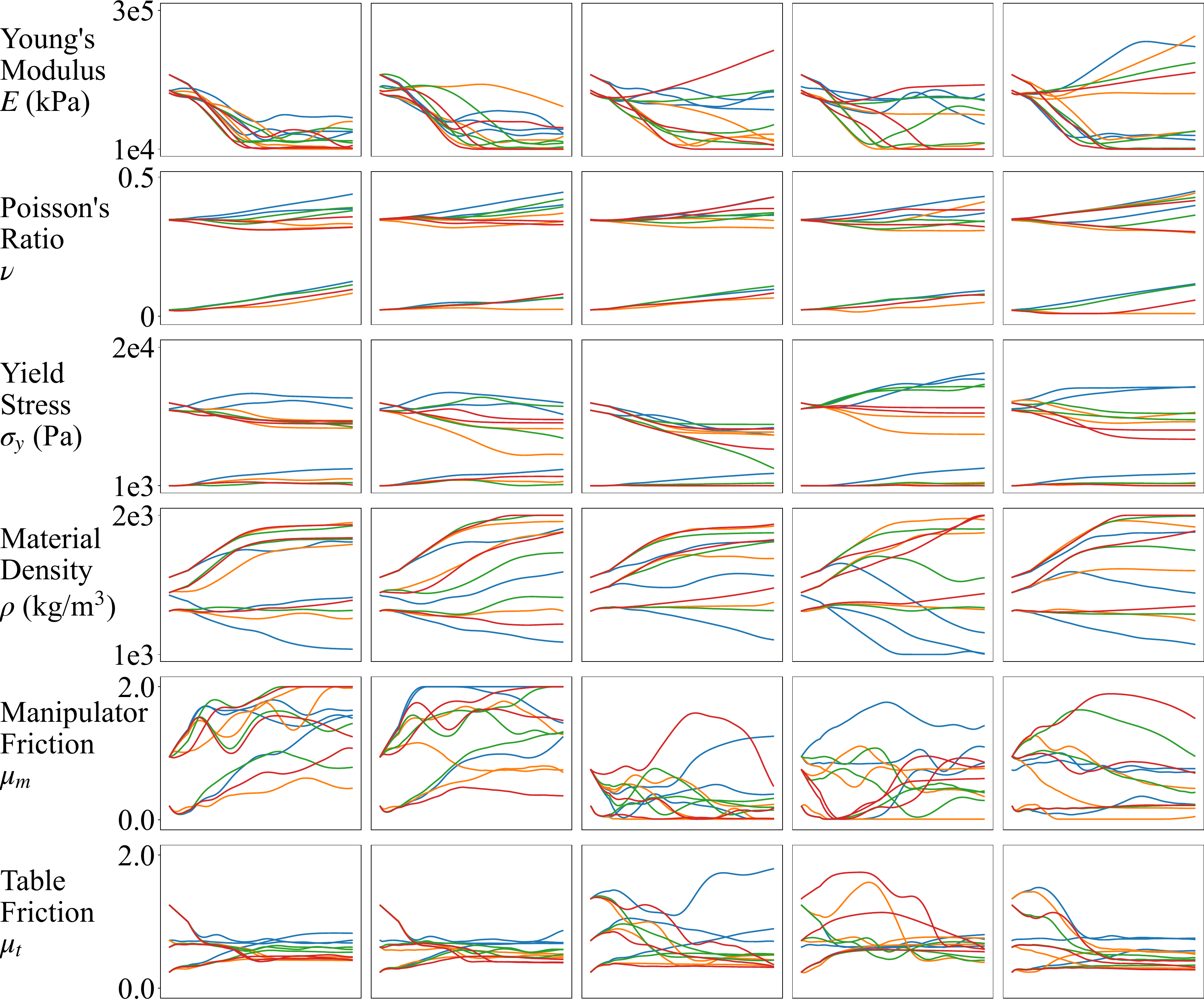}
    \caption{\small Validation losses (top five rows) and parameter values (last six rows) over $100$ gradient updates at level-2 contact complexity. Each column presents the results of optimising with a different dataset. Each row shows the changes of an evaluation metric or a parameter, denoted on the left. In each figure, different colours indicate the results of minimising a different loss function, as labelled by the legend on the top. For the top five rows, each colour has three dotted lines corresponding to the results of three random seeds and a solid line corresponding to their means. For the last four rows, each colour has three solid lines corresponding to the results of three random seeds.}
    \label{fig:loss-param-curves-lv2}
\end{figure*}

\noindent\textbf{\textit{Quantitative results:}} We start by analysing the quantitative results of the parameter identification task. The changes of the validation losses and parameter values over the course of optimisation at both contact complexity levels are shown in Figure~\ref{fig:loss-param-curves-lv1} and~\ref{fig:loss-param-curves-lv2}. The best parameter values corresponding to the lowest validation heightmap loss among the three random seeds are summarised in Table~\ref{tab:best-param}.

First of all, we start by observing the tendency of convergence. The top five rows in Figure~\ref{fig:loss-param-curves-lv1} and~\ref{fig:loss-param-curves-lv2} show that, at both contact complexity levels, most of the individual validation loss curves (dotted lines) tend to stabilise and converge, which indicates that DPSI can effectively converge to local minima. The last four rows in Figure~\ref{fig:loss-param-curves-lv1} and the last six rows in Figure~\ref{fig:loss-param-curves-lv2} reveal that the parameters converge to different solutions. Table~\ref{tab:best-param} also shows that multiple parameter solutions exist for a low validation heightmap distance at both levels of experiments. These observations mean the parameter identification task at both contact complexity levels does converge but has multiple local minima and solutions, aligned with the large flat regions observed from the loss landscapes.

\begin{table*}
\scriptsize
\centering
\begin{tabular}{DOl|BBBBB|C}
\toprule
                   &     &         & \multicolumn{5}{c|}{Dataset} & \\
Contact complexity &Loss & Metric/Parameter & 12-mix  & 6-mix   & 1-rec.   & 1-round & 1-cyl.& Avg. Height map (mm) \\
\midrule
&                    & Min. height map (mm)      & 2267.44 & 2176.21 & \textbf{1973.81}  & 2183.30   & 2327.01& 2185.55\\
&\multirow{2}{*}{PCD}& Young's modulus (kPa)     & 300000  & 10000   & 300000   & 300000    & 294991&\\
&\multirow{2}{*}{CD} & Poisson's ratio   & 0.480   & 0.480   & 0.480    & 0.480     & 0.348&\\
&                    & Yield stress    (kPa)  & 20000   & 12367    & 1000     & 7527      & 20000&\\
&                    & Material density (kg/m$^3$) & 1000    & 1380    & 1225     & 1000      & 1000&\\
\cline{2-9}
&                    & Min. height map (mm)      & 2287.21 & 2293.42 & 2285.98  & 2311.87   & 2290.71 &2273.83\\
&\multirow{2}{*}{PRT}& Young's modulus (kPa)     & 189900  & 300000  & 179373   & 196393    & 20335\\
&\multirow{2}{*}{CD} & Poisson's ratio   & 0.165   & 0.010   & 0.223    & 0.010     & 0.178\\
&                    & Yield stress    (kPa)  & 7373    & 10286   & 7016     & 12358     & 2466\\
Level&                    & Material density (kg/m$^3$) & 1157    & 1088    & 1210     & 1000      & 1797\\
\cline{2-9}
1&                    & Min. height map (mm)     & 2024.78 & 2039.95 & 2040.51  & 2186.90   & 2272.50& 2112.93\\
&\multirow{2}{*}{PCD}& Young's modulus (kPa)     & 265420  & 237337  & 270940   & 300000    & 193790\\
&\multirow{2}{*}{EMD} & Poisson's ratio  & 0.445   & 0.439   & 0.419    & 0.480     & 0.436\\
&                    & Yield stress    (kPa)  & 1403    & 1036     & 1382     & 5839      & 10138\\
&                    & Material density (kg/m$^3$) & 1487    & 1379    & 1482     & 1000      & 1009\\
\cline{2-9}
&                    & Min. height map (mm)      & 2000.36 & \textbf{1999.14 }  & \textbf{1980.35}  & 1999.34   & 2016.10 &1999.06\\
&\multirow{2}{*}{PRT}& Young's modulus (kPa)     & 300000  & 300000  & 300000   & 300000    & 271247\\
&\multirow{2}{*}{EMD}& Poisson's ratio   & 0.453   & 0.480   & 0.480    & 0.478     & 0.442\\
&                    & Yield stress    (kPa)  & 1000    & 1000     & 1059     & 1000      & 1000\\
&                    & Material density (kg/m$^3$) & 1542    & 1670    & 1214     & 1518      & 2000\\
\cline{2-9}
&\multicolumn{7}{l}{\xt{*Validation height map (mm) with hand-picked parameters: 2122.59}}\\
\hline
\hline
&                    & Min. height map (mm)           & 3163.64  & 3091.73  & 3622.58  & 3423.00  & 3125.04 &3285.20 \\
&\multirow{4}{*}{PCD}& Young's modulus (kPa)          & 47700  & 42136  & 91785  & 132840  & 28679  \\
&\multirow{4}{*}{CD} & Poisson's ratio        & 0.444  & 0.451  & 0.364  & 0.095  & 0.403  \\
&                    & Yield stress    (kPa)       & 11407  & 10479  & 7839  & 15590  & 14514  \\
&                    & Material density (kg/m$^3$)      & 1810  & 1919  & 1819  & 1000  & 1880  \\
&                    & Table friction       & 0.725  & 0.696  & 1.816  & 0.696  & 0.715  \\
&                    & Manipulator friction & 1.495  & 1.448  & 1.264  & 1.037  & 0.704  \\
\cline{2-9}
&                    & Min. height map (mm)      & 3088.31  & 2865.14  & 3104.15  & 3031.56  & 2999.36 & 3017.70 \\
&\multirow{4}{*}{PRT}& Young's modulus (kPa)     & 66514  & 47786  & 28023  & 24068  & 256786  \\
&\multirow{4}{*}{CD} & Poisson's ratio   & 0.331  & 0.376  & 0.317  & 0.309  & 0.451  \\
&                    & Yield stress    (kPa)  & 1904  & 1809  & 8363  & 8063  & 1537  \\
&                    & Material density (kg/m$^3$) & 1267  & 1284  & 1926  & 1965  & 1225  \\ 
&                    & Table friction  & 0.473  & 0.488  & 0.414  & 0.419  & 0.487  \\
Level&                    & Manipulator friction & 0.494  & 0.604  & 0.249  & 0.293  & 0.084  \\
\cline{2-9}
2&                    & Min. height map (mm)       & 2933.23  & \textbf{2667.43}  & 2959.43  & 2895.74  & 2979.04 & 2886.97\\
&\multirow{4}{*}{PCD}& Young's modulus (kPa)     & 48459  & 14959  & 135207  & 113474  & 12017  \\
&\multirow{4}{*}{EMD} & Poisson's ratio  & 0.392  & 0.400  & 0.374  & 0.339  & 0.369  \\
&                    & Yield stress    (kPa)  & 1417  & 1176  & 1359  & 1340  & 10921  \\
&                    & Material density (kg/m$^3$) & 1319  & 1409  & 1313  & 1331  & 1995  \\
&                    & Table friction & 0.544  & 0.525  & 0.508  & 0.453  & 0.434  \\
&                    & Manipulator friction & 0.805  & 1.308  & 0.194  & 0.381  & 0.878  \\
\cline{2-9}
&                    & Min. height map (mm)       & \textbf{2598.28}  & 3060.10  & 2847.69  & \textbf{2760.83}  & 3172.49 & 2887.88\\
&\multirow{4}{*}{PRT}& Young's modulus (kPa)     & 27779  & 10000  & 222856  & 144827  & 10000  \\
&\multirow{4}{*}{EMD} & Poisson's ratio  & 0.359  & 0.330  & 0.435  & 0.383  & 0.303  \\
&                    & Yield stress    (kPa)  & 1008  & 9635  & 1000  & 1000  & 7377  \\
&                    & Material density (kg/m$^3$) & 1396  & 2000  & 1492  & 1445  & 2000  \\
&                    & Table friction   & 0.487  & 0.373  & 0.311  & 0.543  & 0.411  \\
&                    & Manipulator friction & 1.013  & 1.478  & 0.015  & 0.625  & 1.433  \\
\cline{2-9}
&\multicolumn{7}{l}{\xt{*Validation height map (mm) with hand-picked parameters: 2985.5688}}\\
\hline
\end{tabular}
\caption{The parameter values corresponding to the lowest validation heightmap distance in each optimisation case at both contact complexity levels. Bolded texts denote the three lowest losses at each level. \xt{It can be seen that the PRT EMD loss tends to produce the best results and DPSI can find parameters that outperform hand-picked parameters (from the geotechnical literature) with even just one interaction datapoint.}}
\label{tab:best-param}
\end{table*}

However, the parameter distributions in Figure~\ref{fig:loss-param-curves-lv2} clearly show that some parameters converge to much smaller and distinct value regions than those found in the level-1 experiments.
More specifically, the values of Young's modulus ($E$), Poisson's ratio ($\nu$) and yield stress ($\pmb{\sigma}_y$) become more converged, while they are more dispersed in the level-1 results. Interestingly, the material density ($\rho$) remains quite dispersed at both contact complexity levels. This is not surprising because, compared to the level-1 motions, the higher contact complexity induced by the shifting parts of the level-2 motions naturally poses a stricter selection range for the physics parameter values. 

\xt{Also, some parameter values are closer to empirical studies of soft/hard/saturated clay in the level-2 experiments. For example, Young's modulus values are closer to the reported range of $5000$ to $54000$ kPa, the material density seems to be closer to $1400$ kg/m$^3$, and Poisson's ratio is closer to the reported range of $0.4$ to $0.5$~\citep{waheed2023evaluation,StructX}. However, the recovered yield stress values are higher than the reported values for clays ($210$ to $600$ kPa)~\citep{rehman2018yield} because the reported values were found to make the material collapse in simulation. 
Based on the empirical values from~\cite{waheed2023evaluation,StructX} and the loss landscapes, we hand-pick a set of parameter values and calculate the validation losses at both levels. As denoted in Table~\ref{tab:best-param}, the resultant performances are reasonable but not as good as the performances of the DPSI method.}

Secondly, the top five rows in Figure~\ref{fig:loss-param-curves-lv1} reveal that, in most cases, the CD and EMD losses have a negative correlation. For example, the red curves of the first column show that minimising the PRT EMD loss reduces both EMD losses and the heightmap distance, but increases both CD losses. Also, Table~\ref{tab:best-param} shows that optimising the CD losses tends to produce large yield stress values ($\pmb{\sigma}_y$), while optimising the EMD losses tends to produce small ones. Similar observations can be made from the level-2 results, but only the PCD CD validation losses tend to exhibit opposite tendencies compared to the other losses (observe the blue lines that deviate from others in Figure~\ref{fig:loss-param-curves-lv2}). This is aligned with the analyses of the loss landscapes at both contact complexity levels, where reversed distributional patterns of the CD and EMD losses were observed. Furthermore, the EMD losses tend to produce smaller validation heightmap distances at both levels as shown by the green and red lines in Figure~\ref{fig:loss-param-curves-lv1} and~\ref{fig:loss-param-curves-lv2} and the average heightmap distances in Table~\ref{tab:best-param}, indicating better sim2real alignment. This again is aligned with the analysis of the level-2 loss landscapes, where only the PCD CD loss distributes significantly differently along the $\pmb{\sigma}_y$ axis.

Thirdly, the results show that it is possible to obtain comparable performances with just one datapoint at both contact complexity levels. This can be concluded by comparing the results horizontally: 1) most validation curves of the top five rows in Figs.~\ref{fig:loss-param-curves-lv1} and \ref{fig:loss-param-curves-lv2} with 1-datapoint show highly similar tendencies with the 6-mix and 12-mix datasets, and 2) the last four rows in Figs.~\ref{fig:loss-param-curves-lv1} and last six rows in Figure~\ref{fig:loss-param-curves-lv2} also show that the parameters found with different datasets mostly converge to similar value regions. Moreover, Table~\ref{tab:best-param} shows that the lowest validation heightmap distances at level-1 contact complexity and the third lowest at level-2 are achieved by optimising with the 1-rec. and 1-round datasets. These results are aligned with the analysis of the loss landscapes, where similar loss distributions are observed among different datasets.

\begin{figure*}
\centering
\begin{subfigure}{\columnwidth}
\caption{Level-1 contact, object configuration 1}\label{subfig:in-dist-vis-lv1-obj1}
\includegraphics[width=0.99\columnwidth]{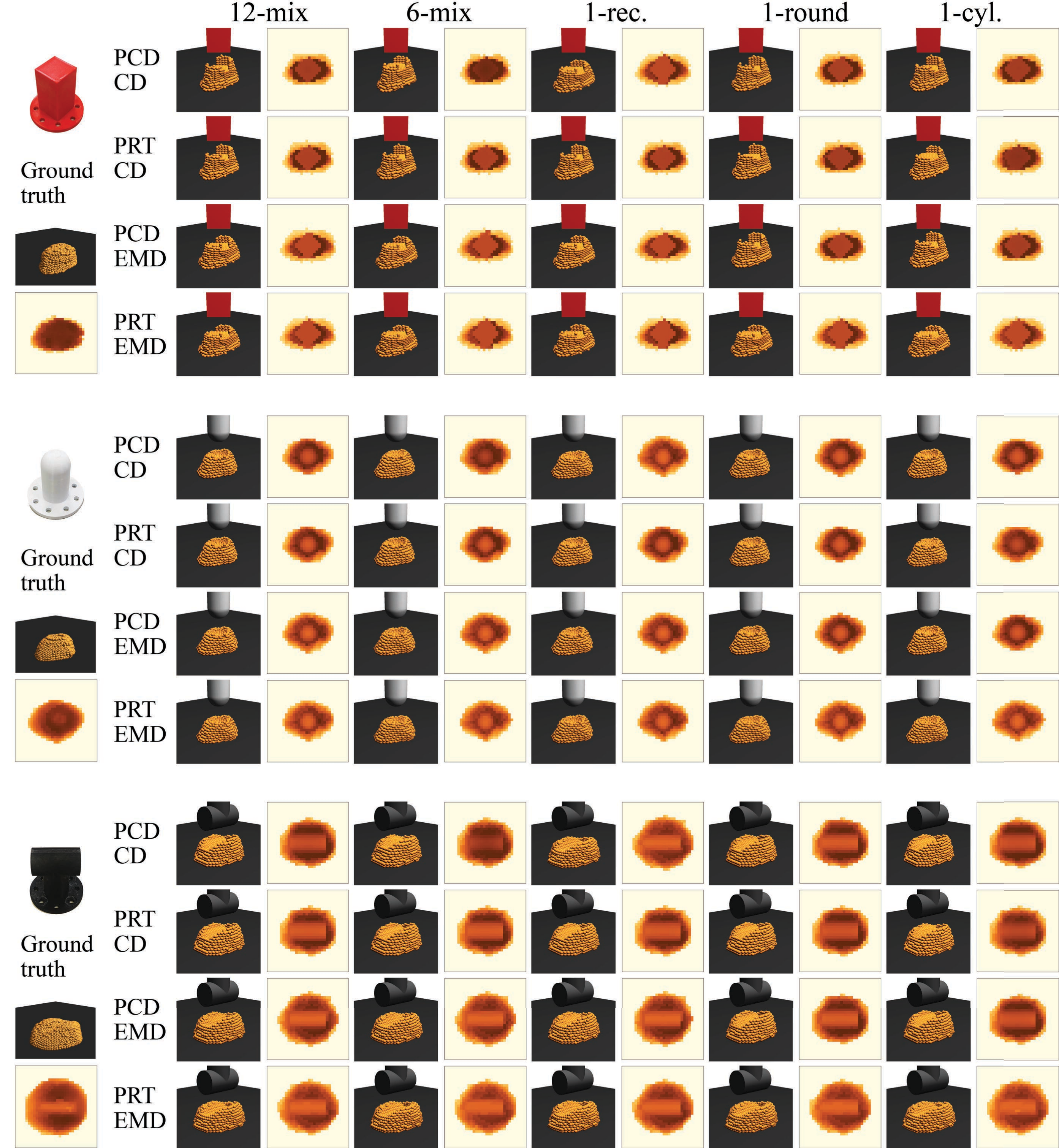}
\end{subfigure}\hfil
\begin{subfigure}{\columnwidth}
\caption{Level-1 contact, object configuration 2}\label{subfig:in-dist-vis-lv1-obj2}
\includegraphics[width=0.99\columnwidth]{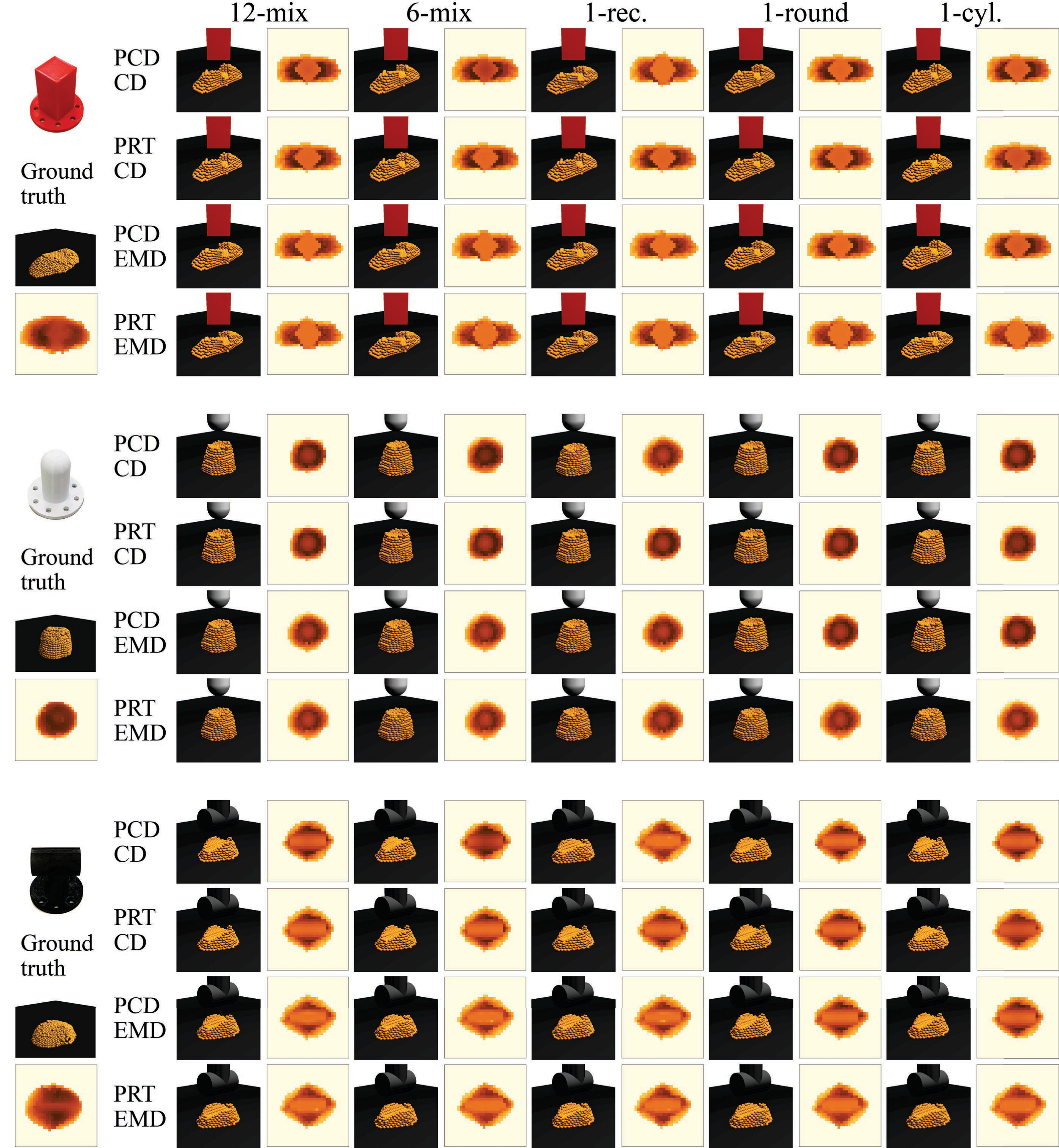}
\end{subfigure}\\
\begin{subfigure}{\columnwidth}
\caption{Level-2 contact, object configuration 1}\label{subfig:in-dist-vis-lv2-obj1}
\includegraphics[width=0.99\columnwidth]{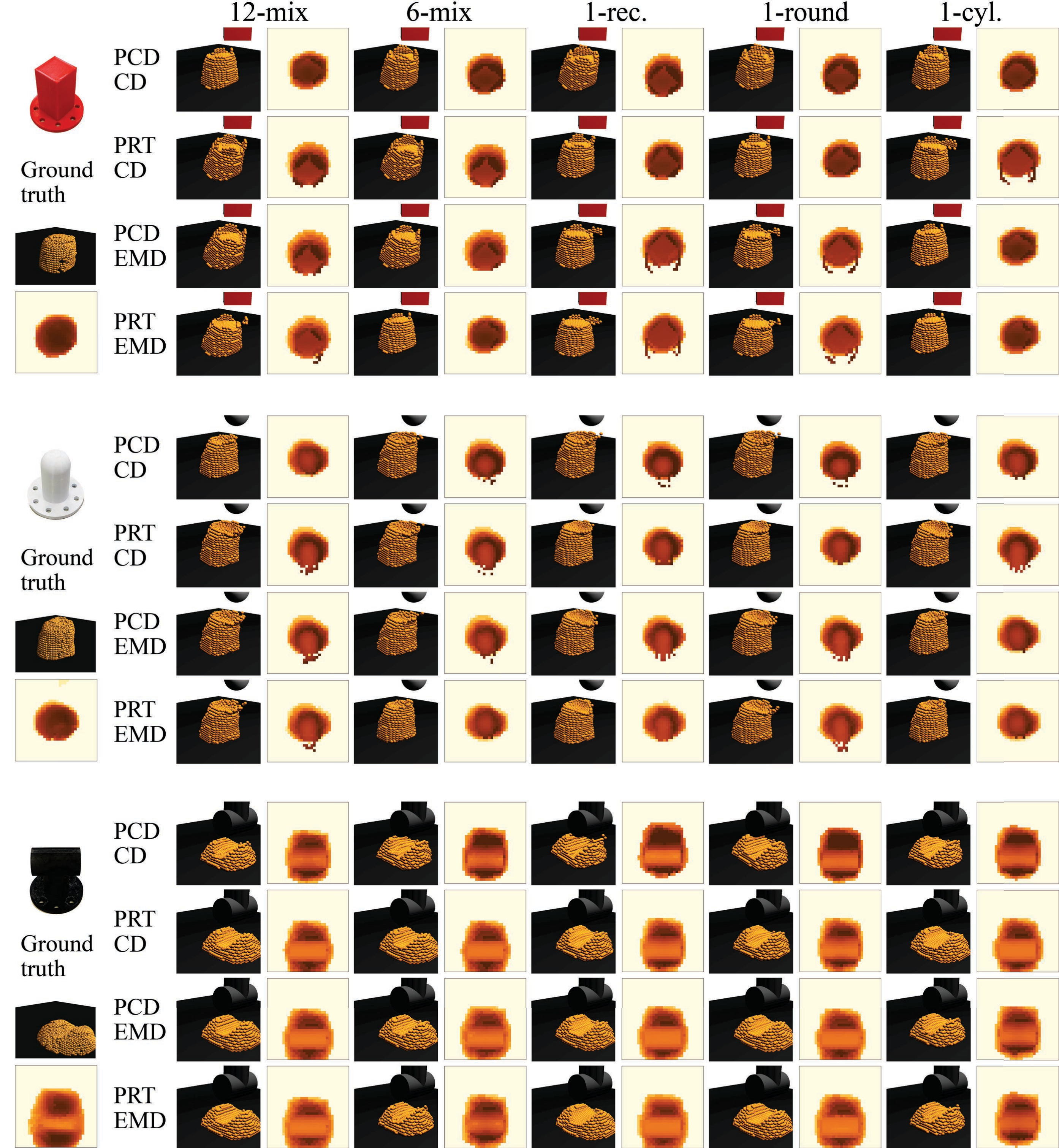}
\end{subfigure}\hfil
\begin{subfigure}{\columnwidth}
\caption{Level-2 contact, object configuration 2}\label{subfig:in-dist-vis-lv2-obj2}
\includegraphics[width=0.99\columnwidth]{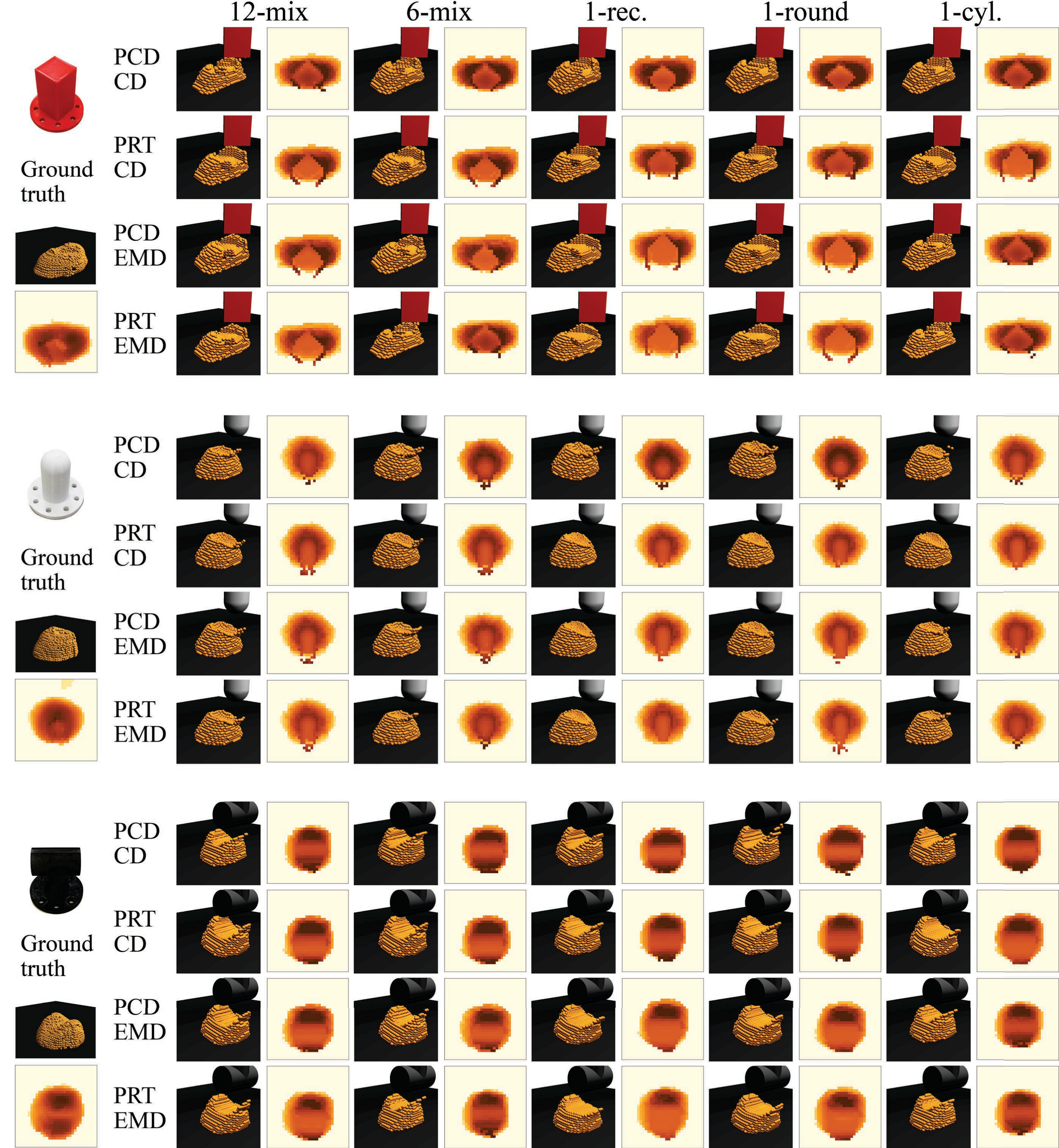}
\end{subfigure}
\caption{The particles and heightmaps of the objects after applying the second motions at contact level 1 (a-b) and 2 (c-d), simulated with the best set of physics parameters. Darker colours of the heightmaps indicate greater heights of the object. For each combination of the loss function and dataset at each level, three trajectories on two object configurations are simulated (3 effectors $\times$ 2 datapoints). There are three groups in each subfigure, with the ground truth point clouds and heightmaps placed in the leftmost column. In each group, a row shows the results corresponding to a loss function and a column results corresponding to a dataset.}
\label{fig:in-dist-vis}
\end{figure*}

\noindent\textbf{\textit{Qualitative results}:} We now examine the manipulation processes simulated using the best physics parameters corresponding to the lowest validation heightmap losses presented in Table~\ref{tab:best-param}. Figure~\ref{fig:in-dist-vis} shows the resultant particles and heightmaps after the manipulation of objects at two contact complexity levels with three end-effectors (two object configurations are included for each contact level and end-effector).

Firstly, all the simulated particles and heightmaps post-manipulation are highly similar to the ground truths across the loss functions and datasets. This shows that minimising any of the four loss functions with any of the five datasets can indeed reproduce visually plausible manipulation results close to the real-world system. It indicates that DPSI is not data-hungry and it is robust to the choice of common point-based loss objectives. These characteristics are highly preferred in robotic applications, as they lead to smaller data collection costs, simpler observation preprocessing, and simpler loss function engineering.

\begin{figure}[h]
    \centering
    \includegraphics[width=0.99\columnwidth]{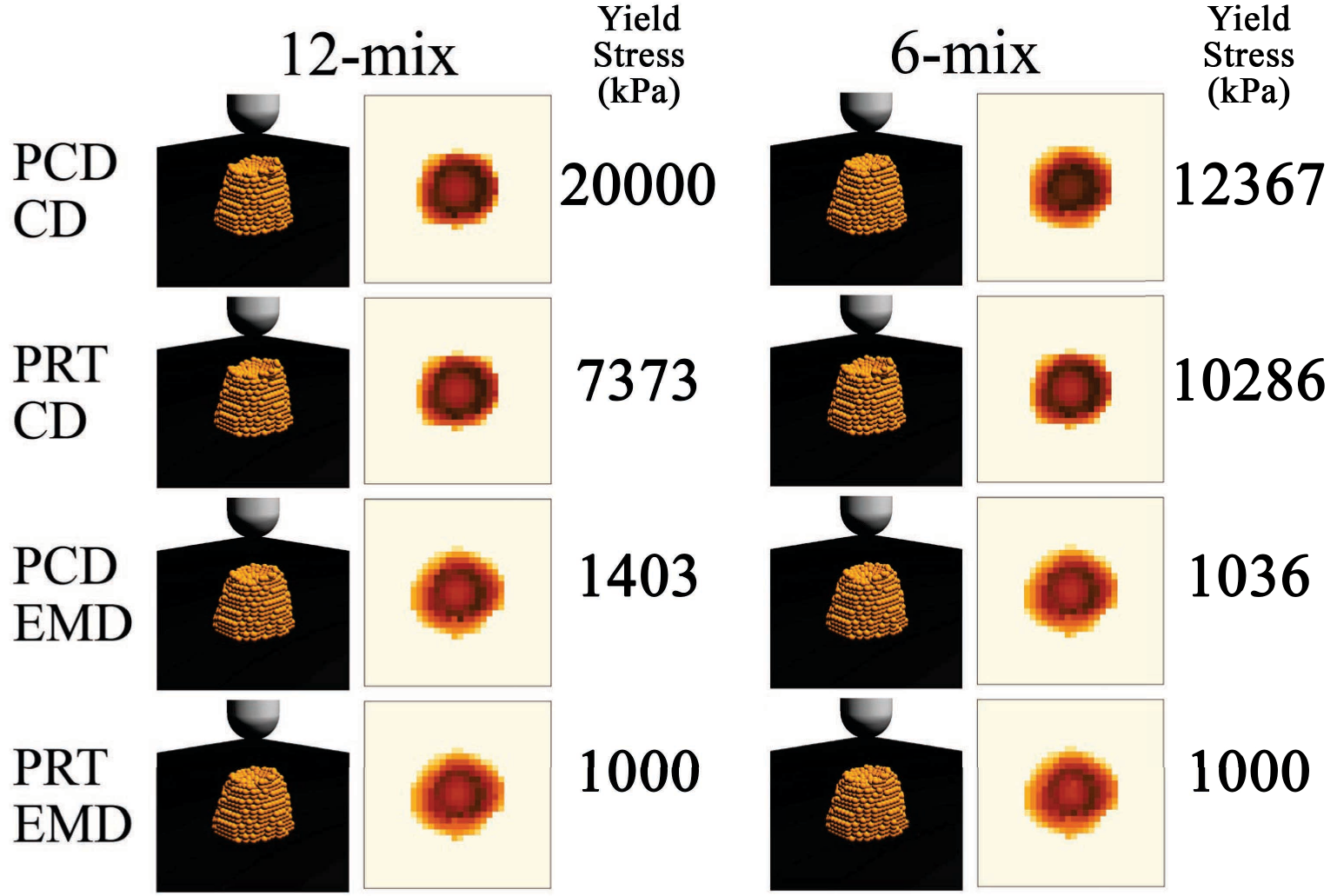}
    \caption{Selected examples from Figure~\ref{subfig:in-dist-vis-lv1-obj2} to show that the CD losses produce darker-coloured heightmaps while the EMD losses produce lighter-coloured and wider heightmaps, which can be intuitively explained by the optimised yield stress values and their potential difference in spatial focus.}
    \label{fig:example-1}
\end{figure}

Secondly, by examining the details of the heightmaps more carefully, it shows that optimising the CD losses tends to produce heightmaps with darker colours (which represent greater heights of the object) while optimising the EMD losses tends to produce heightmaps where the objects look slightly bigger and wider. This is especially true at level-1 contact complexity, \xt{where the heightmaps in the first two rows for each object configuration in Subfigures~\ref{subfig:in-dist-vis-lv1-obj1} and~\ref{subfig:in-dist-vis-lv1-obj2} are generally darker than those in the last two rows (see figure~\ref{fig:example-1} for enlarged examples)}. At level 2, the PRT CD loss sometimes changes its focus on the $x$ and $y$ directions, producing lighter-colour heightmaps (see the second rows of the heightmaps associated with the rectangle and round end-effectors in Subfigures~\ref{subfig:in-dist-vis-lv2-obj1} and~\ref{subfig:in-dist-vis-lv2-obj2}). This observation indicates that optimising the CD losses tends to produce particles that match the height of the ground truth shapes, i.e., in the $z$ direction, while optimising the EMD losses tends to match the $x$ and $y$ directions.

\begin{figure}[h]
    \centering
    \includegraphics[width=0.85\columnwidth]{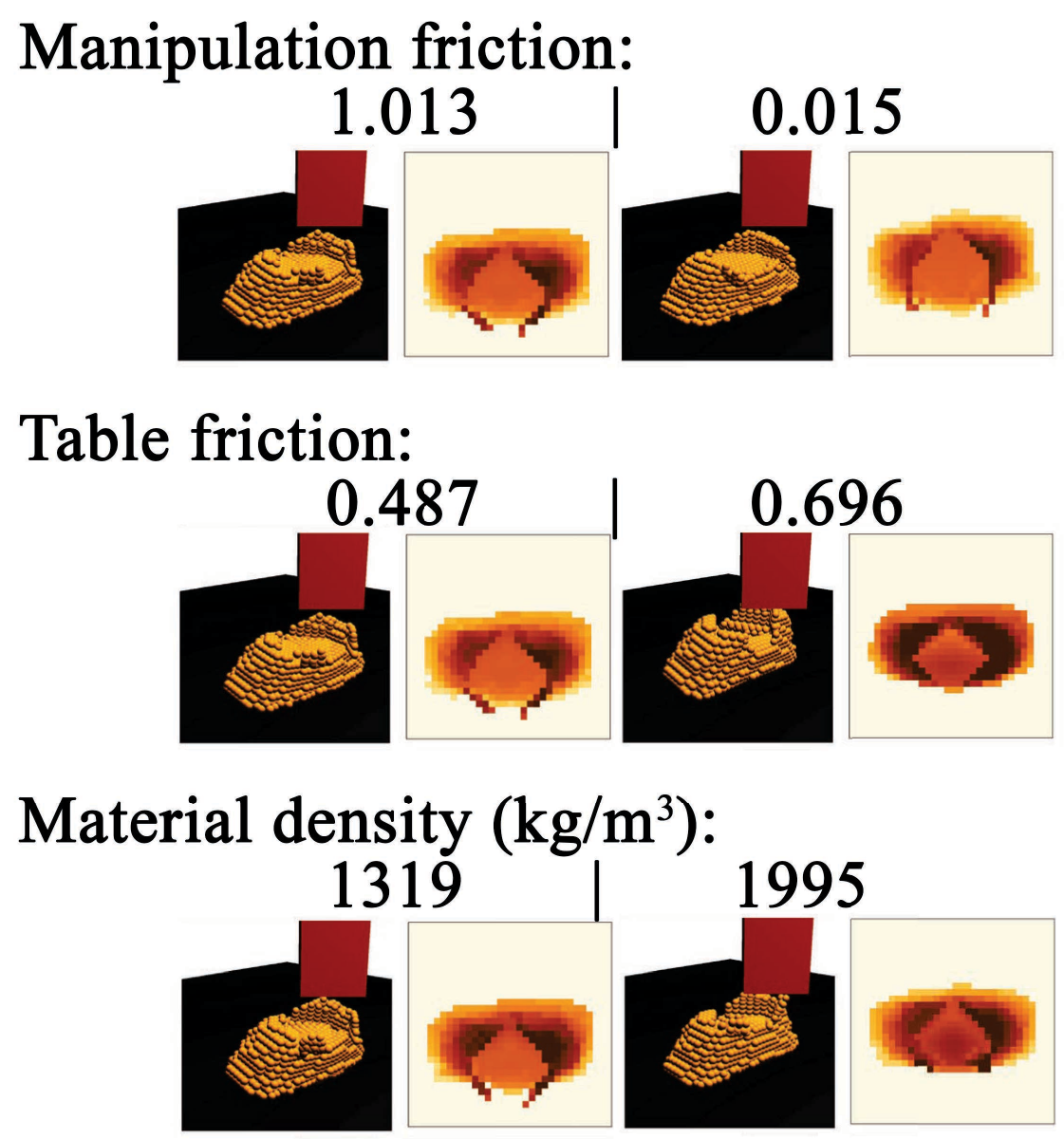}
    \caption{Selected examples from Figure~\ref{subfig:in-dist-vis-lv2-obj2} to show the physically realistic associations between object movements and three physics parameters, demonstrating the physical plausibility of different physics parameter values.}
    \label{fig:example-2}
\end{figure}

This may be caused by the fact that these losses distribute differently along the yield stress axis (the CD and EMD losses at level 1, the PRT CD loss at level 2). According to physics intuitions, a smaller yield stress will cause the object to yield more easily and respond more drastically to the poking forces, hence the more spreading in the $x$ and $y$ directions and more compressing in the $z$ direction, and vice versa. One can see that a greater yield stress value in Table~\ref{tab:best-param} corresponds to a darker heightmap in Figure~\ref{fig:in-dist-vis}, and vice versa.


Thirdly, though the particles and heightmaps are very similar, they are produced by quite different parameter combinations. From the results \xt{at contact level 2 in Subfigures~\ref{subfig:in-dist-vis-lv2-obj1} and~\ref{subfig:in-dist-vis-lv2-obj2}, one can observe a correlation} between the friction coefficients and the material density -- the three key parameters that determine how much the object will be moved in the shifting motion direction. The parameters in Table~\ref{tab:best-param} and their corresponding visualisations exhibit the following physically plausible relationships (see Figure~\ref{fig:example-2} for a few examples), 
\begin{enumerate}
    \item with similar weights ($\rho$), the object is moved at a longer distance when the manipulator frictions ($\mu_m$) is greater (compare the parameters and visualisations between the results from the PCD EMD loss with the 12-mix and 1-rec. datasets)
    \item with similar manipulator frictions ($\mu_m$), the object is moved further when the table friction ($\mu_t$) is low (compare the parameters and visualisations between the PCD CD loss with the 1-round dataset and the PRT EMD loss with the 12-mix dataset)
    \item with similar friction values, the object is moved further when it is lighter (compare the results from the PCD EMD loss with the 12-mix and the 1-cyl. datasets)
\end{enumerate}

These relationships between parameters and visualisations align with our understanding of real-world physics, demonstrating that DPSI can produce physically plausible and interpretable parameter values. \xt{Moreover, the variations in parameter values suggest that, aside from the mismatch of the models and the materials, there is hardly any fixed set of parameter values that can simulate a material in every condition. These parameters will always have some differences in different conditions (e.g., heat, light, humidity, etc.) and they change over time (e.g., drying). Therefore, the fact that DPSI can estimate the parameters with only one interaction makes it highly promising for fast parameter re-identification that can be performed online to recalibrate simulation accuracy efficiently, eliminating the need to keep track of previously identified parameter values.}

Fourthly, the visualisations in Figure~\ref{fig:in-dist-vis} also reveal limitations and potential improvement directions of the physics model. By comparing with the ground truths, the simulated contact areas of the objects always deform more sharply with insufficient elastic returning. The real objects, on the other hand, tend to respond more elastically after being plastically deformed, hence the higher and smoother surfaces shown in the ground truths. This is more obvious for contact areas with sharp edges, such as the four edges of the rectangle end-effector or the sharp sides of the cylinder end-effector. In addition, at level-2 cases, with the effectors shifting horizontally, some simulated particles are unrealistically displaced and stay floating. In the real world, the displaced parts would fall because of gravity. What's more, the shifting motion in the real world causes the whole object to tilt in the moving direction, while in the simulation the contact impact tends to remain in a much smaller region around the contact area. These inconsistencies can be caused not only by inaccurate modelling, but also by various computation approximations, such as time integration method, simulation step size, contact handling, etc.

\noindent\textbf{\textit{Summary}:} In short, this subsection presents a detailed examination of the in-distribution performances of DPSI with motions at two levels of contact complexity. By analysing the validation loss curves, the parameter values and their manipulation visualisations, we can answer the proposed questions:
\begin{itemize}
    \item The loss functions can be minimised, demonstrating the feasibility of DPSI even in the presence of noisy and incomplete point cloud observations.
    \item The loss functions produce similar visualisations, but do not always agree with each other on the found parameter values. The CD and EMD losses seem to focus on quite different spatial aspects of the point sets. They distribute differently along the yield stress ($\pmb{\sigma}_y$) axis and produce different yield stress values. However, it seems that, as the optimisation motion becomes more complex, the loss functions agree more on the parameter solutions.
    \item There are many local minima and possible parameter solutions, but they produce visually and physically similar manipulation results with physically intuitive interpretability.
    \item The number of datapoints has a minor influence on the optimisation performances both quantitatively and qualitatively, indicating that DPSI is not data-hungry even in the presence of real-world perception challenges.
    \item Discussions on the visualisations and parameter values show that DPSI can produce physics parameter values with physically realistic and interpretations. 
\end{itemize}

\subsection{Generalisation}
This subsection will look at the out-of-distribution performance of DPSI by visualising the manipulation processes of three unseen and much longer motions that induce more complex contact dynamics with the best parameters (Table~\ref{tab:best-param}) found at the level-2 contact complexity identification task.

\begin{figure}[t]
\footnotesize
\centering
\begin{subfigure}{\columnwidth}
\vspace{-6mm}
\caption{Triple-poking, object configuration 1}
\vspace{-1mm}
\includegraphics[width=0.905\linewidth]{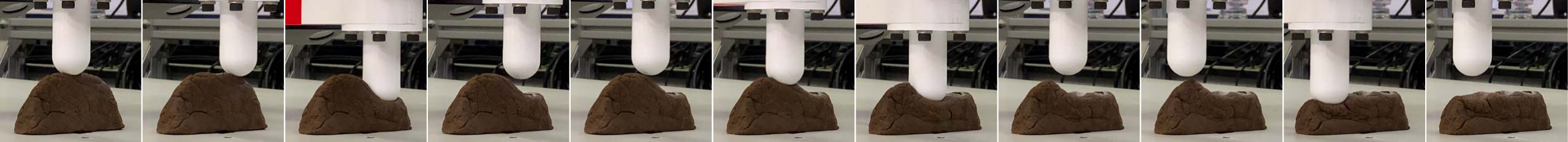}
\includegraphics[width=0.085\linewidth]{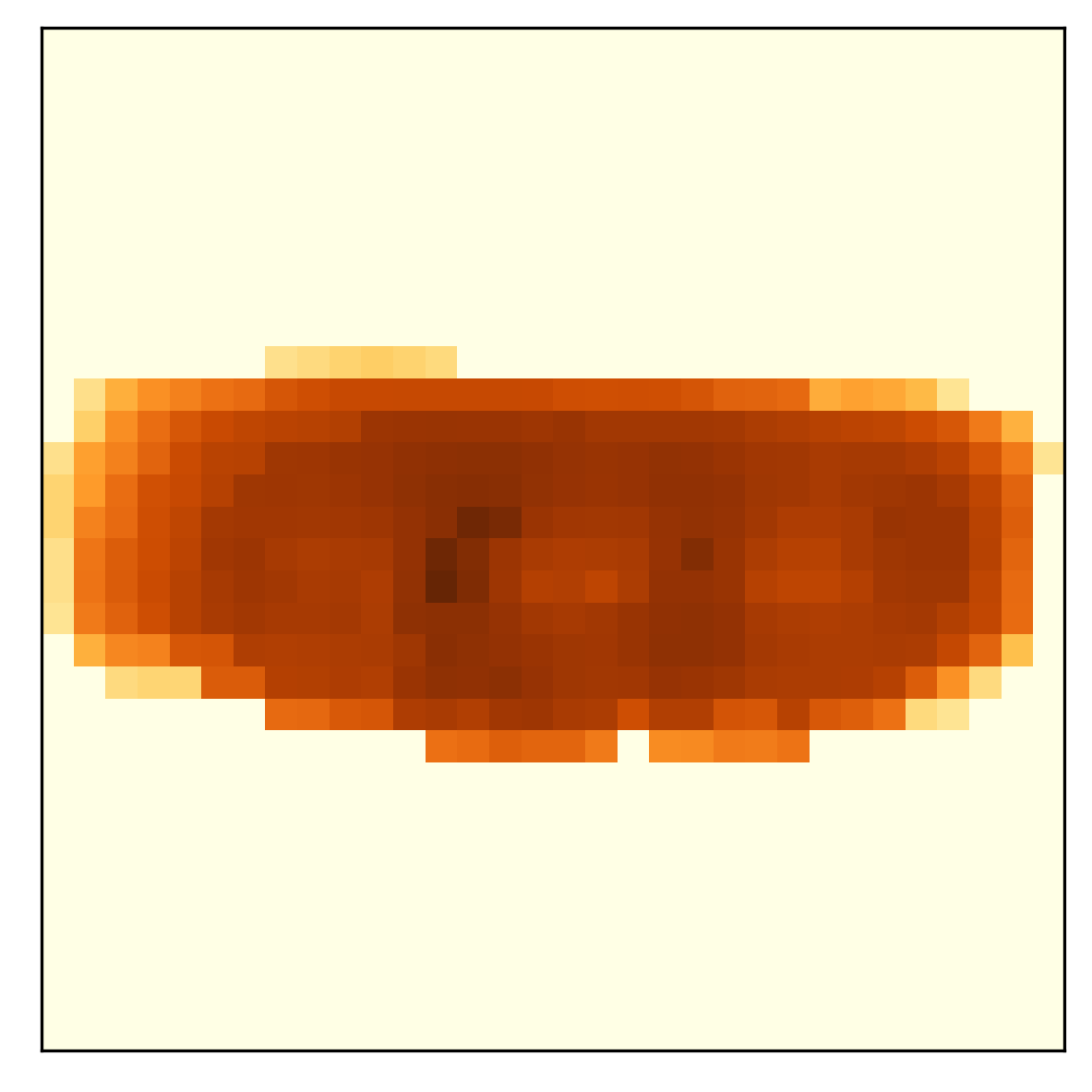}\\
\includegraphics[width=0.905\linewidth]{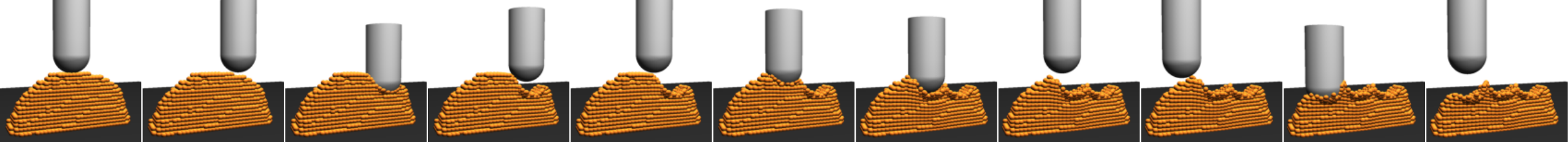}
\includegraphics[width=0.085\linewidth]{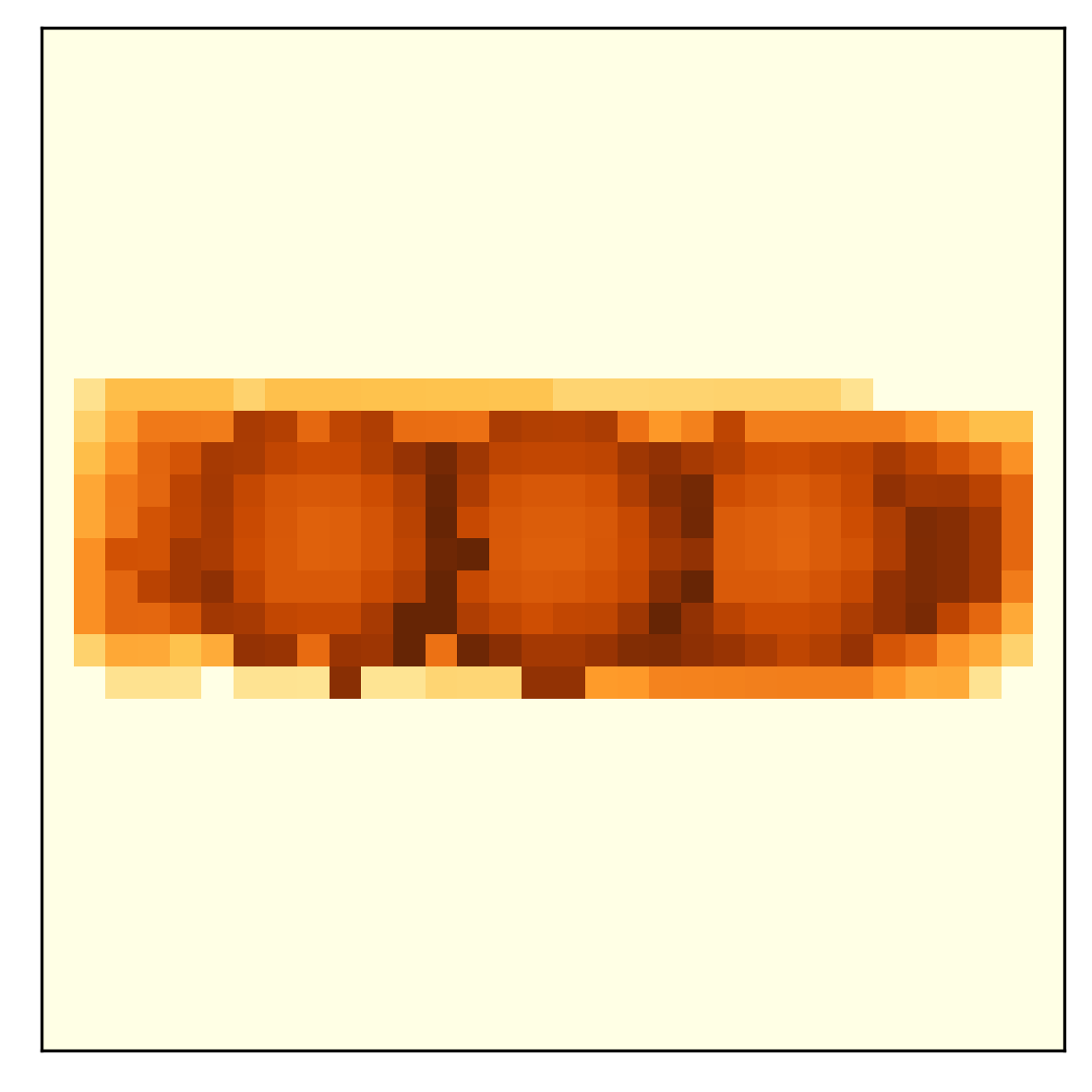}\\
\includegraphics[width=0.905\linewidth]{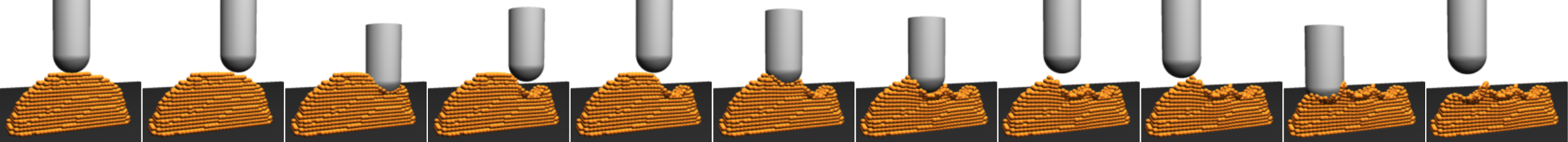}
\includegraphics[width=0.085\linewidth]{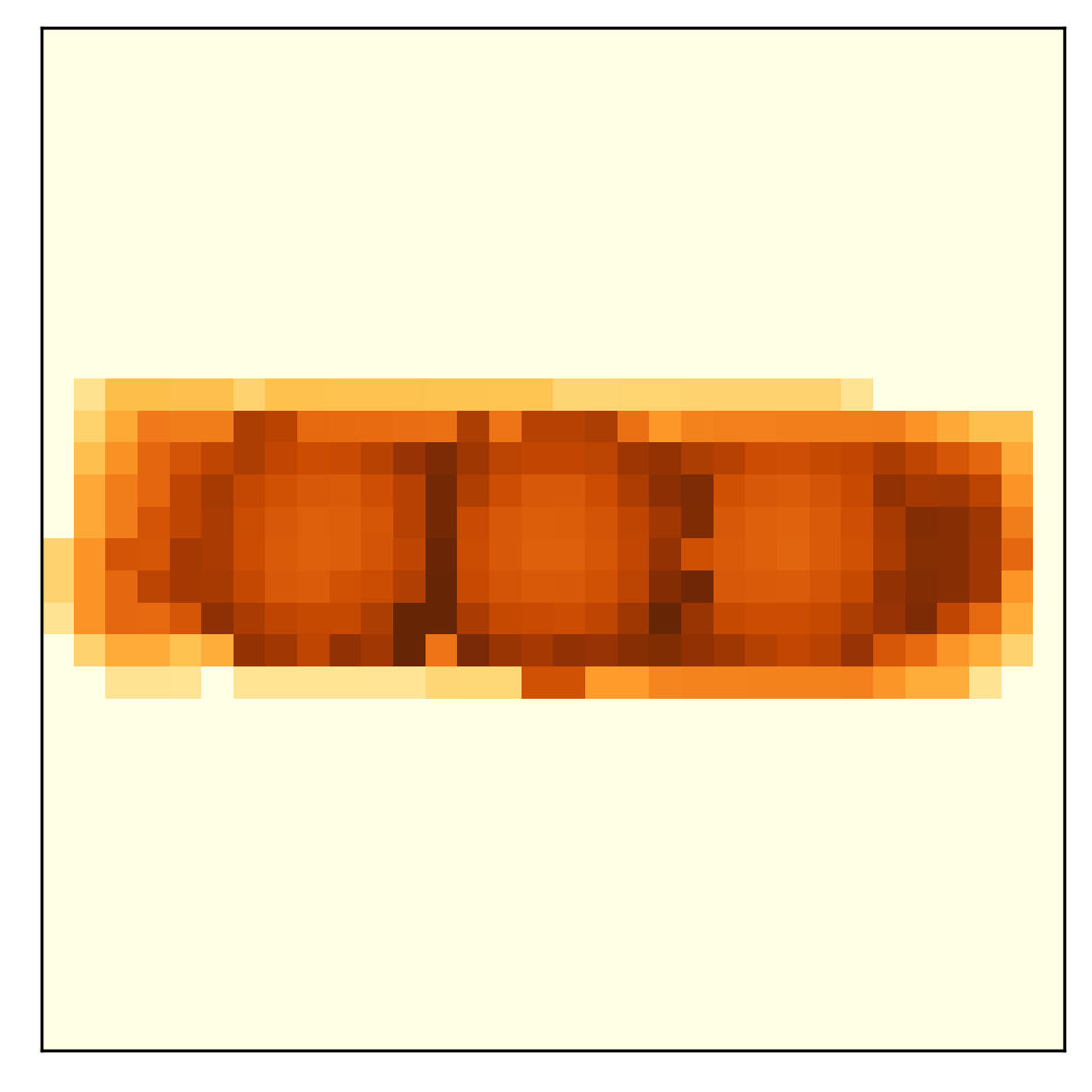}\\
\includegraphics[width=0.905\linewidth]{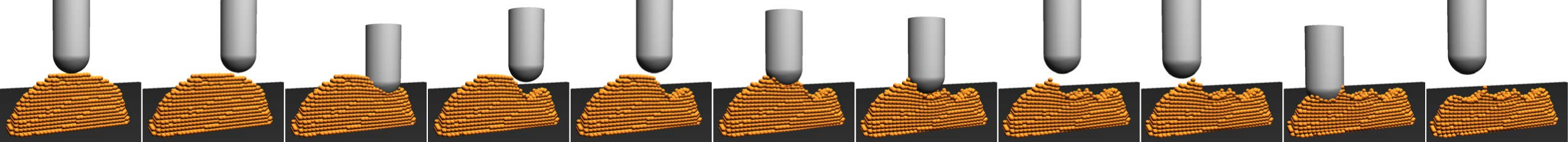}    
\includegraphics[width=0.085\linewidth]{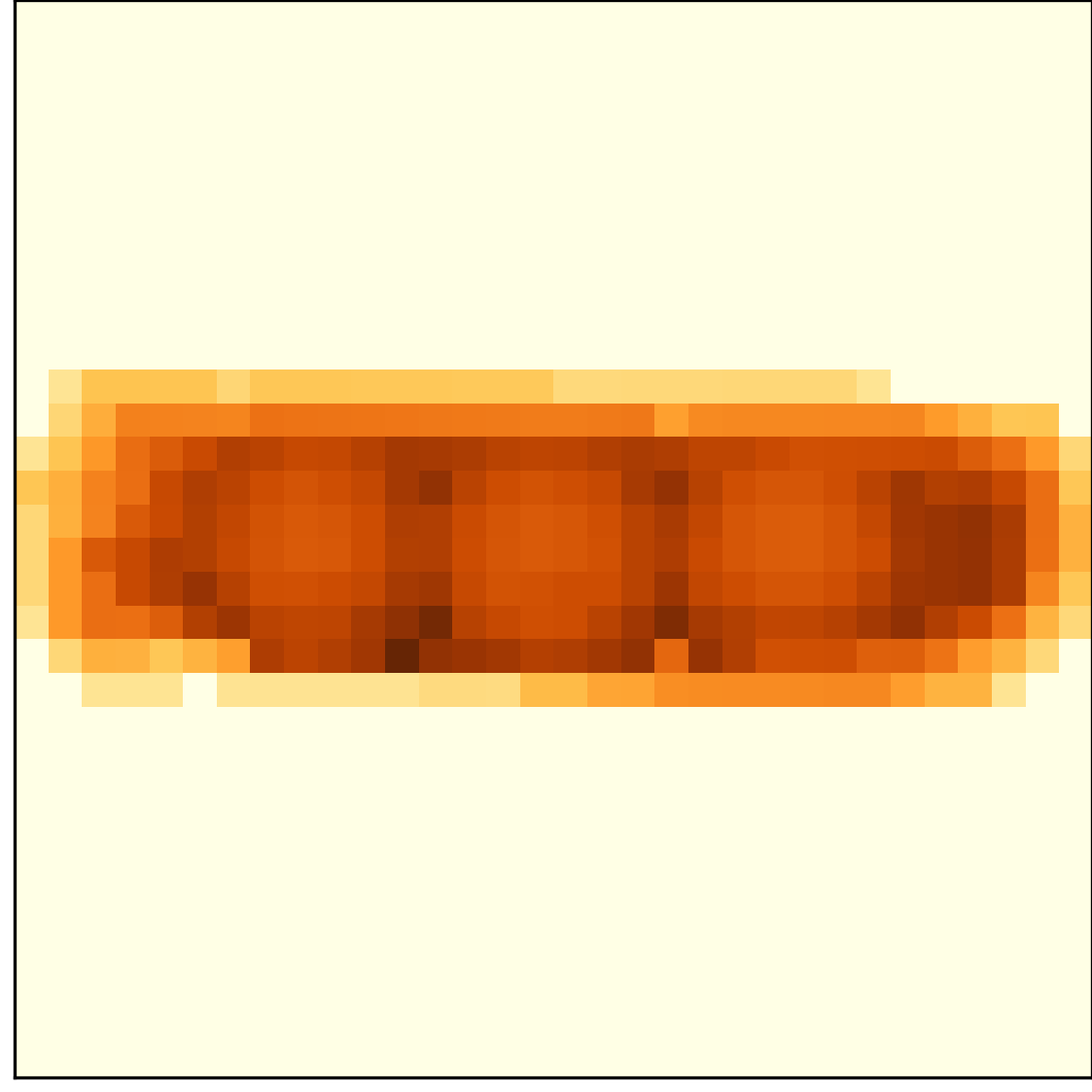}
\vspace{-6mm}
\end{subfigure}

\begin{subfigure}{\columnwidth}
\caption{Triple-poking, object configuration 2}
\vspace{-1mm}
\includegraphics[width=0.905\linewidth]{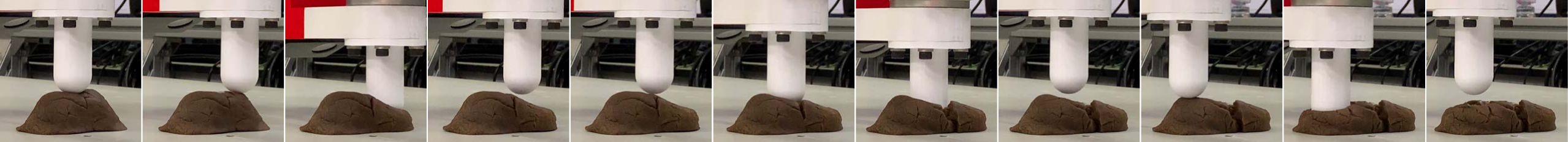}
\includegraphics[width=0.085\linewidth]{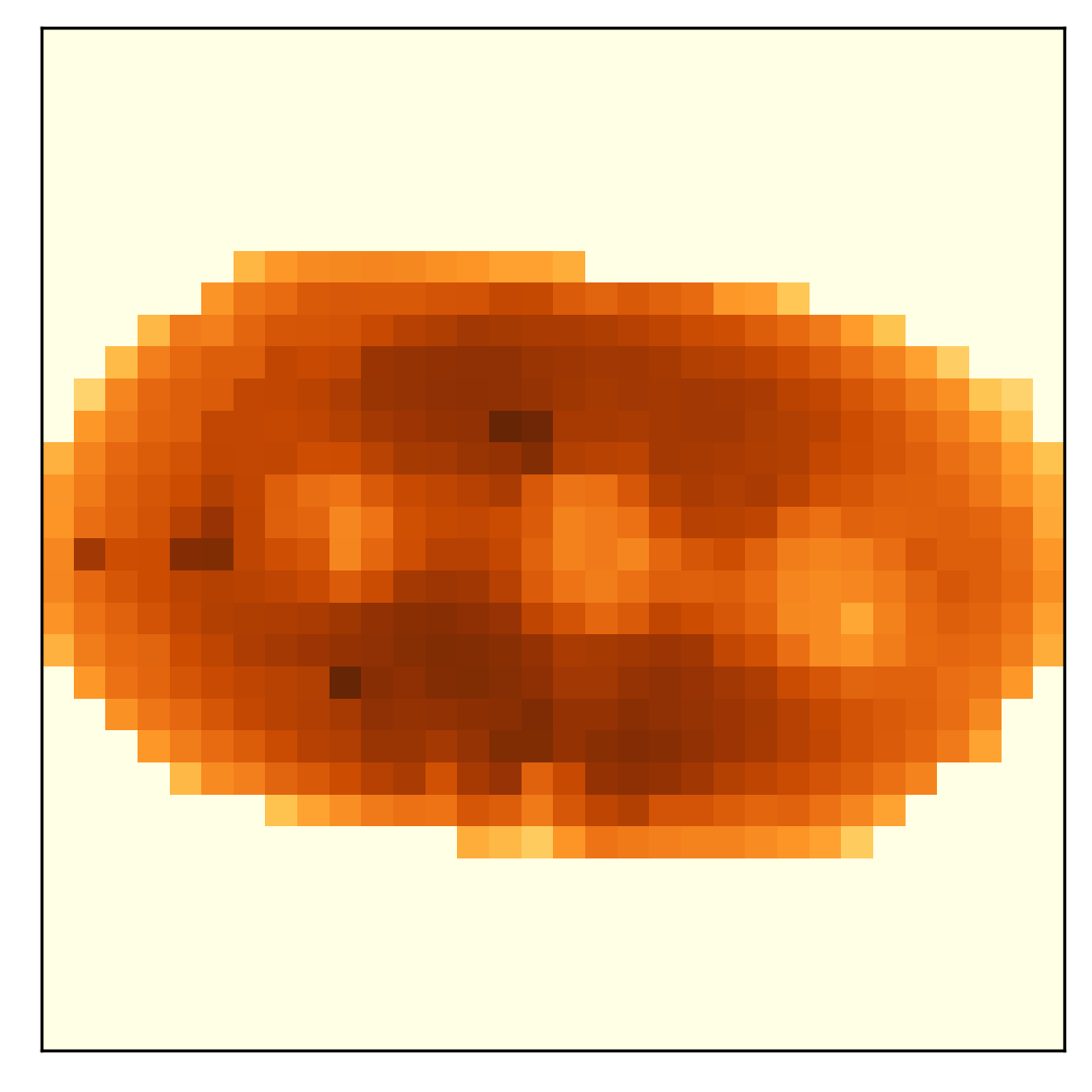}\\
\includegraphics[width=0.905\linewidth]{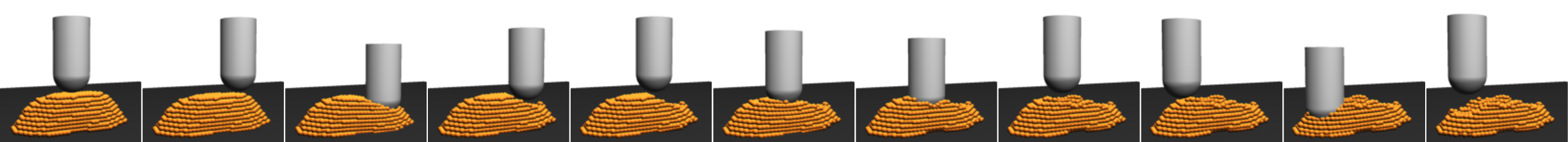}
\includegraphics[width=0.085\linewidth]{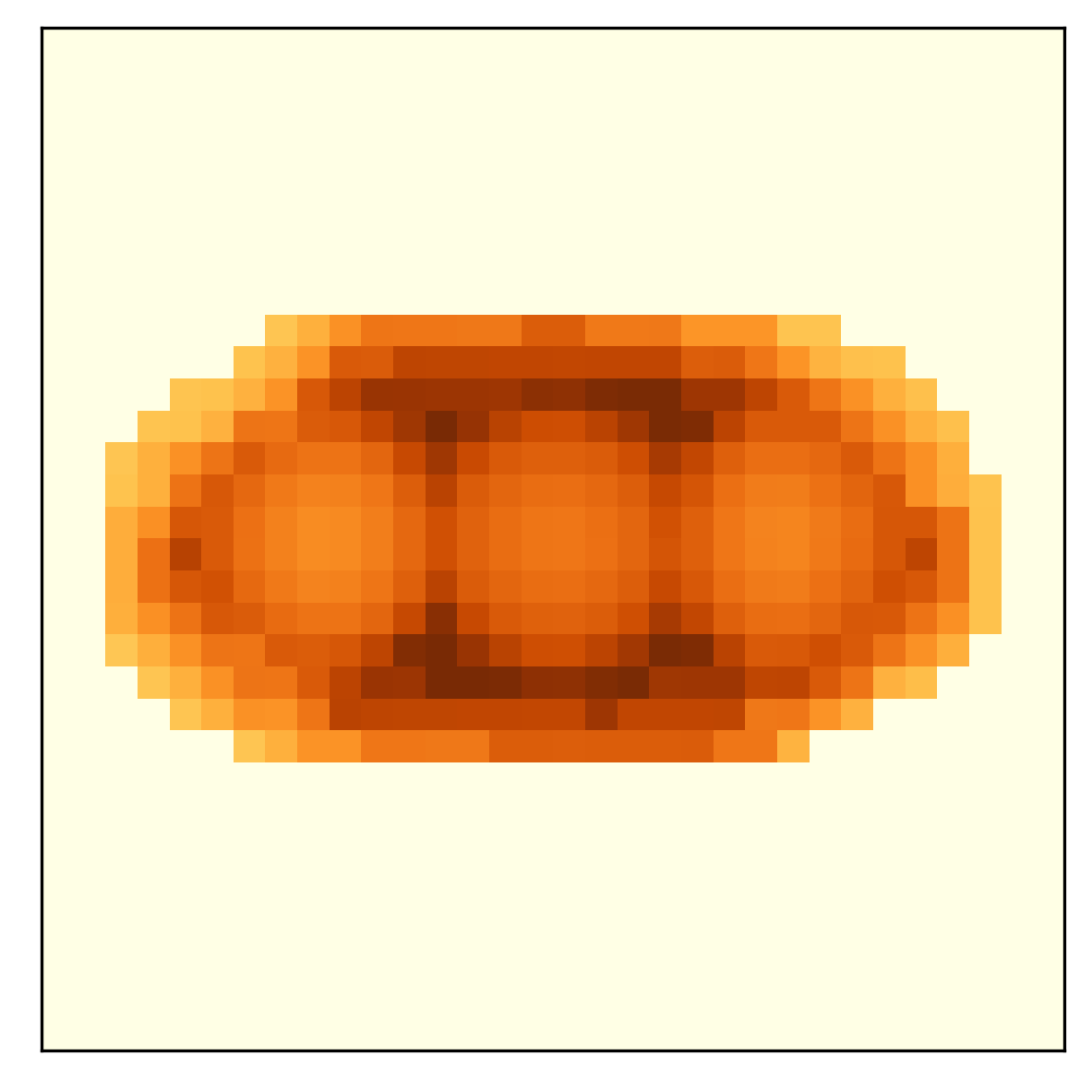}\\
\includegraphics[width=0.905\linewidth]{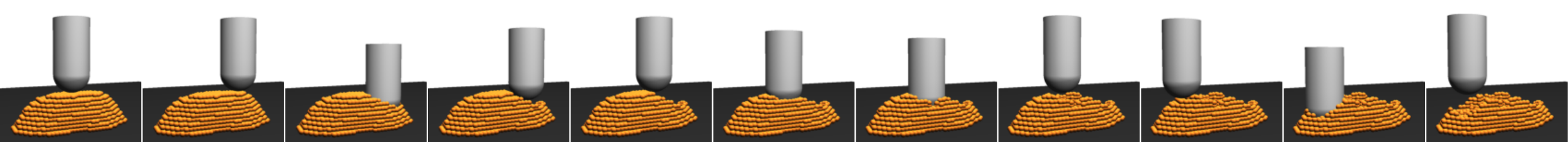}
\includegraphics[width=0.085\linewidth]{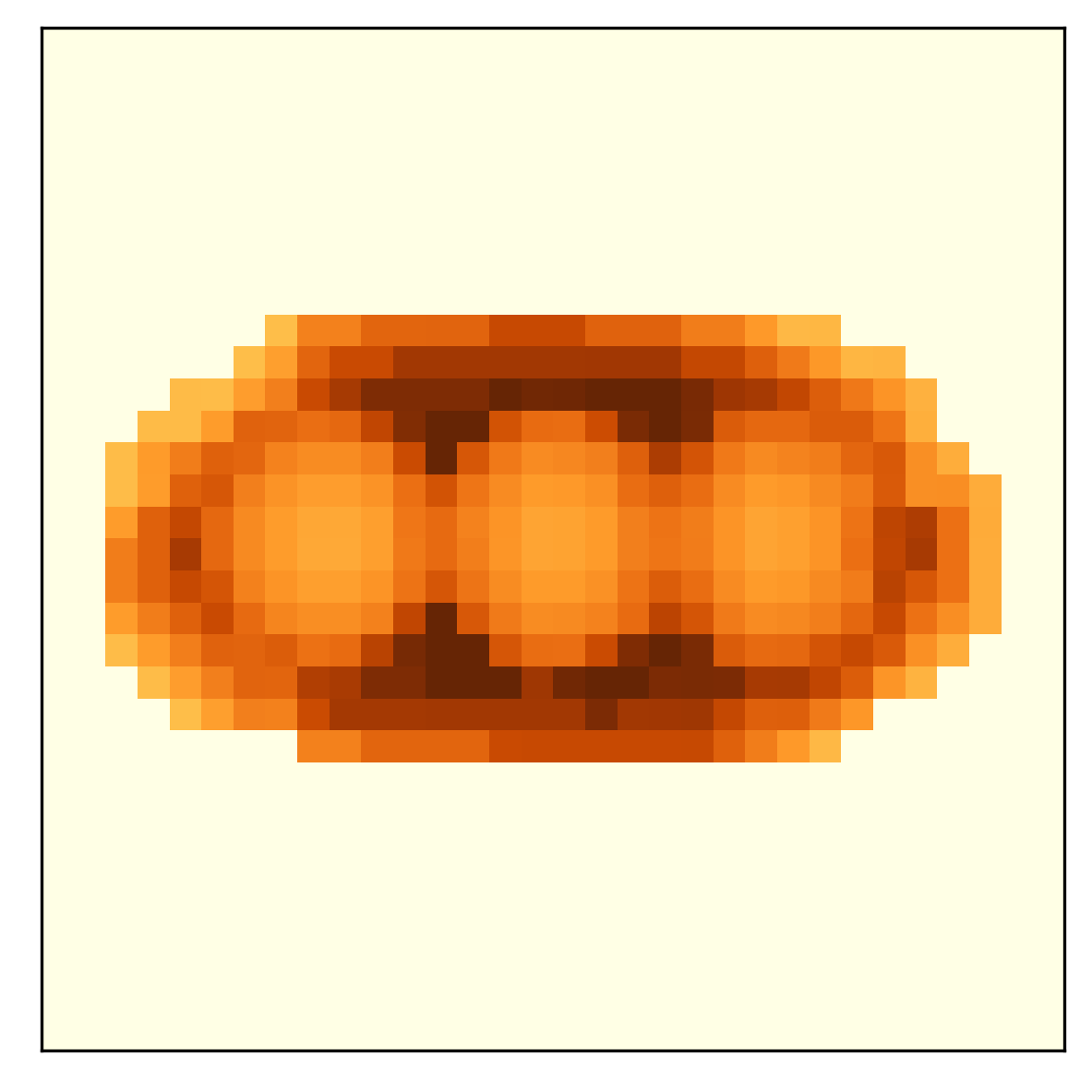}\\
\includegraphics[width=0.905\linewidth]{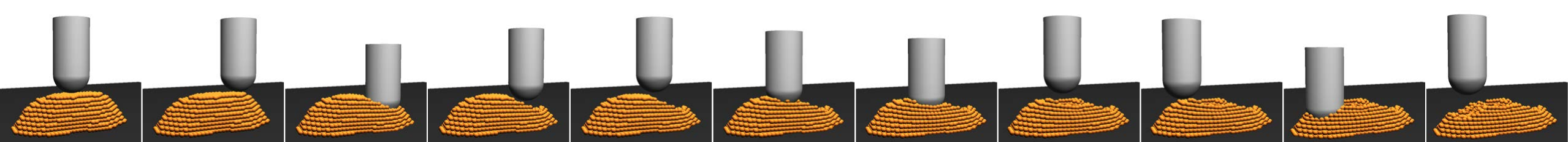}    
\includegraphics[width=0.085\linewidth]{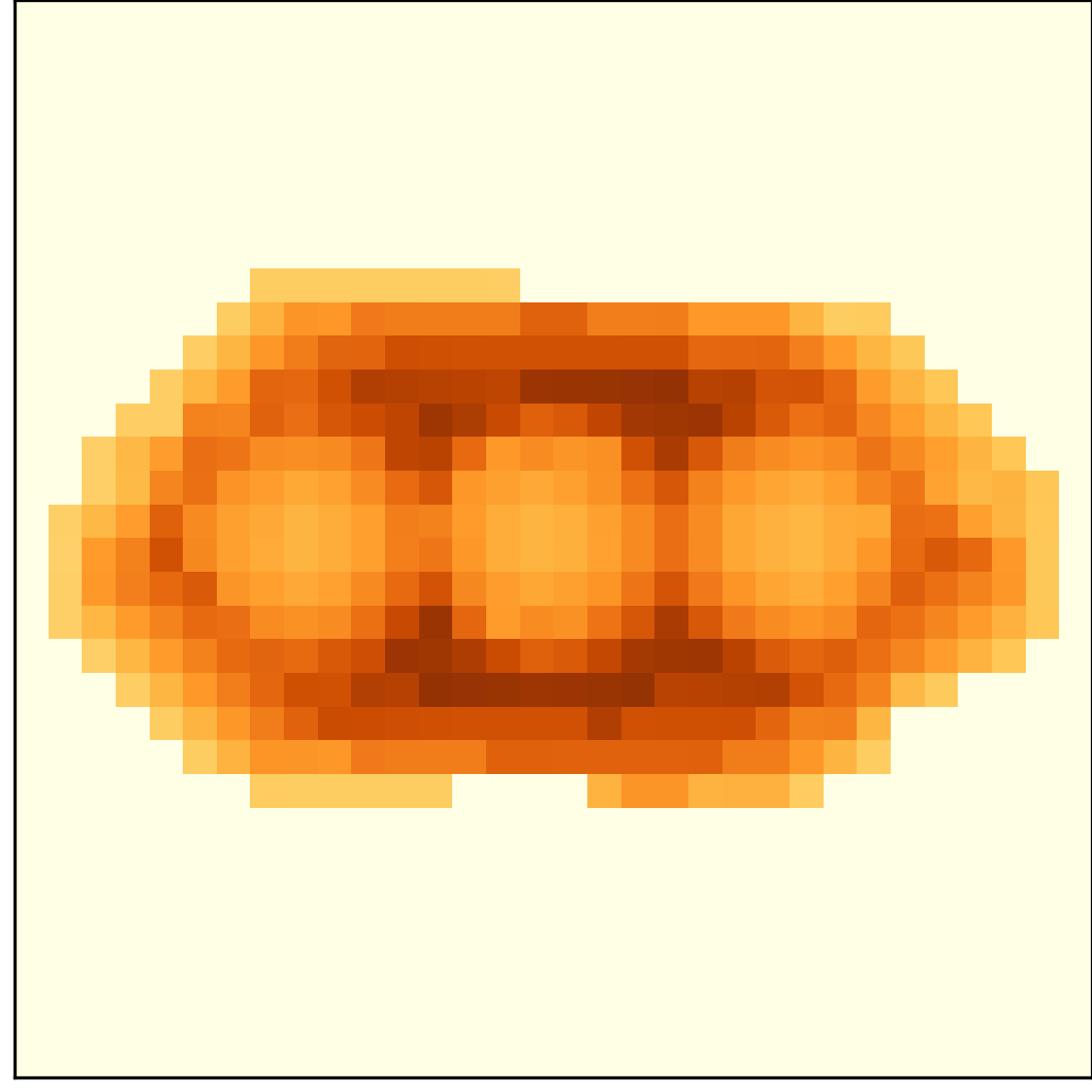}
\vspace{-6mm}
\end{subfigure}

\begin{subfigure}{\columnwidth}
\caption{Flattening, object configuration 1}
\vspace{-1mm}
\includegraphics[width=0.905\linewidth]{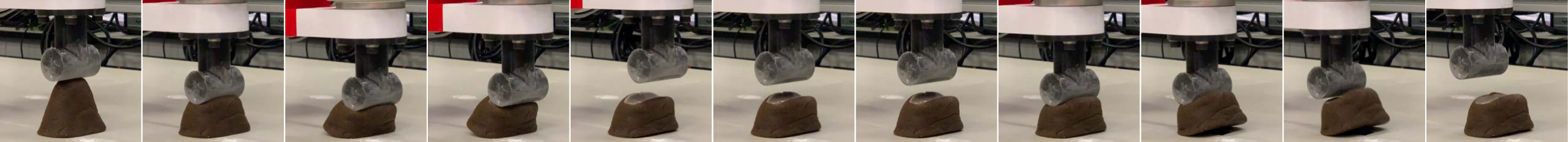}
\includegraphics[width=0.085\linewidth]{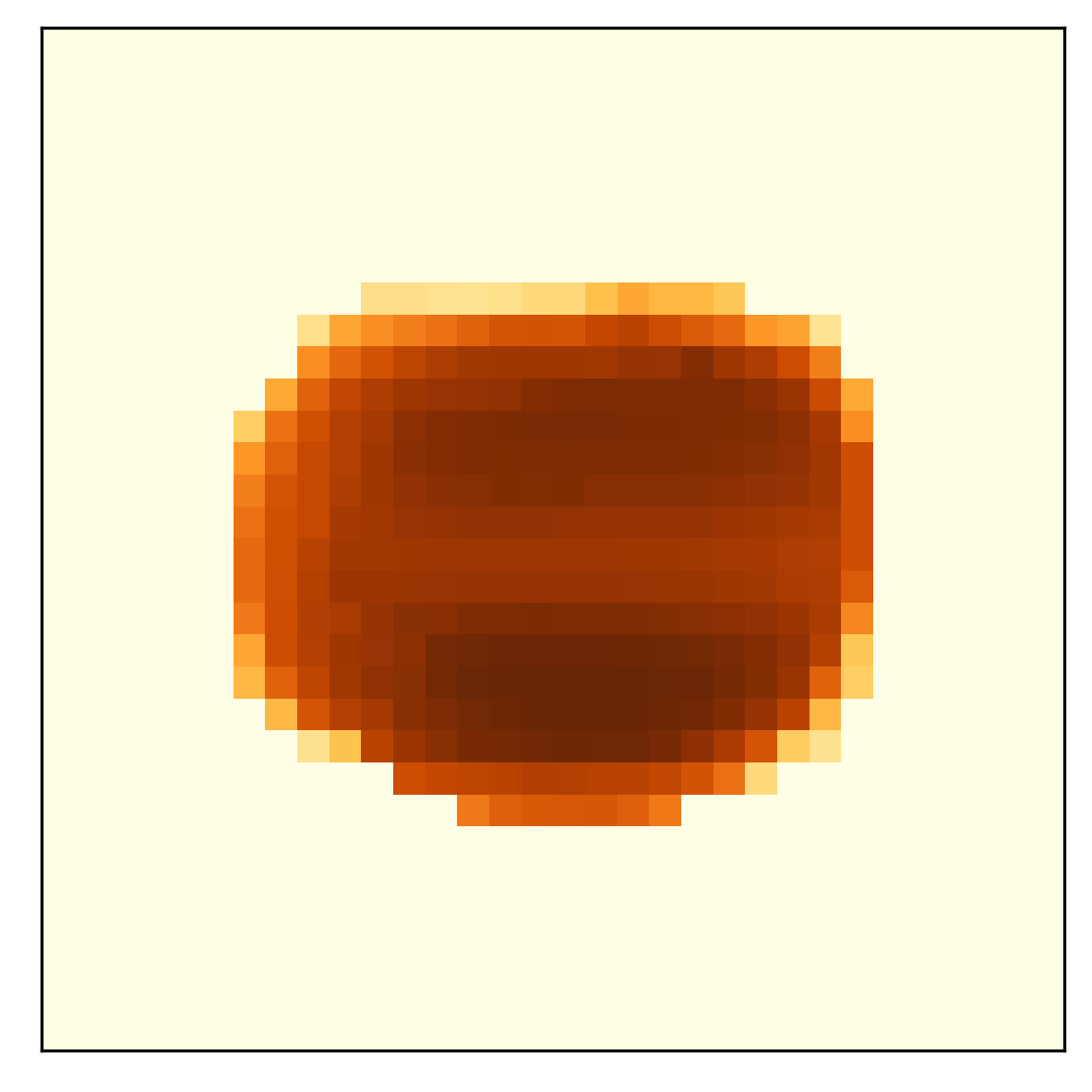}\\
\includegraphics[width=0.905\linewidth]{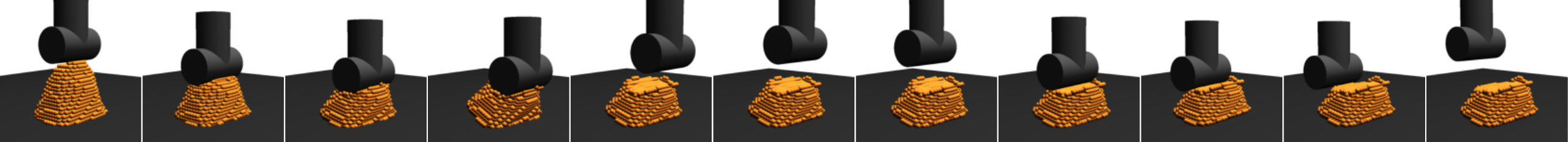}
\includegraphics[width=0.085\linewidth]{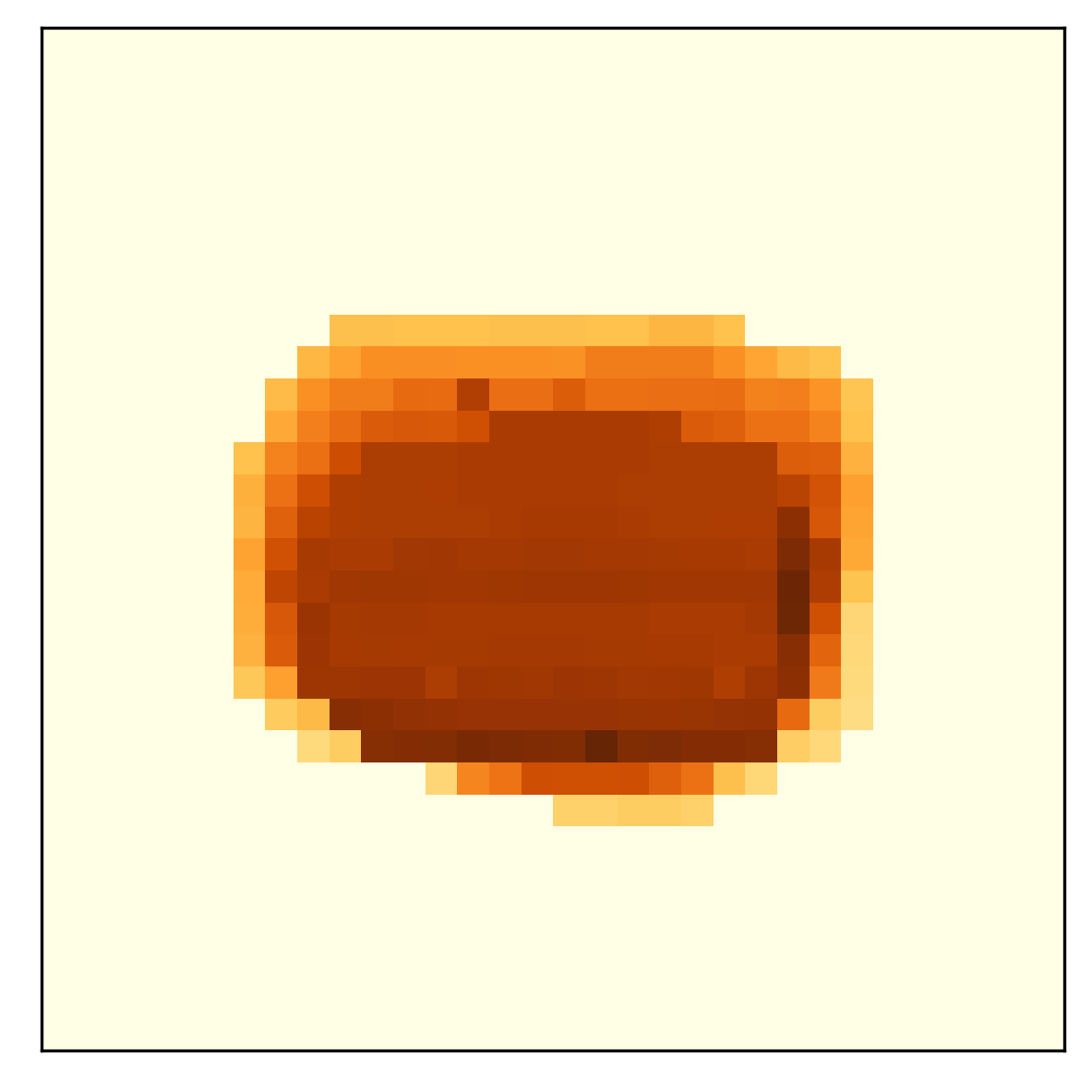}\\
\includegraphics[width=0.905\linewidth]{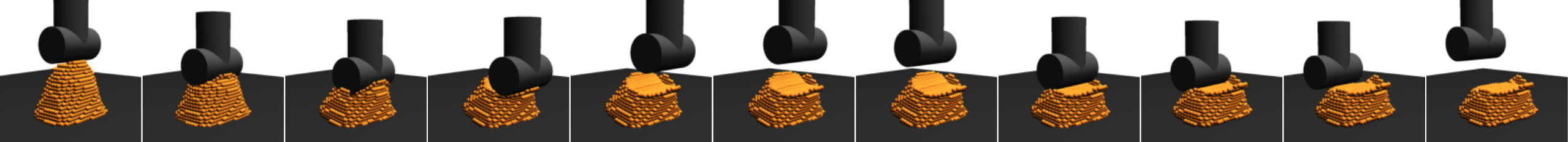}
\includegraphics[width=0.085\linewidth]{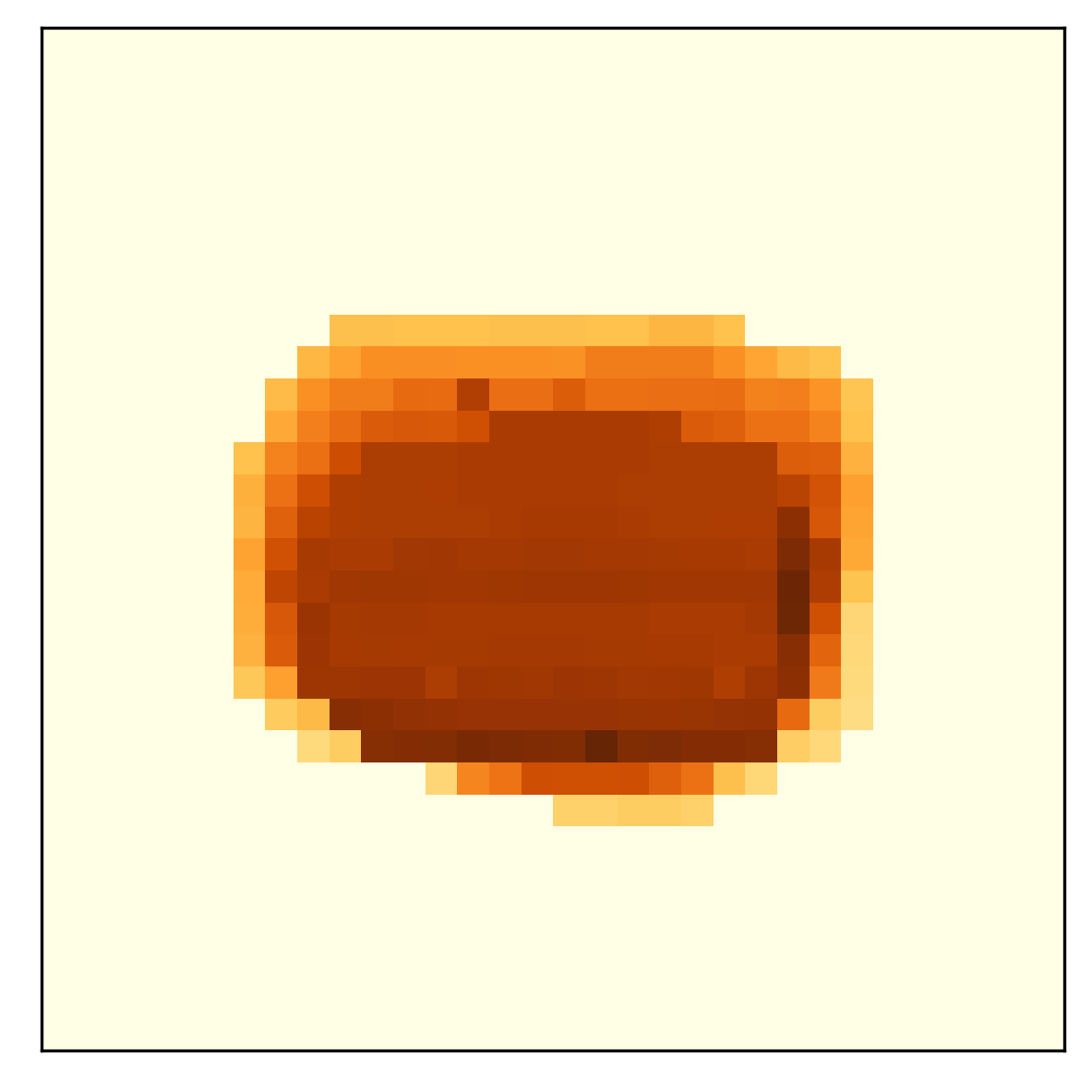}\\
\includegraphics[width=0.905\linewidth]{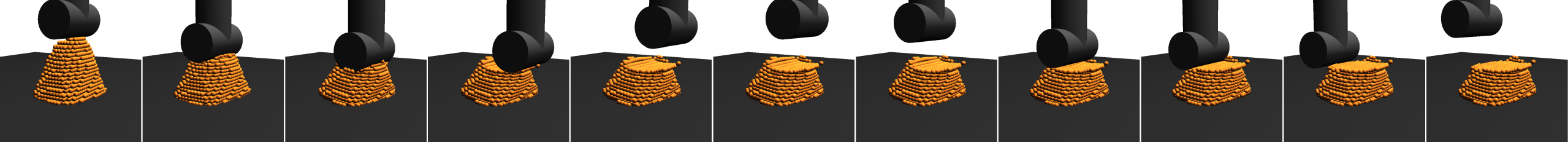}
\includegraphics[width=0.085\linewidth]{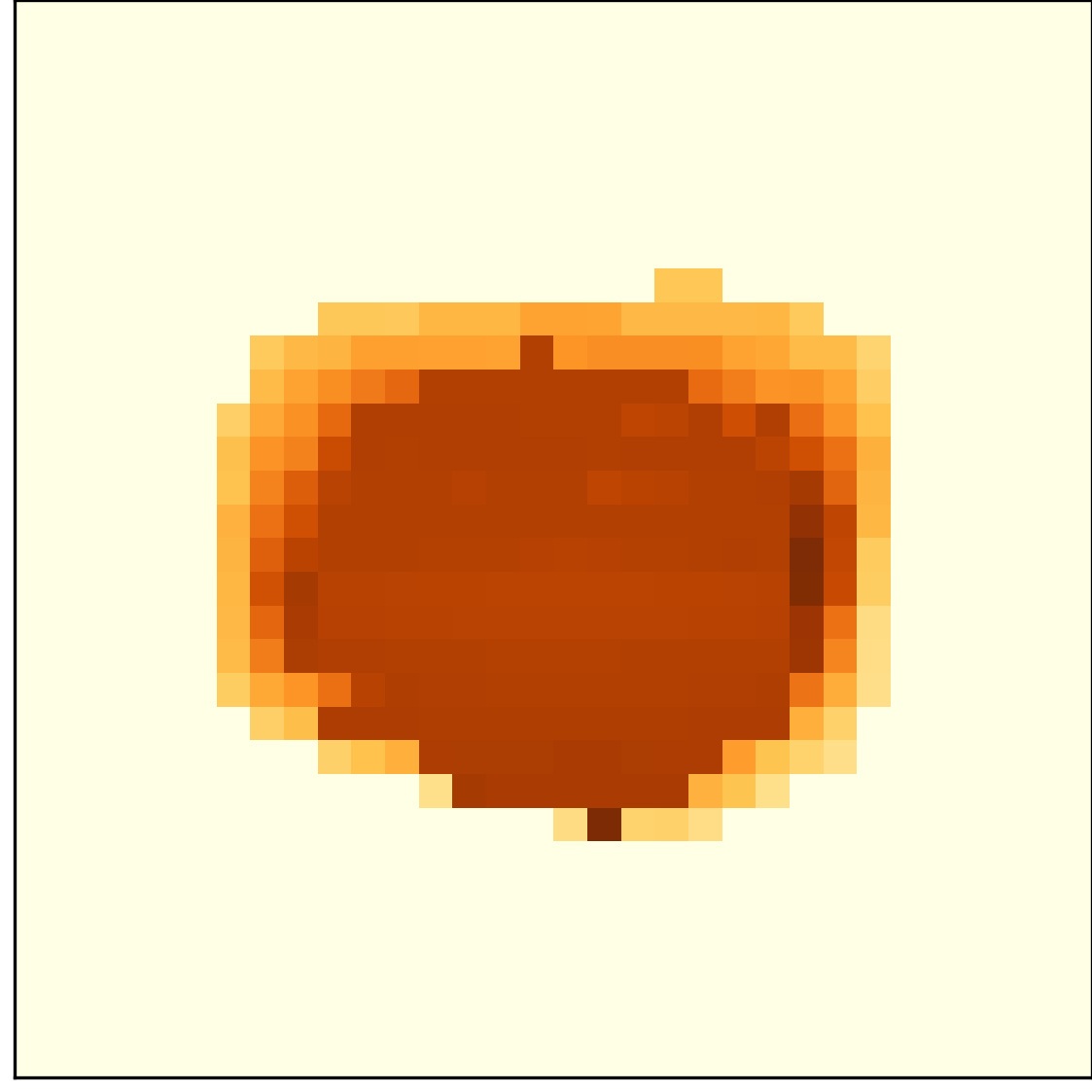}
\vspace{-6mm}
\end{subfigure}

\begin{subfigure}{\columnwidth}
\caption{Flattening, object configuration 2}
\vspace{-1mm}
\includegraphics[width=0.905\linewidth]{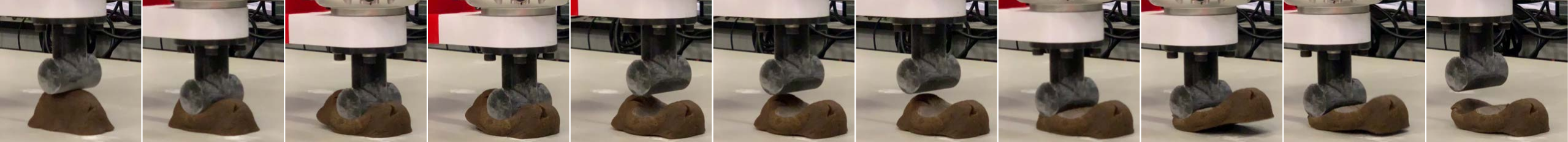}
\includegraphics[width=0.085\linewidth]{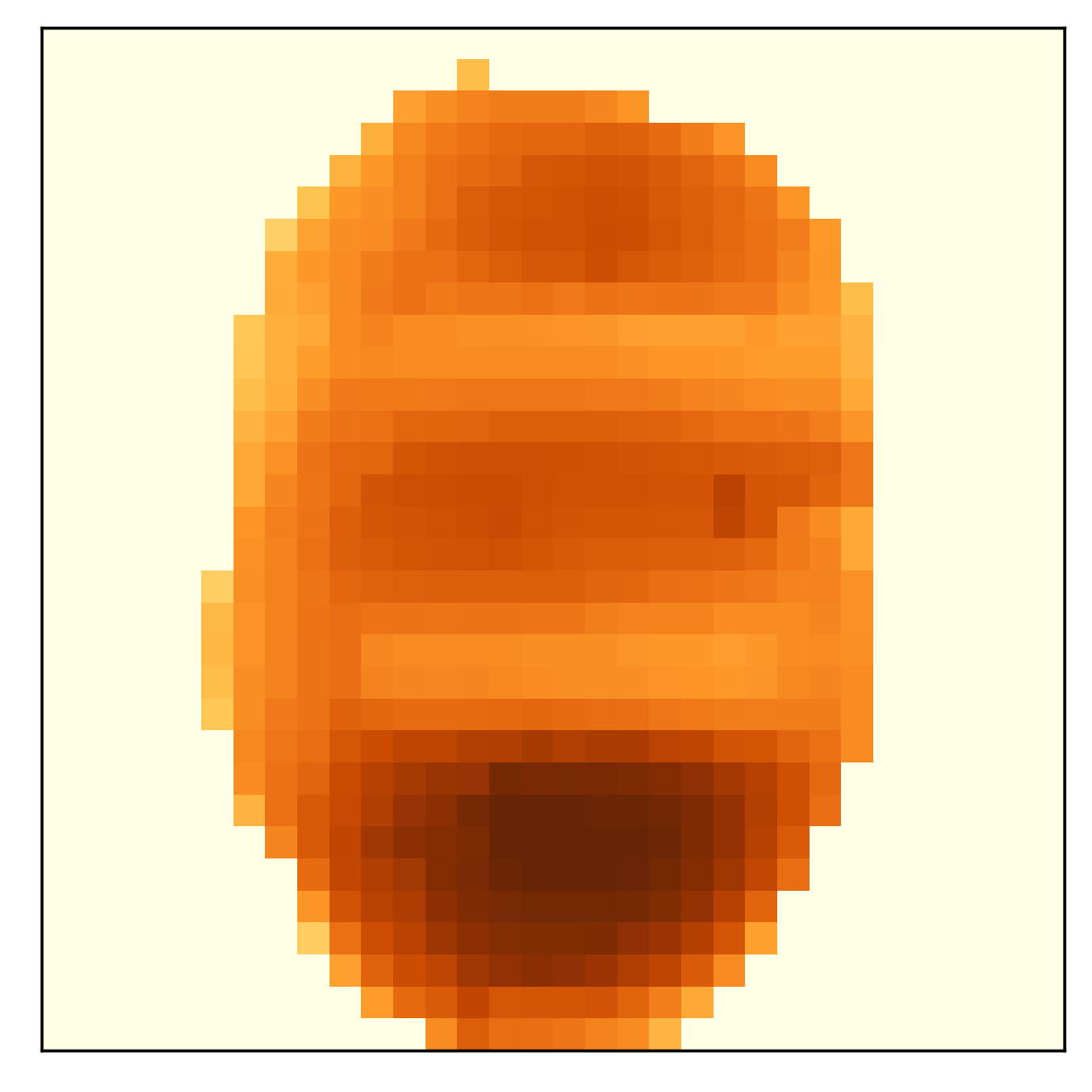}\\
\includegraphics[width=0.905\linewidth]{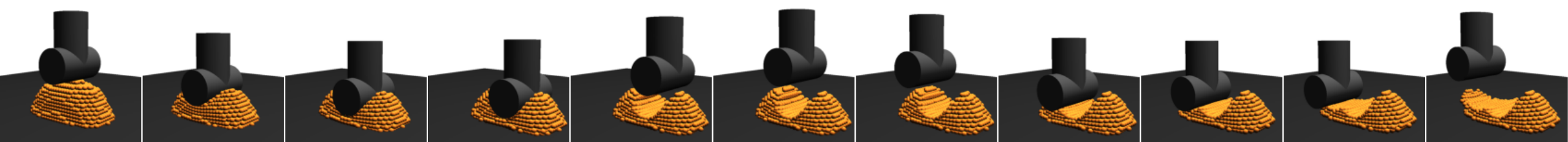}
\includegraphics[width=0.085\linewidth]{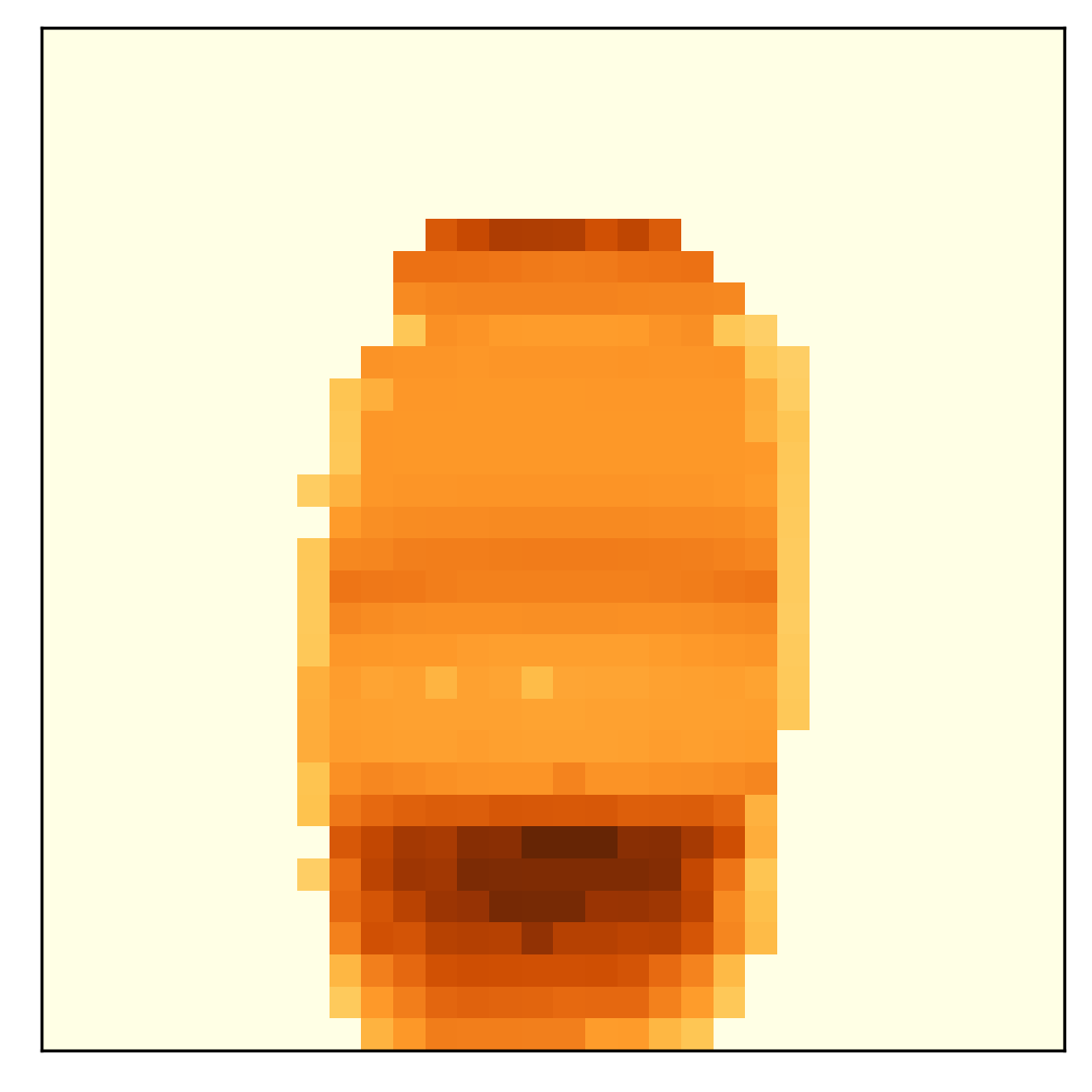}\\
\includegraphics[width=0.905\linewidth]{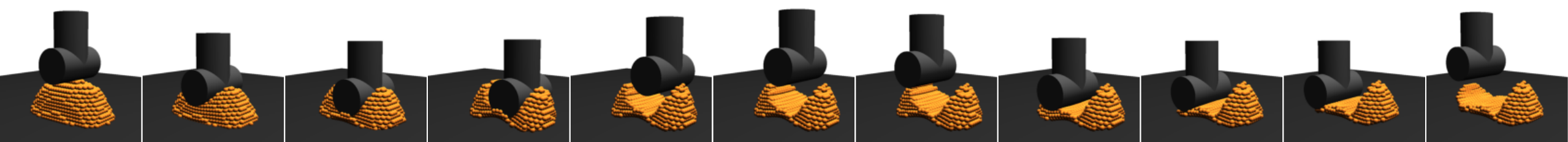}
\includegraphics[width=0.085\linewidth]{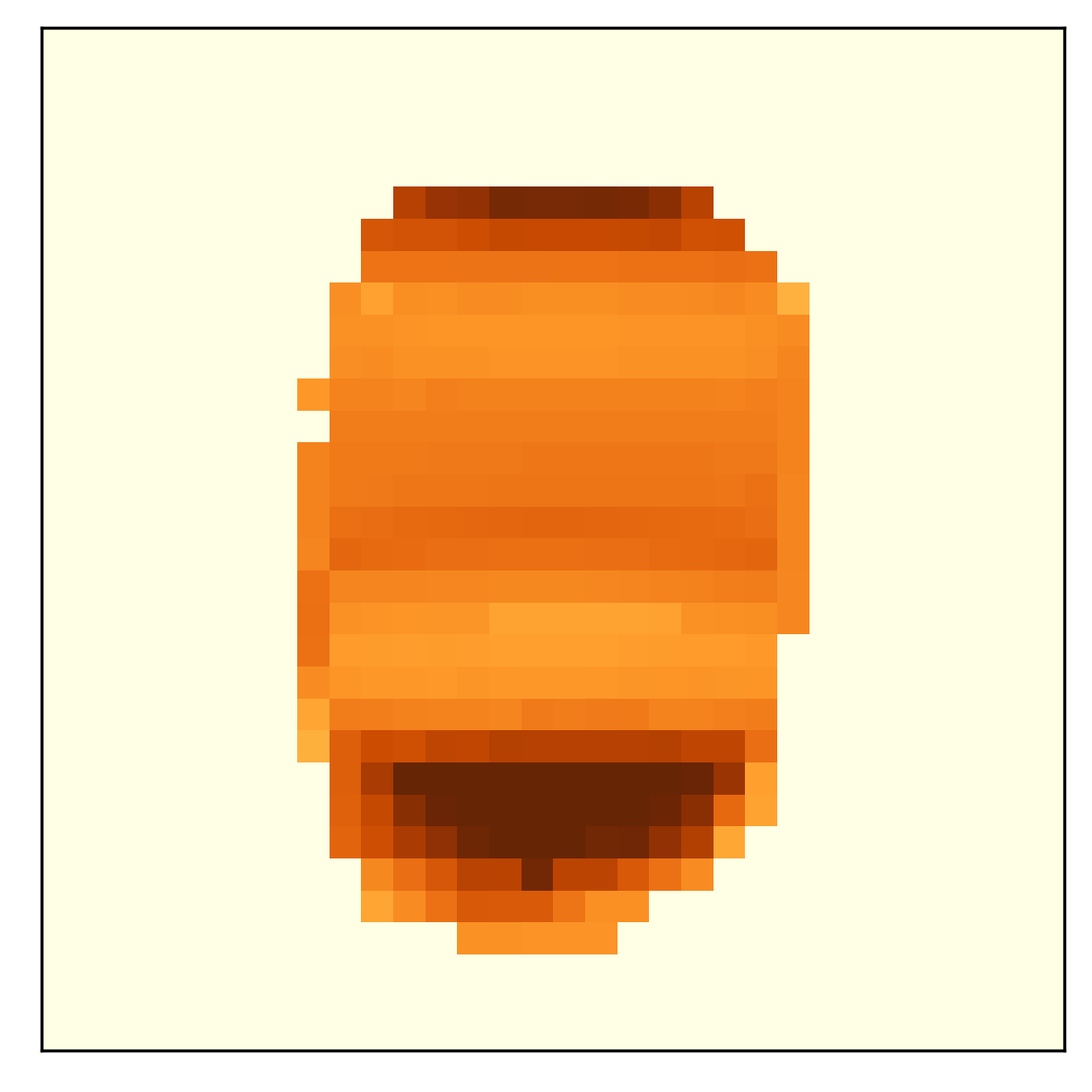}\\
\includegraphics[width=0.905\linewidth]{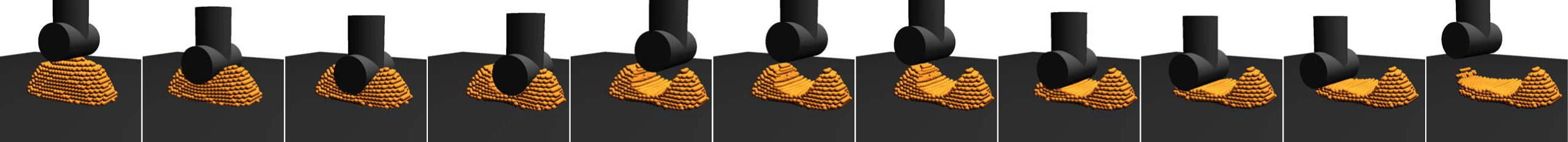}
\includegraphics[width=0.085\linewidth]{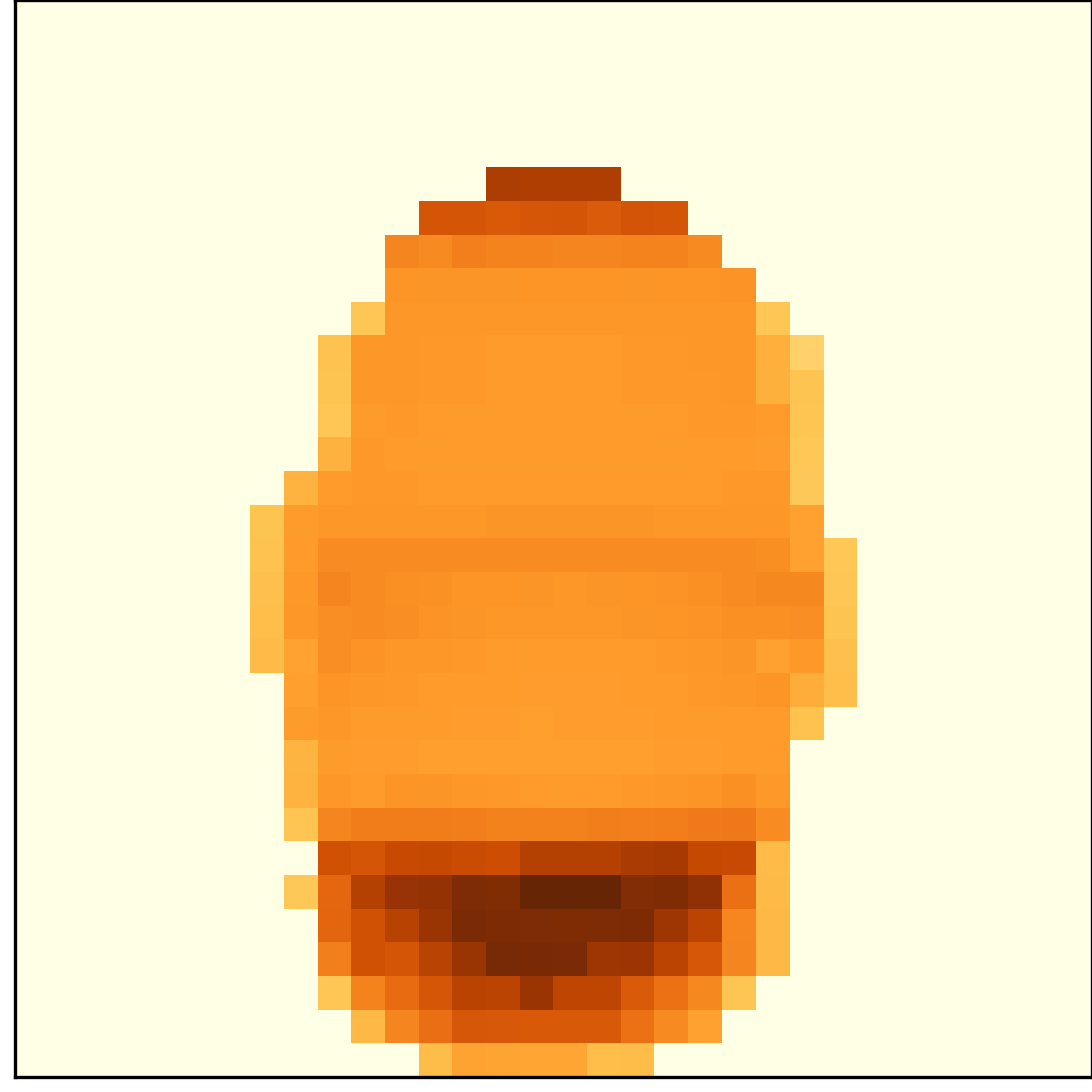}
\vspace{-6mm}
\end{subfigure}
\begin{subfigure}{\columnwidth}
\caption{Poking-180-rotating, object configuration 1}
\vspace{-1mm}
\includegraphics[width=0.905\linewidth]{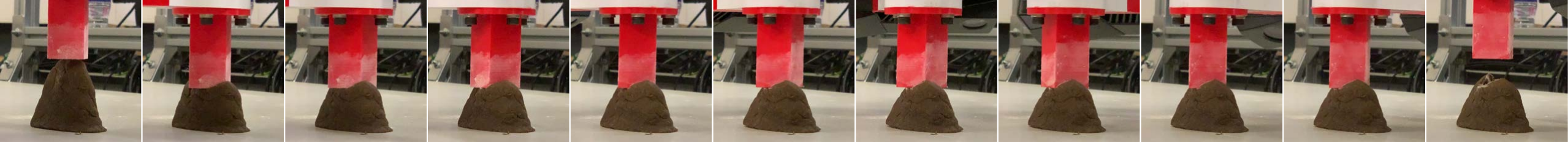}
\includegraphics[width=0.085\linewidth]{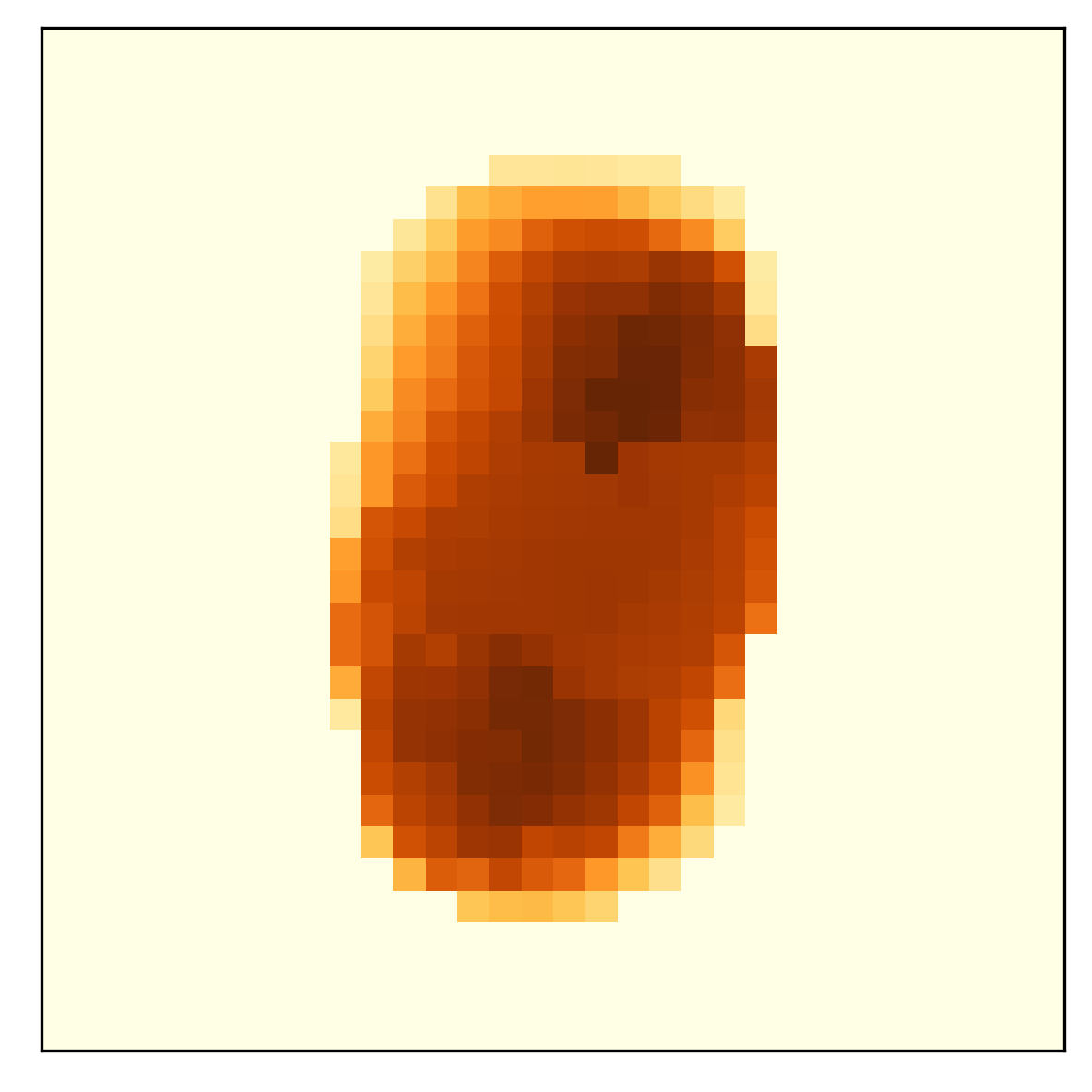}\\
\includegraphics[width=0.905\linewidth]{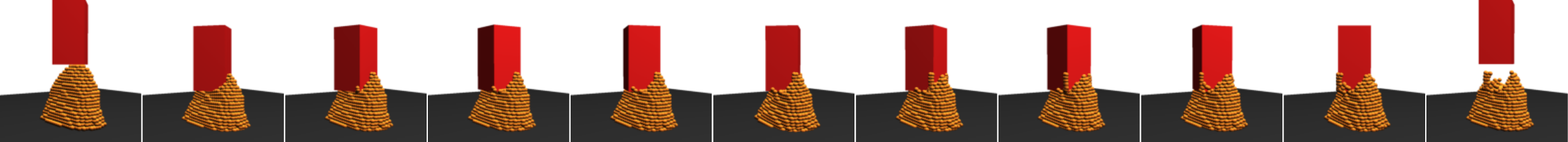}
\includegraphics[width=0.085\linewidth]{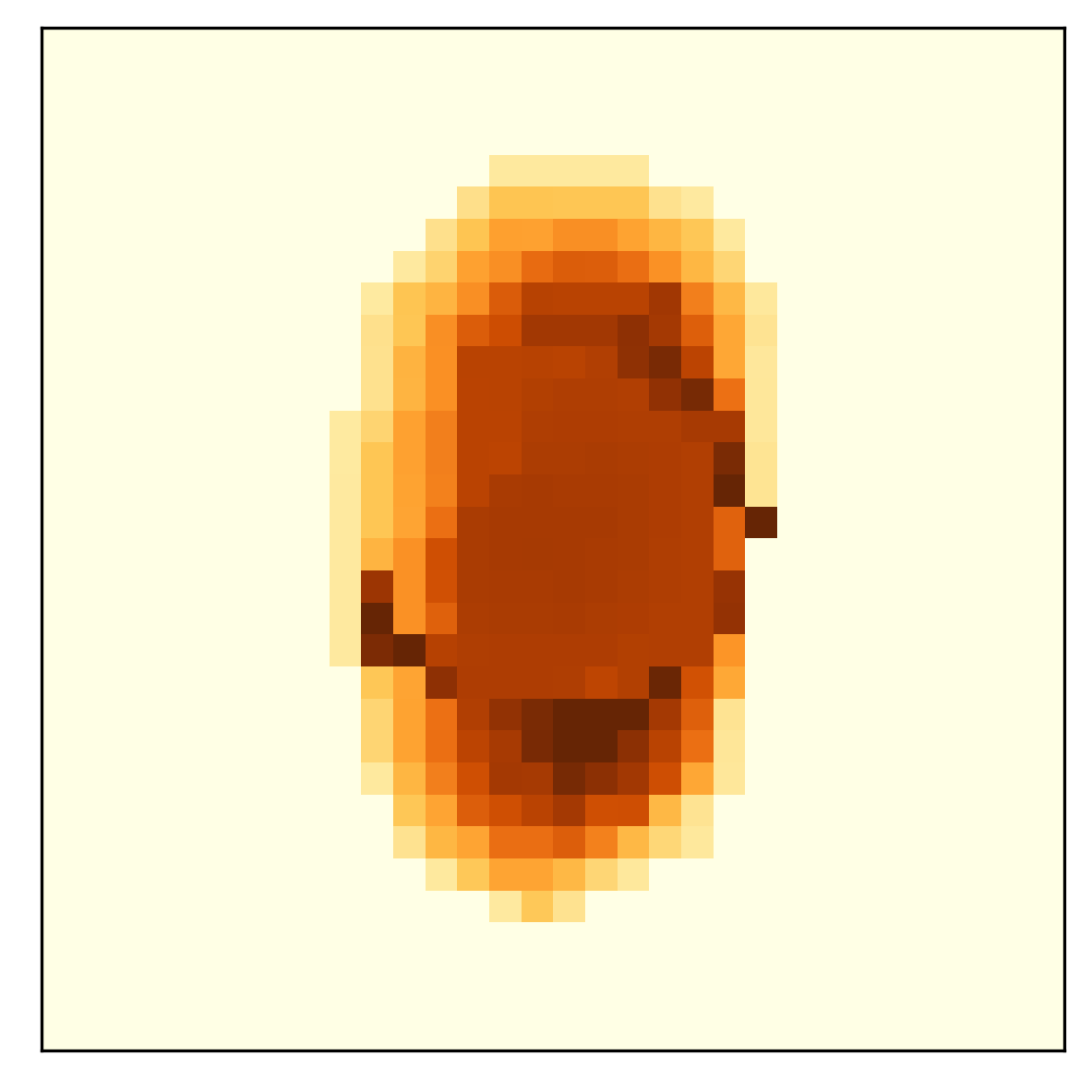}\\
\includegraphics[width=0.905\linewidth]{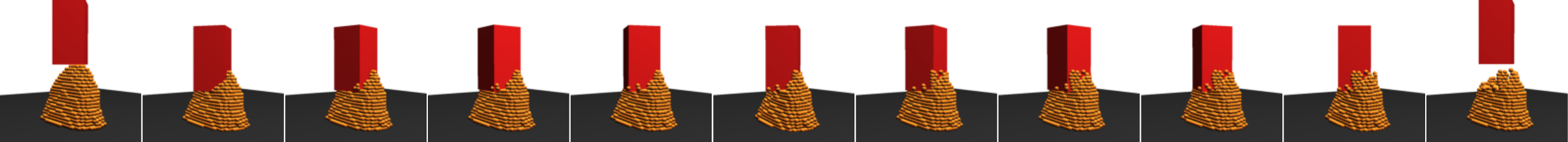}
\includegraphics[width=0.085\linewidth]{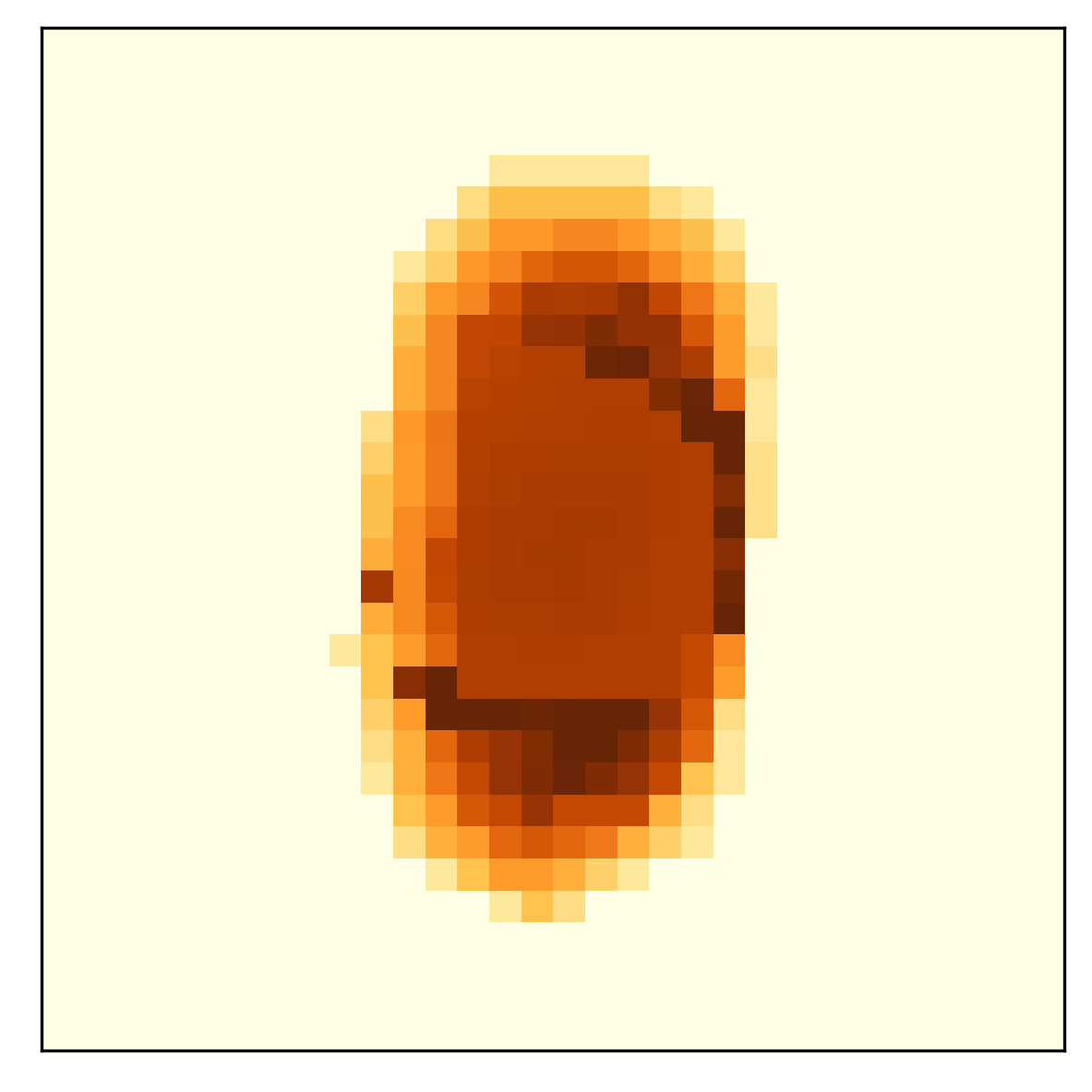}\\
\includegraphics[width=0.905\linewidth]{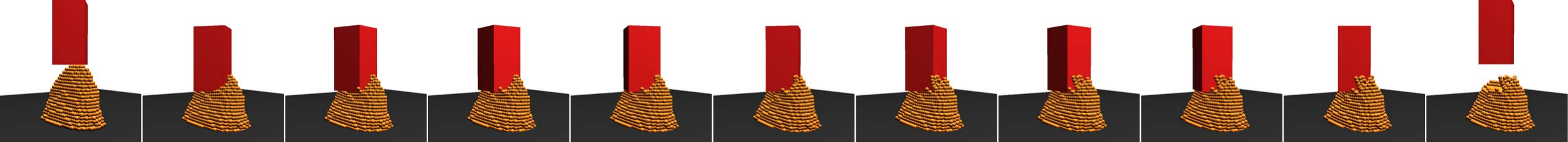}
\includegraphics[width=0.085\linewidth]{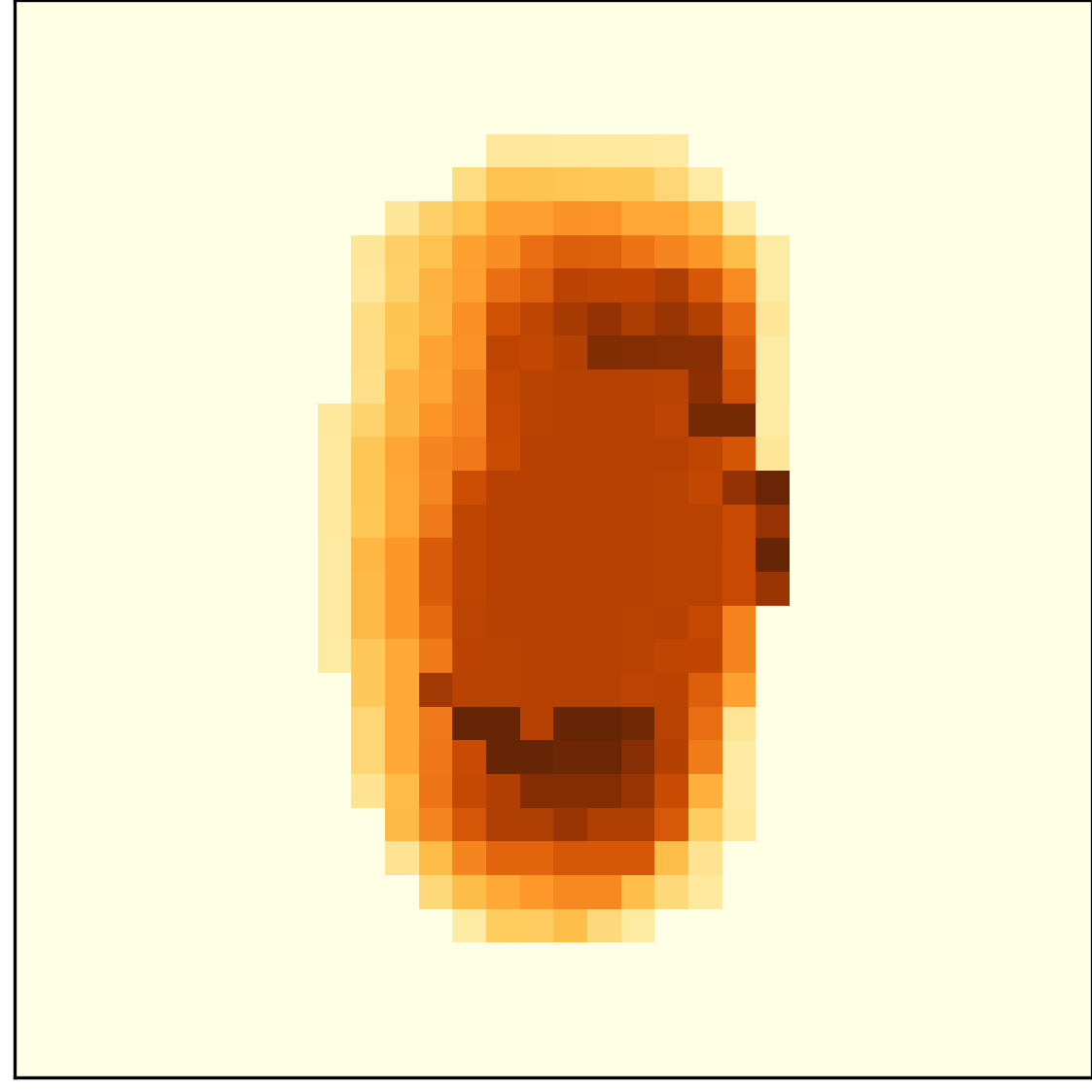}
\vspace{-6mm}
\end{subfigure}
\begin{subfigure}{\columnwidth}
\caption{Poking-180-rotating, object configuration 2}
\vspace{-1mm}
\includegraphics[width=0.905\linewidth]{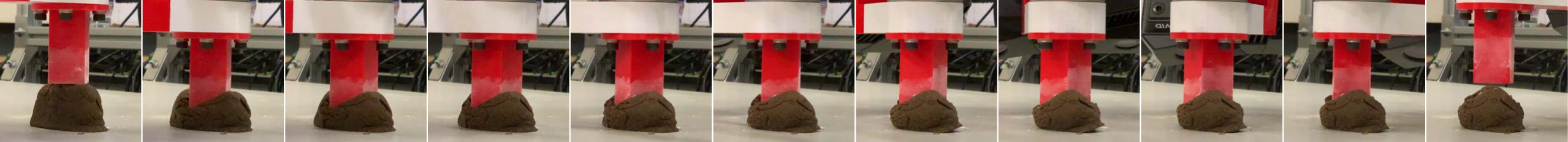}
\includegraphics[width=0.085\linewidth]{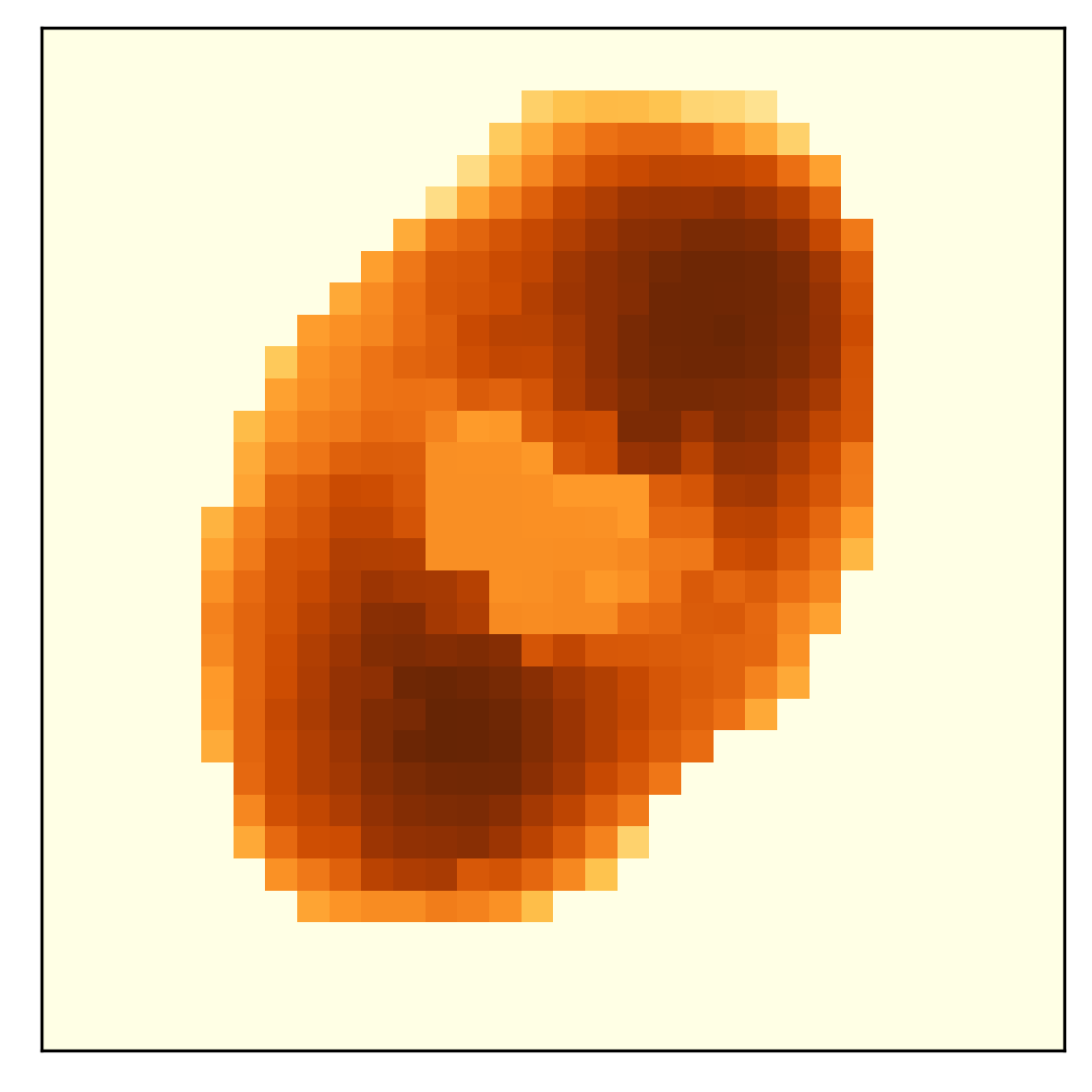}\\
\includegraphics[width=0.905\linewidth]{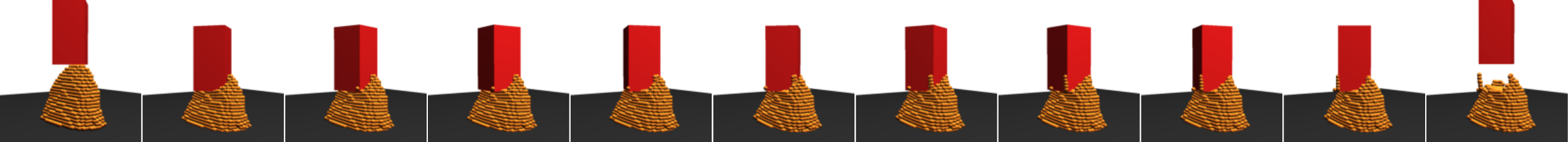}
\includegraphics[width=0.085\linewidth]{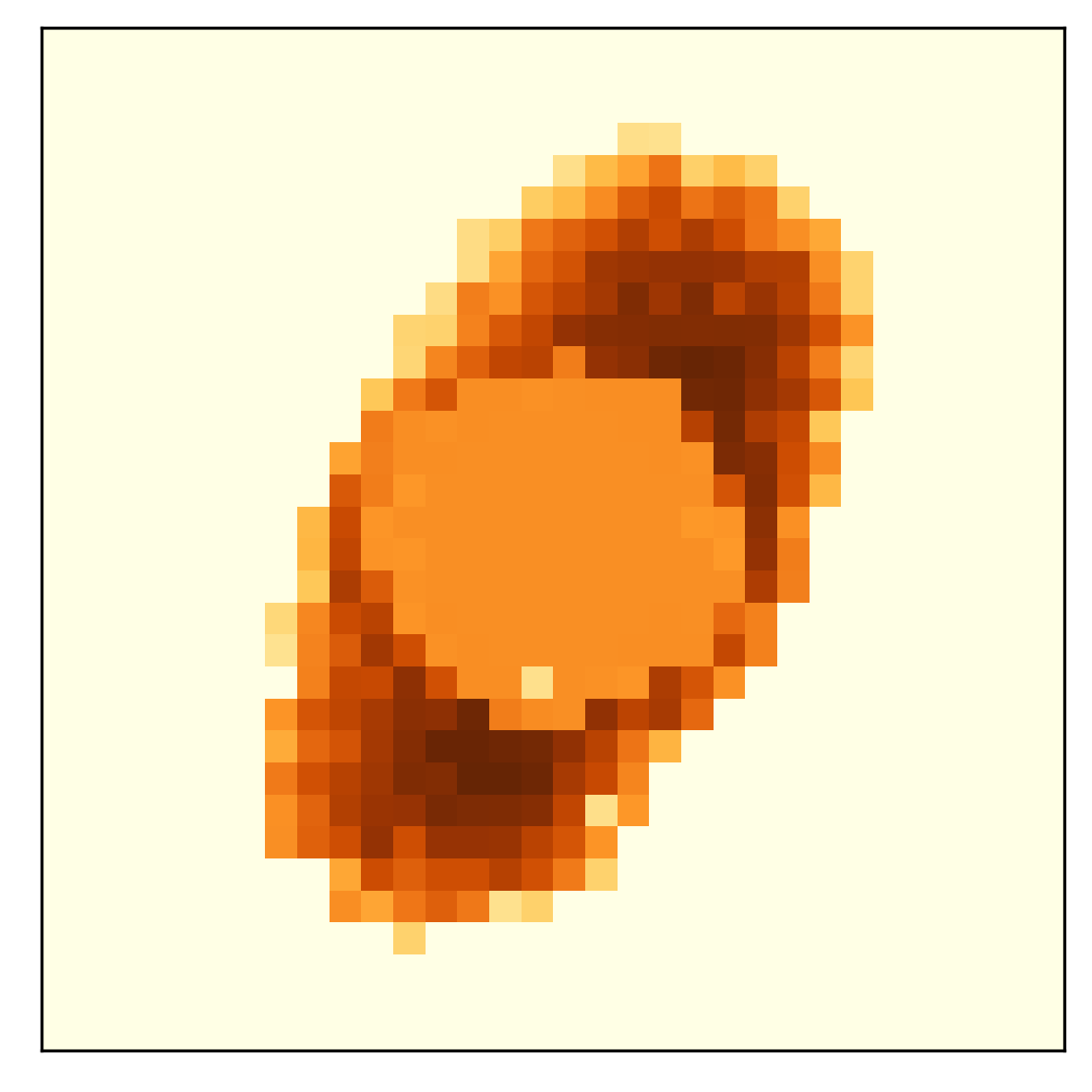}\\
\includegraphics[width=0.905\linewidth]{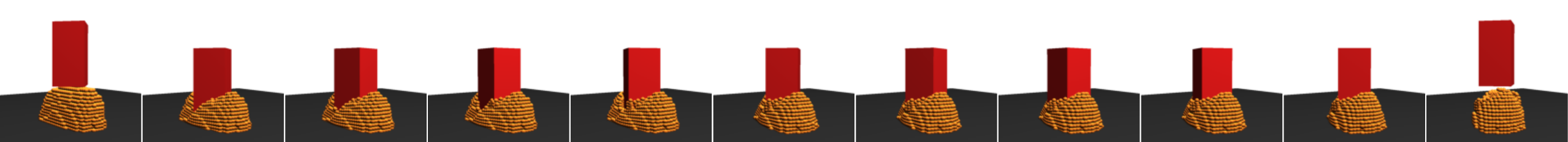}
\includegraphics[width=0.085\linewidth]{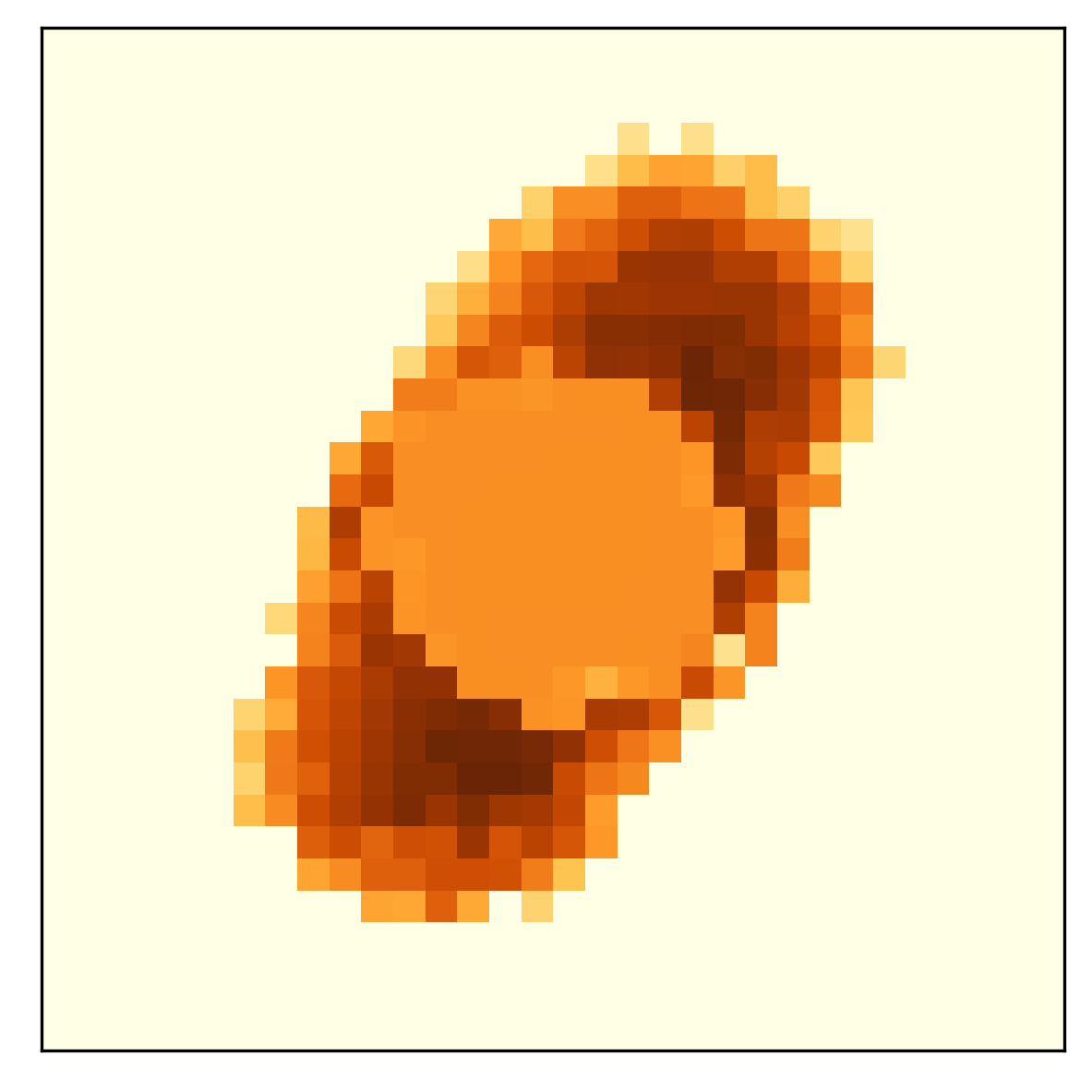}\\
\includegraphics[width=0.905\linewidth]{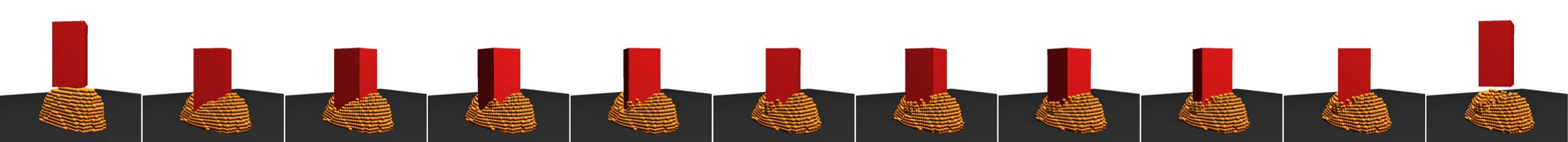}
\includegraphics[width=0.085\linewidth]{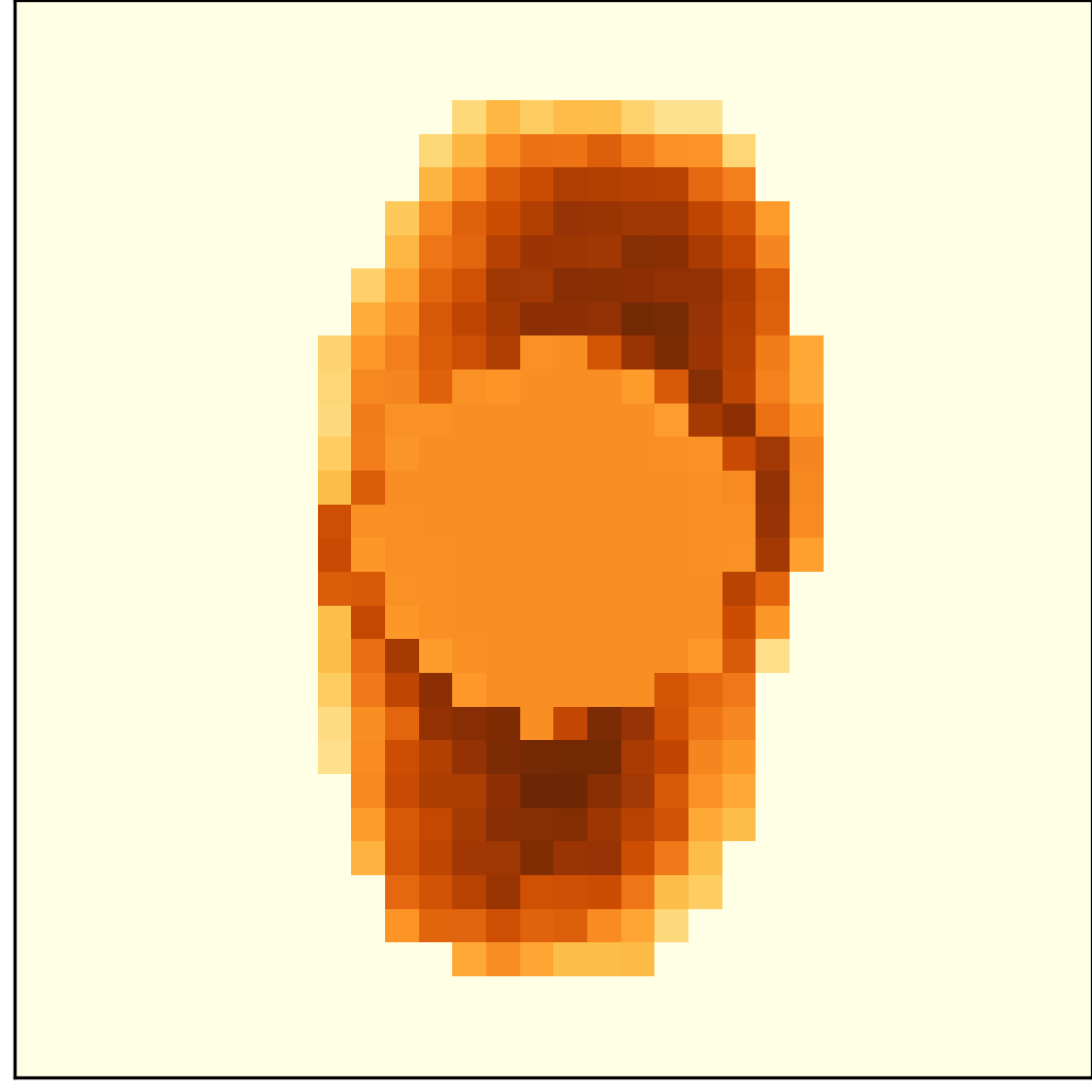}
\includegraphics[width=\linewidth]{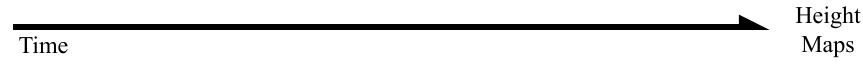}
\end{subfigure}
\vspace{-5mm}
\caption{\xt{The long-horizon manipulation motions. Every subfigure visualises from top to bottom the real manipulation process and its simulations with the top two best DPSI-recovered parameters and hand-picked parameters.}}
\label{fig:out-dis-trs}
\end{figure}

To recall, there are three unseen motions. The triple-poking motion moves the round end-effector to poke the object at three different locations along the $y$ axis. The flattening motion moves the cylinder end-effector to press and flatten the object towards the positive and negative $x$ directions. The poking-180-rotating motion moves the rectangle end-effector to press the object down and then rotate $180$ degrees about the $z$ axis. The triple-poking motion is less relevant to the friction coefficients, while the last two motions involve complex frictional contacts. Table~\ref{tab:unseen-motion-heightmap} summarises the heightmap distances of the simulated results of the unseen motions (each with two object configurations). Figure~\ref{fig:out-dis-trs} visualises the trajectories of the real manipulation trajectory and its simulations with the top two best DPSI-found parameters and the hand-picked parameters. The animation videos of them are included in \textit{Supplementary materials}. Figure~\ref{fig:out-dis-vis} presents the resultant particles and their heightmaps of each motion and object configuration simulated by the DPSI-found parameters. The following observations can be drawn.

First, in Figure~\ref{fig:out-dis-trs}, (as well as the animation video), it can be seen that, for all motion and object configuration pairs, the real and simulated trajectories are highly similar. Figure~\ref{fig:out-dis-vis} also shows that, although discrepancies exist, the simulated particles and their heightmaps of the manipulation results are also highly similar across the loss functions and datasets. This visual similarity proves that DPSI can \xt{recover physics parameters from simple interactions that allow the accurate simulation of} unseen, longer, and more complex manipulation motions, exhibiting robustness against perception and data challenges. \xt{Moreover, we note that it is the single-datapoint cases that achieve $11$ out of $18$ top-three lowest heightmap losses across the three unseen motions as shown in Table~\ref{tab:unseen-motion-heightmap}, where $2$ achieved by the hand-picked parameters and $5$ by the 12mix and 6mix datasets.}

\begin{figure}[h]
    \centering
    \includegraphics[width=0.95\columnwidth]{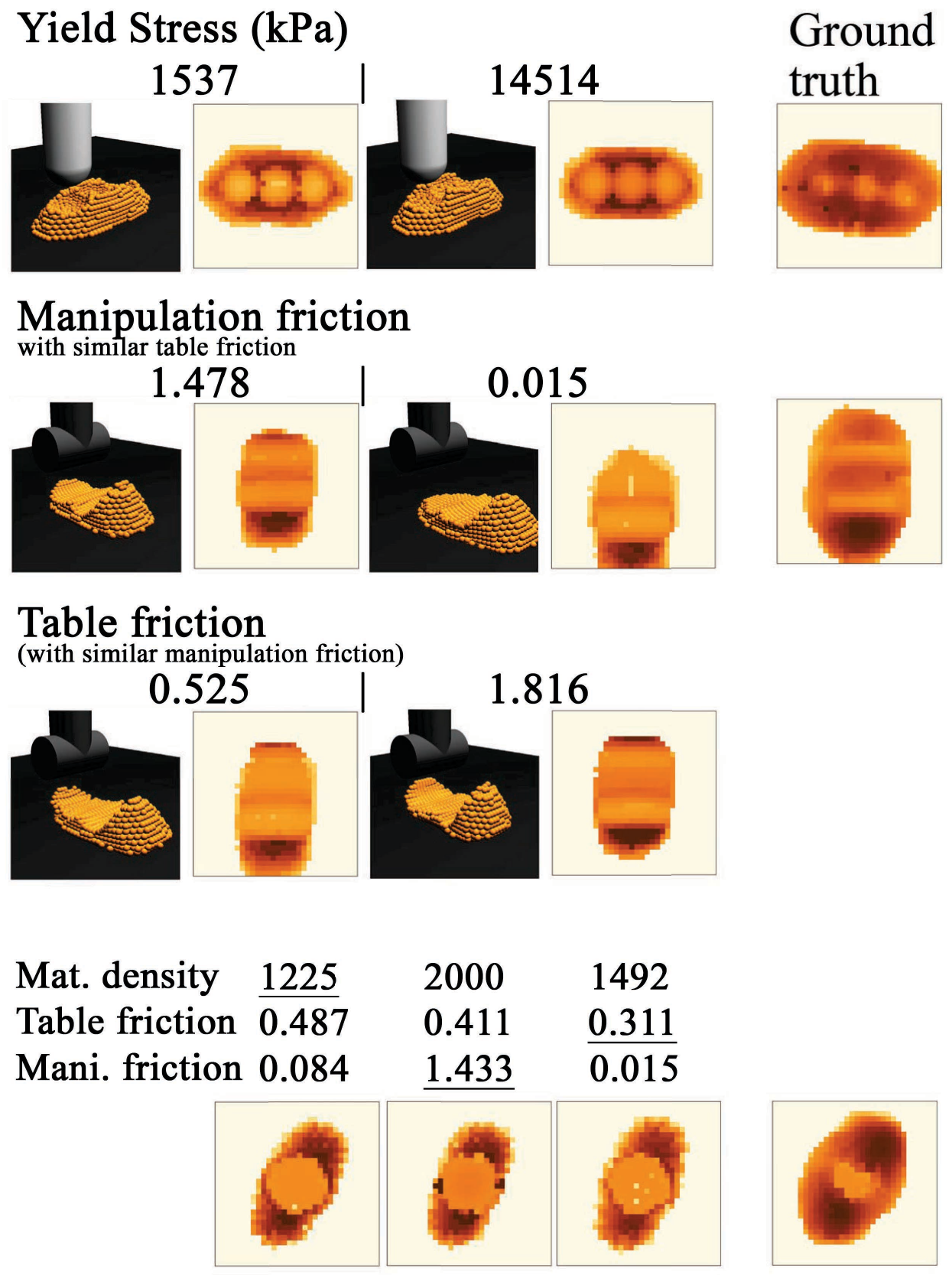}
    \caption{Selected examples from Figure~\ref{fig:out-dis-vis} to show the physically realistic associations between the manipulation results and the found parameter values. Comparisons to the ground truths reveal the modelling limitations (insufficient elastic returning).}
    \label{fig:example-3}
\end{figure}

Secondly, the found physics parameters retain their physical meanings and interpretability for the unseen and more complex motions. Evidence supporting this can be found in Figure~\ref{fig:out-dis-vis} produced by different parameters (see Figure~\ref{fig:example-3} for selected examples). 
\begin{itemize}
    \item The first example is that when the yield stress ($\pmb{\sigma}_y$) is smaller, the resultant heightmaps of the unseen motions tend to be lighter, meaning the objects are more compressed down. 
    \item Another example is that, when moving the cylinder end-effector along the $x$ axis after pressing down the object, a higher manipulator friction coefficient ($\eta_m$) causes larger whole-body displacement of the object, while a smaller value causes the effector to slip away and compress down the particles along its moving direction. This is shown more clearly in the bottom-right case (cylinder end-effector) of Figure~\ref{fig:out-dis-vis}, where the resultant object bodies that are more concentrated to the heightmap centres correspond to larger manipulator friction coefficients, while the objects whose upper part of the heightmaps are flattened correspond to smaller one. 
    \item The last example is that, for the poking-180-rotating motion results shown at the top of Figure~\ref{fig:out-dis-vis}, when the objects are more rotated by the end-effector, at least two of the following three conditions can be verified true from Table~\ref{tab:best-param} at the same time: a relatively small material density, a relatively small table friction coefficient, and a relatively large manipulator friction coefficient. This is physically intuitive as any two of the three conditions would make the object easier to move.
\end{itemize}

\begin{figure}[h]
    \centering
    \includegraphics[width=0.95\columnwidth]{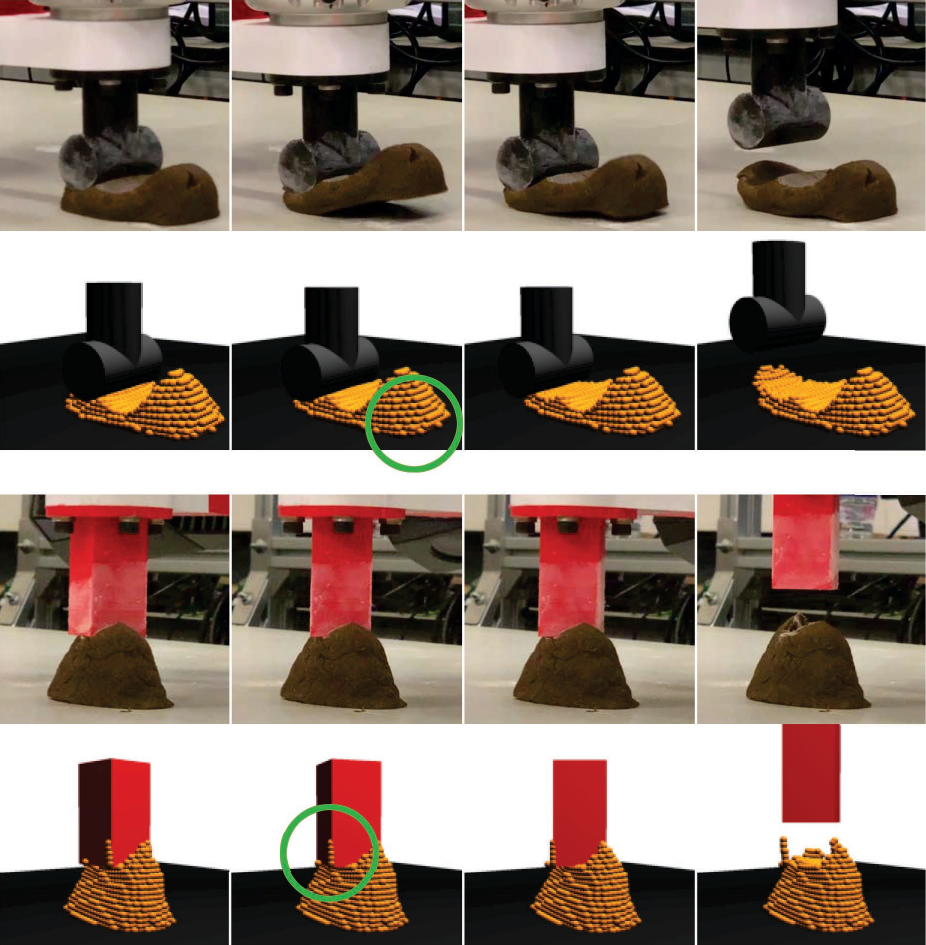}\\
    \includegraphics[width=0.95\columnwidth]{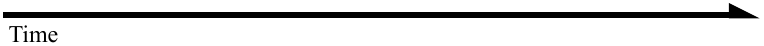}
    \caption{Selected frames from two trajectories in Figure~\ref{fig:out-dis-trs} to show modelling limitations (floating particles and insufficient spreading of contacts). The areas around the green cycles show that the second part of the flattening motion does not tilt the object up, and there are floating particles in the poking-180-rotating motion.}
    \label{fig:example-4}
\end{figure}

\begin{table*}
\footnotesize
\centering
\begin{tabular}{lll|BBBBBc}
\toprule
Motion  & Object    & Num. datapoint    & 12      & 6       & 1 (rec.) & 1 (round) & 1 (cyl.) & Avg. over losses\\
\midrule
        &          & PCD CD  & 2384.72 & 2302.12 & 2394.64 & 3827.77 & \textbf{2288.37} & 2639.52 \\
        & Config.  & PRT CD  & 3318.00 & 2845.03 & 2316.25 & 2309.38 & 4499.31 & 3057.59 \\
        & 1        & PCD EMD & 2700.57 & \underline{2306.85} & 2971.97 & 3346.45 & \textbf{2255.47} & 2716.26 \\
Poking-180&        & PRT EMD & \underline{\textbf{2290.37}} & 2299.37 & 3605.89 & \underline{2361.34} & 2357.05 & 2582.81 \\
        &          & Avg. over dataset    & 2673.42 & 2438.34 & 2822.19 & 2961.24 & 2850.05 & -\\
\cline{3-9}
        &&\multicolumn{6}{l}{Hand-picked parameters: *2132.04}\\
\cline{2-9}
rotating&          & PCD CD  & 5367.10 & 5345.52 & 5473.18 & 4822.09 & 5770.72 & 5355.72 \\
        & Config.  & PRT CD  & 4218.07 & 4227.91 & 5802.93 & 5766.13 & *\textbf{3994.08} & 4801.82\\
        & 2        & PCD EMD & \textbf{4187.56} & \underline{4469.51} & 4300.81 & \textbf{4026.08} & 5467.30 & 4490.25 \\
        &          & PRT EMD & \underline{4175.07} & 4768.15 & 4323.36 & \underline{4226.68} & 4978.39 & 4494.33 \\
        &          & Avg. over dataset    & 4486.95 & 4702.77 & 4975.07 & 4710.25 & 5052.62 & -\\
\cline{3-9}
        &&\multicolumn{6}{l}{Hand-picked parameters: 5283.87}\\
\hline
        &          & PCD CD  & 2956.06 & 2919.90 & 3028.85 & 3089.34 & 2949.11 & 2988.65 \\
        & Config.  & PRT CD  & 2960.77 & 2971.96 & 2962.99 & 2963.06 & *\textbf{2798.67} & 2931.49 \\
        & 1        & PCD EMD & 2895.86 & \underline{2875.01} & \textbf{2839.52} & 2863.56 & 3003.25 & 2895.44 \\
Triple- &          & PRT EMD & \underline{3100.87} & 3035.49 & \textbf{2852.95} & \underline{3034.08} & 3059.61 & 3016.60 \\
        &          & Avg. over dataset    & 2978.39 & 2950.59 & 2921.08 & 2987.51 & 2952.66 & -\\
\cline{3-9}
        &&\multicolumn{6}{l}{Hand-picked parameters: 2983.62}\\
\cline{2-9}
poking  &          & PCD CD  & *\textbf{4175.07} & 4768.15 & 4323.36 & \textbf{4226.68} & 4978.39 & 4630.09 \\
        & Config.  & PRT CD  & 4395.60 & 4399.77 & 4731.59 & 4685.70 & \textbf{4245.47} & 4491.62 \\
        & 2        & PCD EMD & 4430.33 & \underline{4495.40} & 4417.59 & 4426.74 & 4379.40 & 4429.89 \\
        &          & PRT EMD & \underline{4529.65} & 4304.59 & 4300.22 & \underline{4272.90} & 4536.17 & 4388.70 \\
        &          & Avg. over dataset    & 4508.76 & 4467.01 & 4499.59 & 4505.60 & 4444.43 & -\\
\cline{3-9}
        &&\multicolumn{6}{l}{Hand-picked parameters: 4517.24}\\
\hline
        &          & PCD CD  & 2443.54 & 2536.84 & 2153.18 & 2244.94 & 2403.77 & 2356.45 \\
        & Config.  & PRT CD  & 2020.59 & \textbf{1988.33} & 2496.75 & 2547.35 & \textbf{1860.49} & 2182.70 \\
        & 1        & PCD EMD & 2149.35 & *\underline{\textbf{1835.63}} & 2233.75 & 2196.98 & 2545.52 & 2192.25 \\
Flattening&        & PRT EMD & \underline{2275.80} & 2439.66 & 2407.77 & \underline{2287.72} & 2539.93 & 2390.18 \\
        &          & Avg. over dataset    & 2222.32 & 2200.12 & 2322.86 & 2319.25 & 2337.43 & -\\
\cline{3-9}
        &&\multicolumn{6}{l}{Hand-picked parameters: 2051.06}\\
\cline{2-9}
        &          & PCD CD  & 5325.90 & 5378.31 & \textbf{4752.87} & \textbf{4756.35} & 5743.95 & 5191.48 \\
        & Config.  & PRT CD  & 5622.73 & 5387.05 & 5282.32 & 5119.63 & 5817.57 & 5445.86 \\
        & 2        & PCD EMD & 5058.42 & \underline{\textbf{4697.61}} & 5798.36 & 5891.09 & 5190.65 & 5327.23 \\
        &          & PRT EMD & \underline{4996.47} & 5460.32 & 5686.20 & \underline{5152.61} & 5493.27 & 5357.77  \\
        &          & Avg. over dataset    & 5250.88 & 5230.82 & 5379.94 & 5229.92 & 5561.36 & -\\
\cline{3-9}
        &&\multicolumn{6}{l}{Hand-picked parameters: *4483.75}\\
\hline
\bottomrule
\end{tabular}
\caption{The lowest pixel-wise heightmap distances (in mm) achieved by the three unseen motions (each with two object configurations), simulated by the best parameters found at the level 2 contact complexity identification task. Bolded texts denote the top three lowest losses. Underlined texts denote the losses achieved by the cases which have achieved the top-three lowest in-distribution validation performances.}
\label{tab:unseen-motion-heightmap}
\end{table*}

\begin{figure*}
\begin{subfigure}{\columnwidth}
\caption{Object configure 1}
\includegraphics[width=0.99\columnwidth]{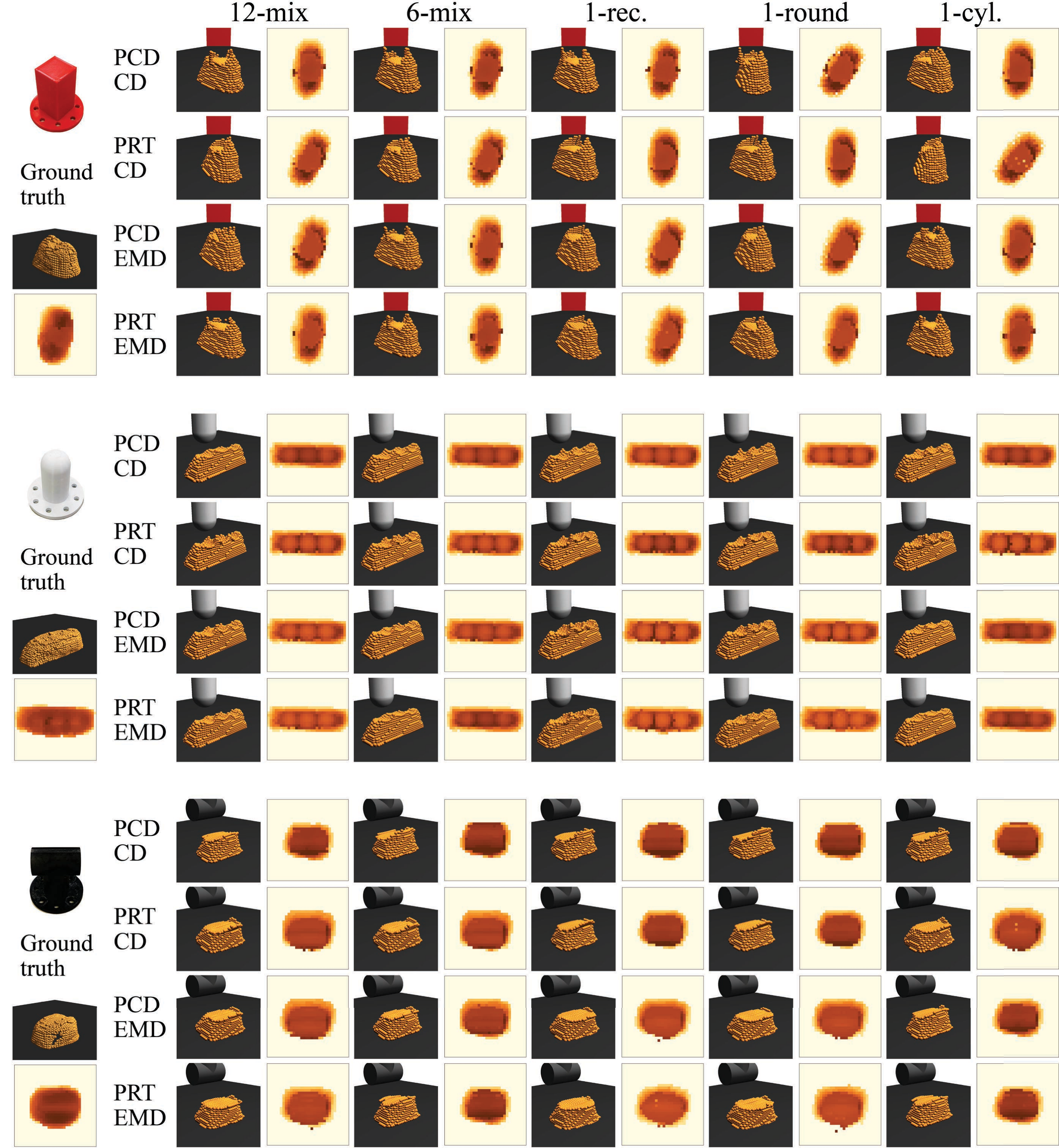}    
\end{subfigure}
\begin{subfigure}{\columnwidth}
\caption{Object configure 2}
\includegraphics[width=0.99\columnwidth]{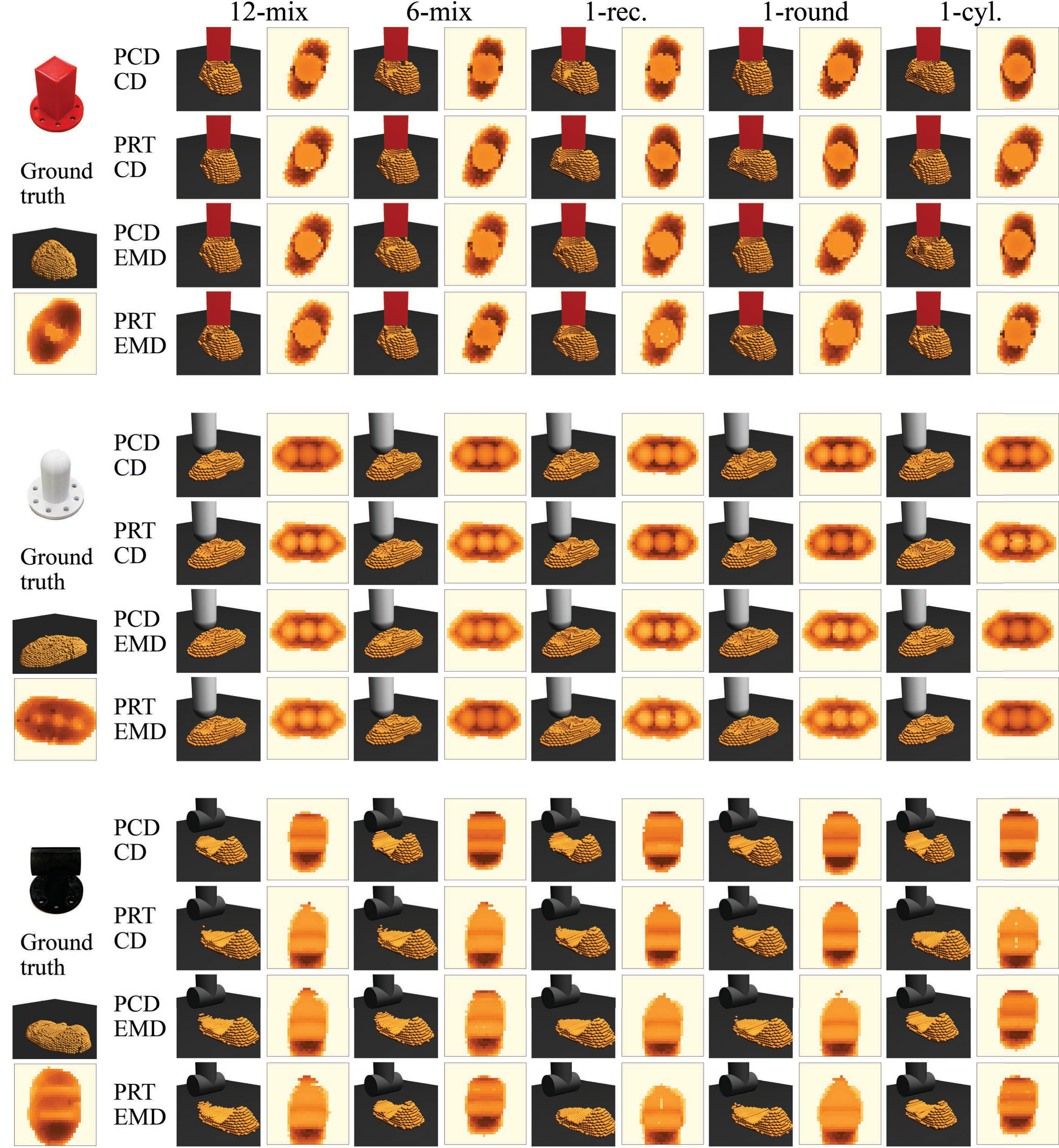}
\end{subfigure}
\caption{The particles and heightmaps of the objects after applying the long-horizon unseen motions, simulated with the best set of physics parameters found at the level 2 contact complexity experiments. Darker colours of the heightmaps indicate greater heights of the object. For each motion, two object configurations are manipulated with three end-effectors. The results for each trajectory are grouped with the ground truth placed on the left. In each group, a row shows the results corresponding to a loss function and a column results corresponding to a dataset.}
\label{fig:out-dis-vis}
\end{figure*}

Thirdly, the found solutions are not globally optimal. Despite the visual similarity, Table~\ref{tab:unseen-motion-heightmap} shows that the best generalisation results are not always achieved by the cases with the best in-distribution validation performances, as the underlined and bolded texts rarely coincide. \xt{In addition, though the hand-picked parameters do achieve the best results in two out of six object simulations, they are not the best fit for every scenario. This again suggests the importance of re-fitting the parameters on the fly for every new scenario as there are hardly two objects that are identical especially for elastoplastic materials.} Qualitatively, from the heightmaps and trajectories in Figures~\ref{fig:out-dis-vis} and~\ref{fig:out-dis-trs}, one can also recognise certain visual discrepancies between the real objects and their simulation counterparts, indicating a certain degree of mismatch between the found parameters and the actual solutions. This could be improved by further diversifying the data (and the motions that produce them) used for optimisation. Other directions for improvement may include \xt{better observational noise removal or better physics model selection/approximation.}

Finally, the modelling limitations observed in the in-distribution visualisation still exist in the unseen motion visualisations (Figures.~\ref{fig:out-dis-trs} and \ref{fig:out-dis-vis}). The insufficient elastic returning at contact areas, especially with sharp manipulator edges, can be seen from all visualisations where the primary contact areas of the object rarely return to heights close to the real object (compare also the enlarged heightmaps in Figure~\ref{fig:example-3}). Floating displaced particles can be observed from the poking-180-rotating motion visualisations, where many particles are carried away during rotations. The insufficient spreading of the influence of contacts can be recognised from the first object configuration of the flattening motion visualisations where the object is not tilted enough in the second half of the trajectory. See Figure~\ref{fig:example-4} for enlarged examples.

In summary, the results and discussions above strongly support the argument that DPSI \xt{can recover physics parameters that produce the accurate simulation of} unseen, longer, and more complex deformable object manipulation dynamics by optimising with very few real-world noisy and incomplete point-cloud data collected by simple and short interacting motions. 

\subsection{Cloud slime \& soil materials}
\begin{figure}[t]
\centering
\begin{subfigure}{\columnwidth}
\caption{Cloud slime}
\includegraphics[width=0.905\columnwidth]{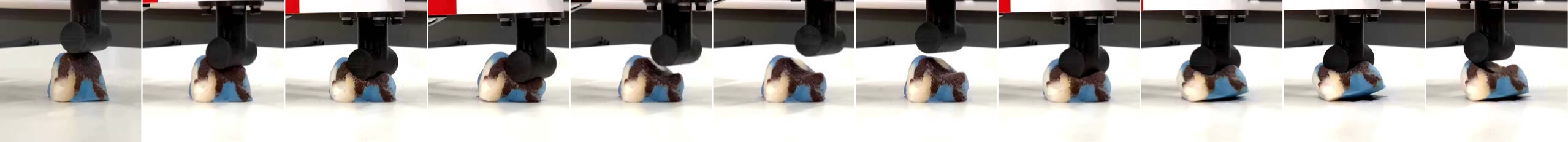}
\includegraphics[width=0.085\columnwidth]{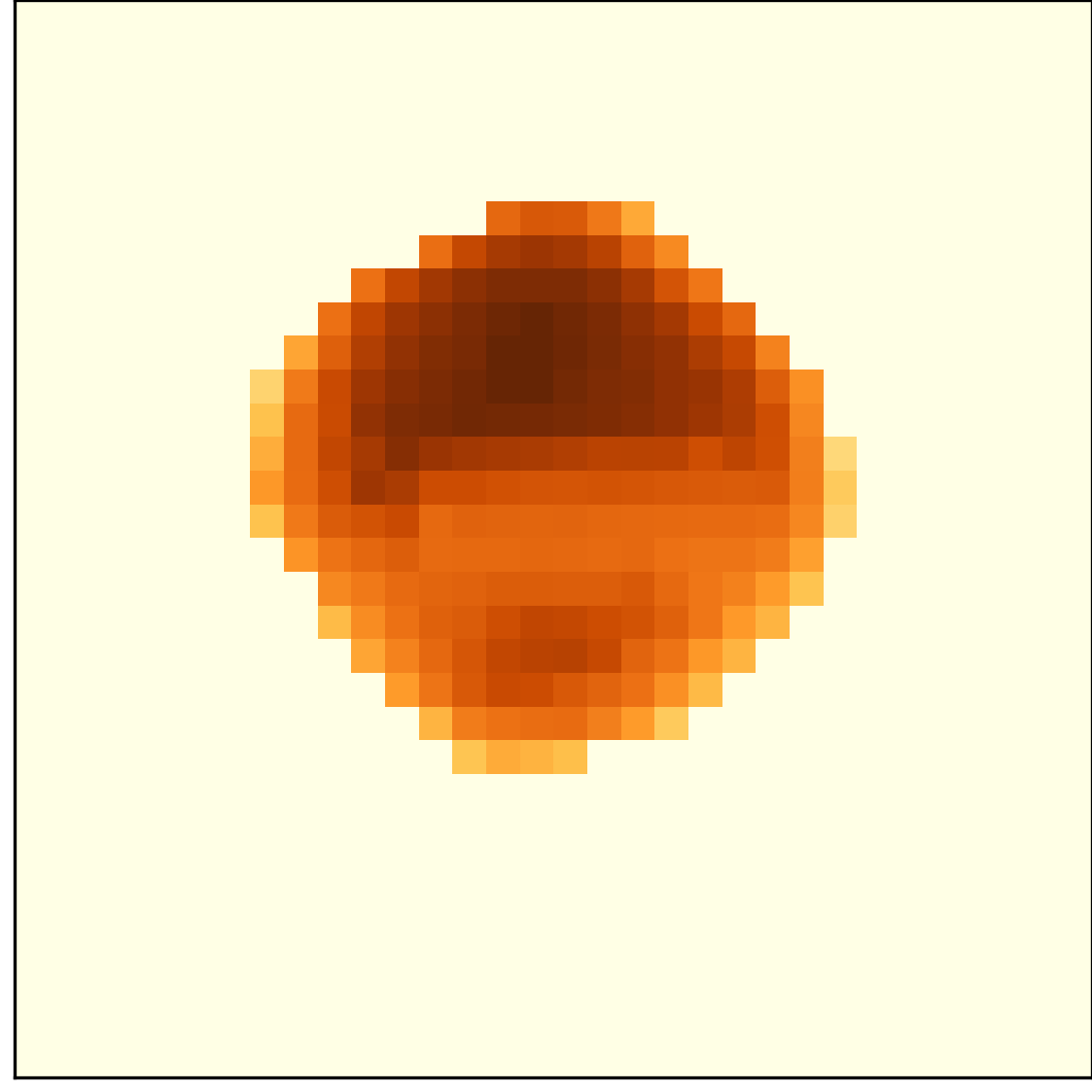}\\
\includegraphics[width=0.905\columnwidth]{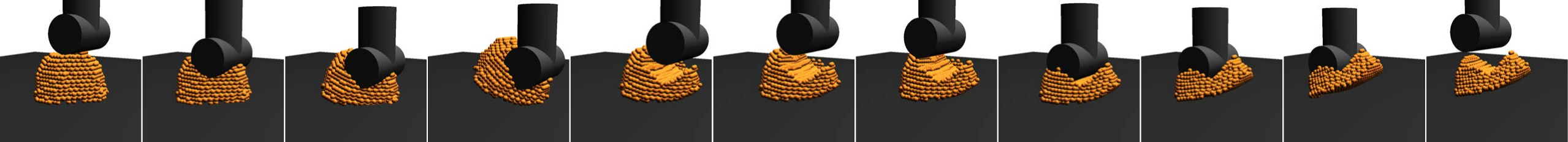}
\includegraphics[width=0.085\columnwidth]{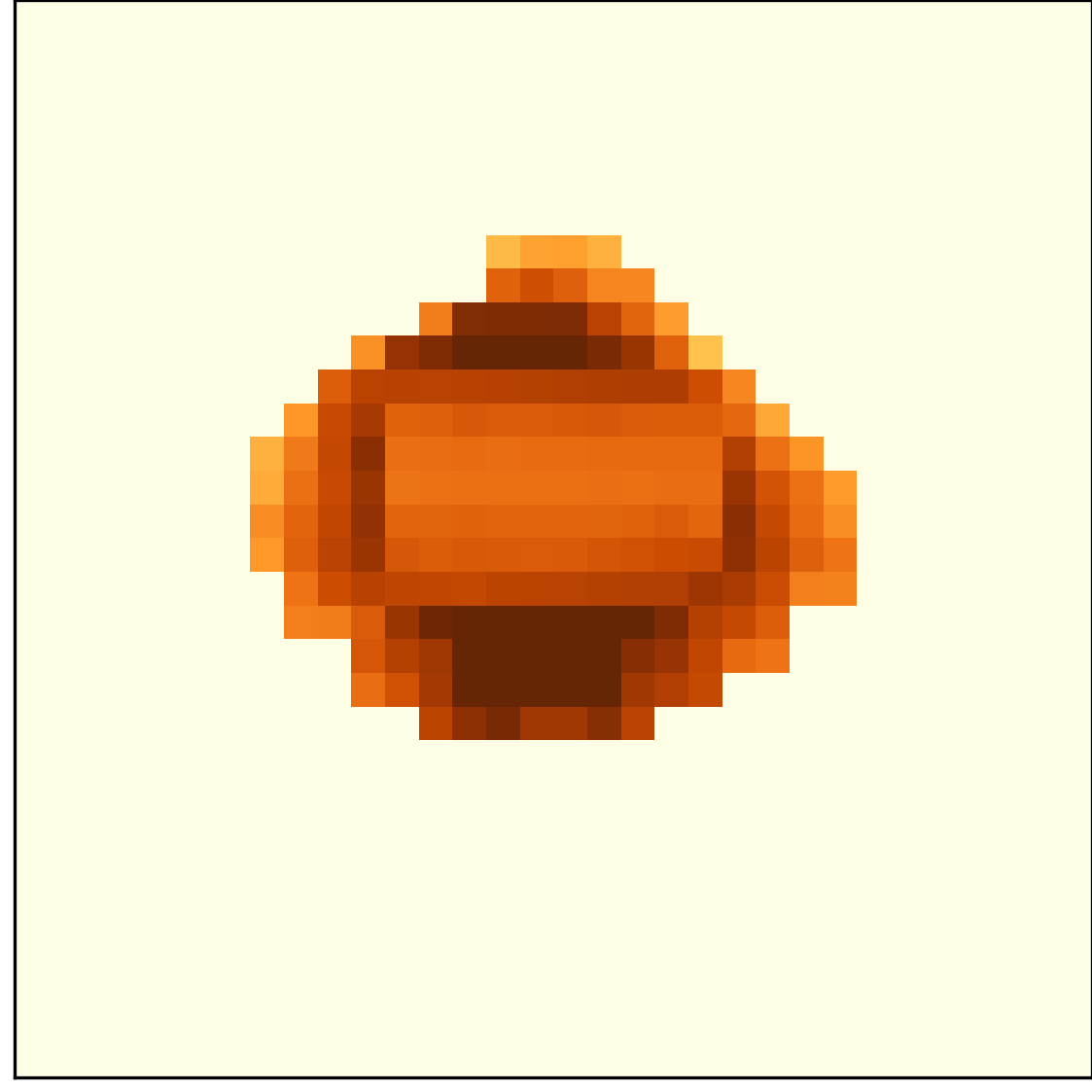}\\
\includegraphics[width=0.905\columnwidth]{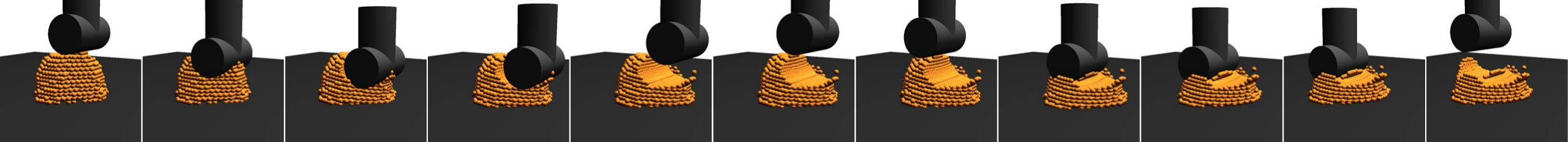}
\includegraphics[width=0.085\columnwidth]{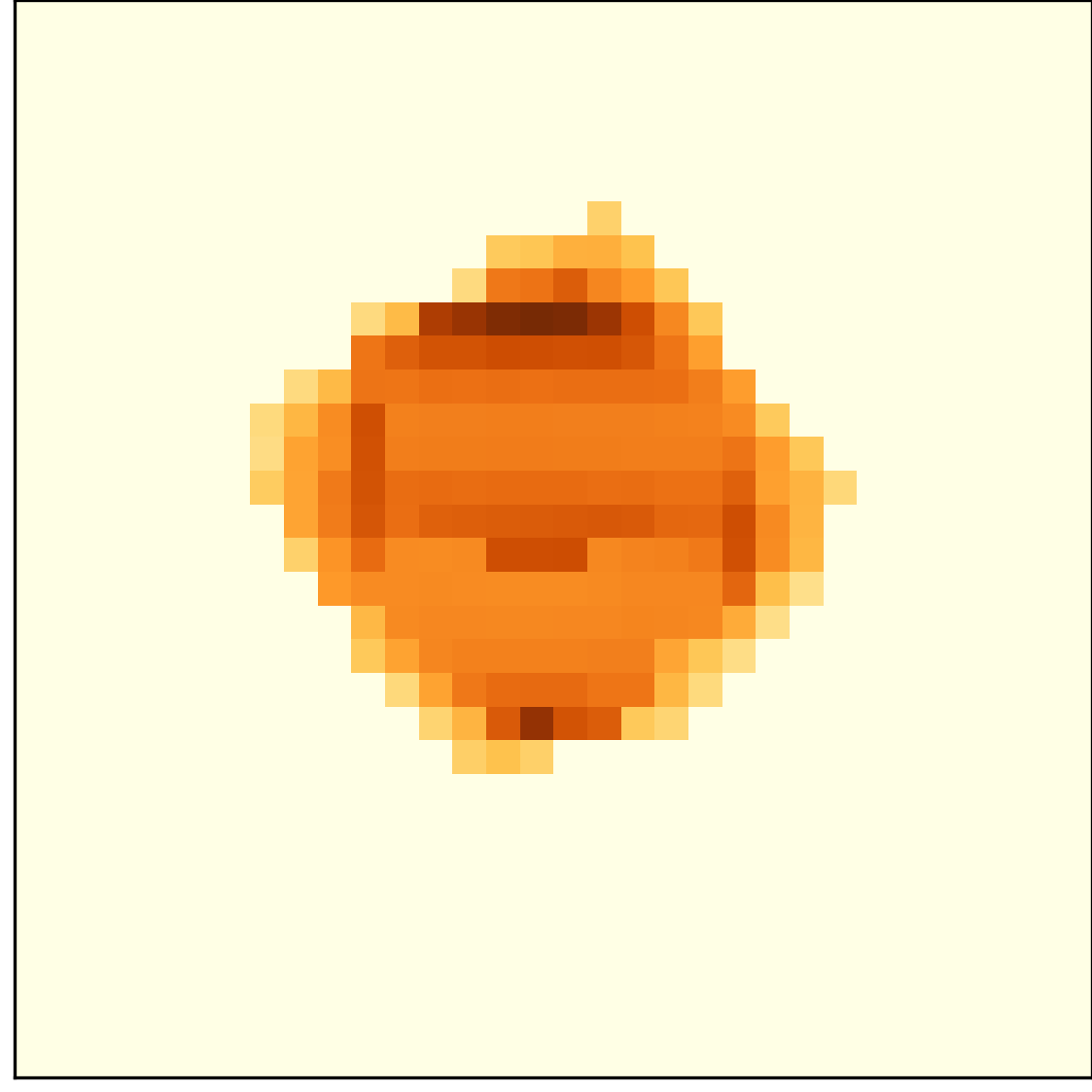}\\
\includegraphics[width=0.905\columnwidth]{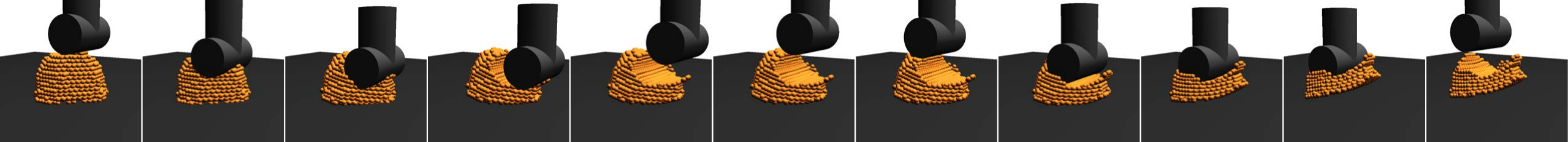}
\includegraphics[width=0.085\columnwidth]{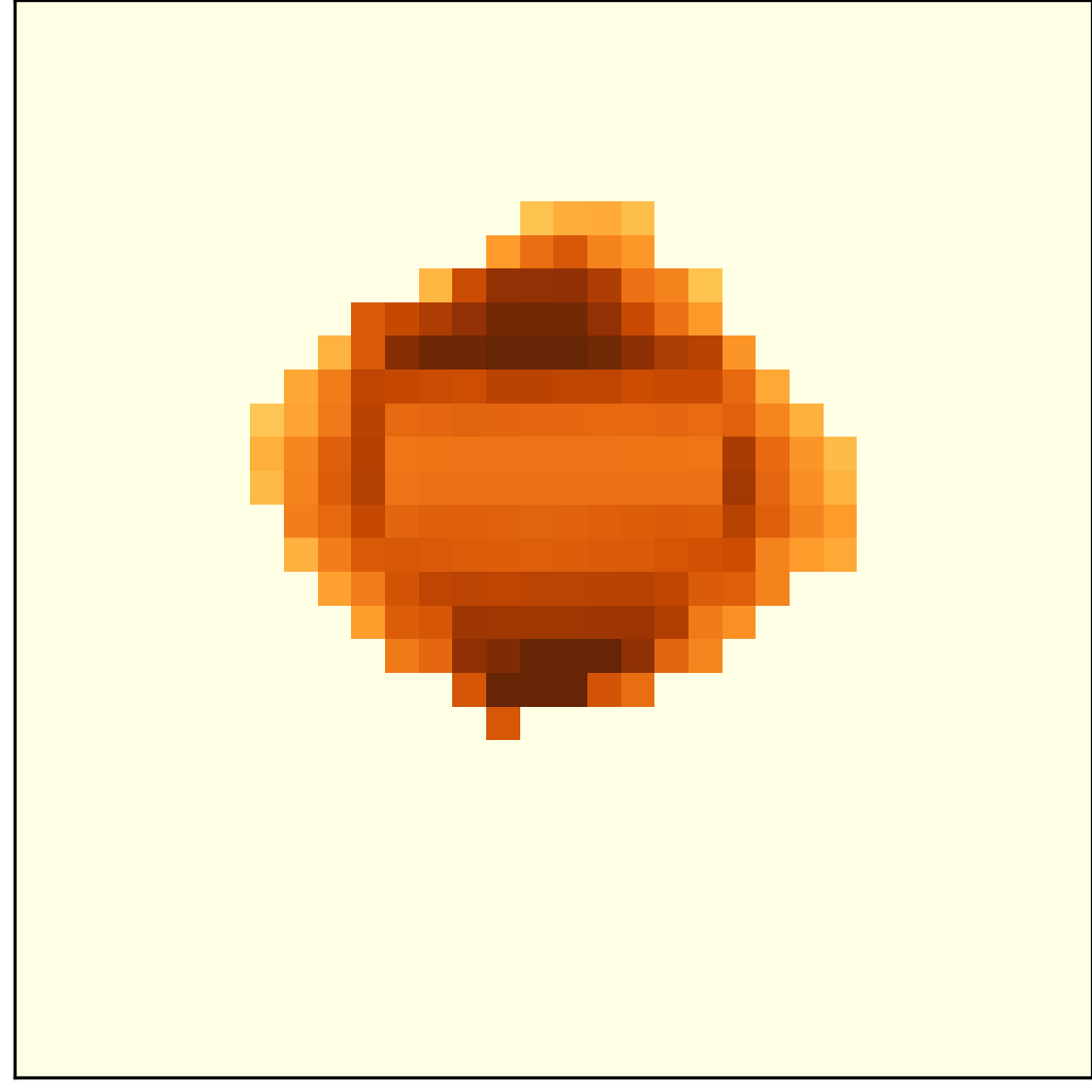}
\label{subfig:slime}
\end{subfigure}
\begin{subfigure}{\columnwidth}
\vspace{-6mm}
\caption{Soil}
\includegraphics[width=0.905\columnwidth]{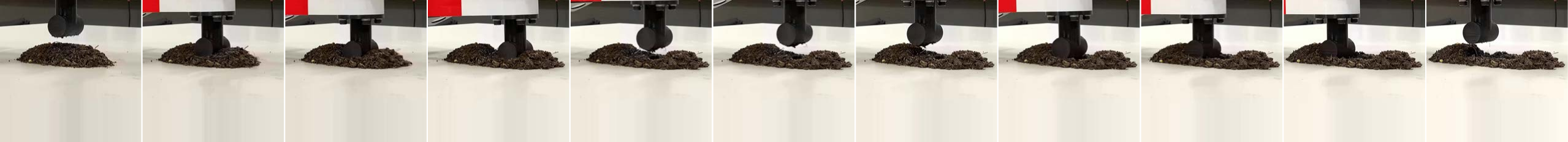}
\includegraphics[width=0.085\columnwidth]{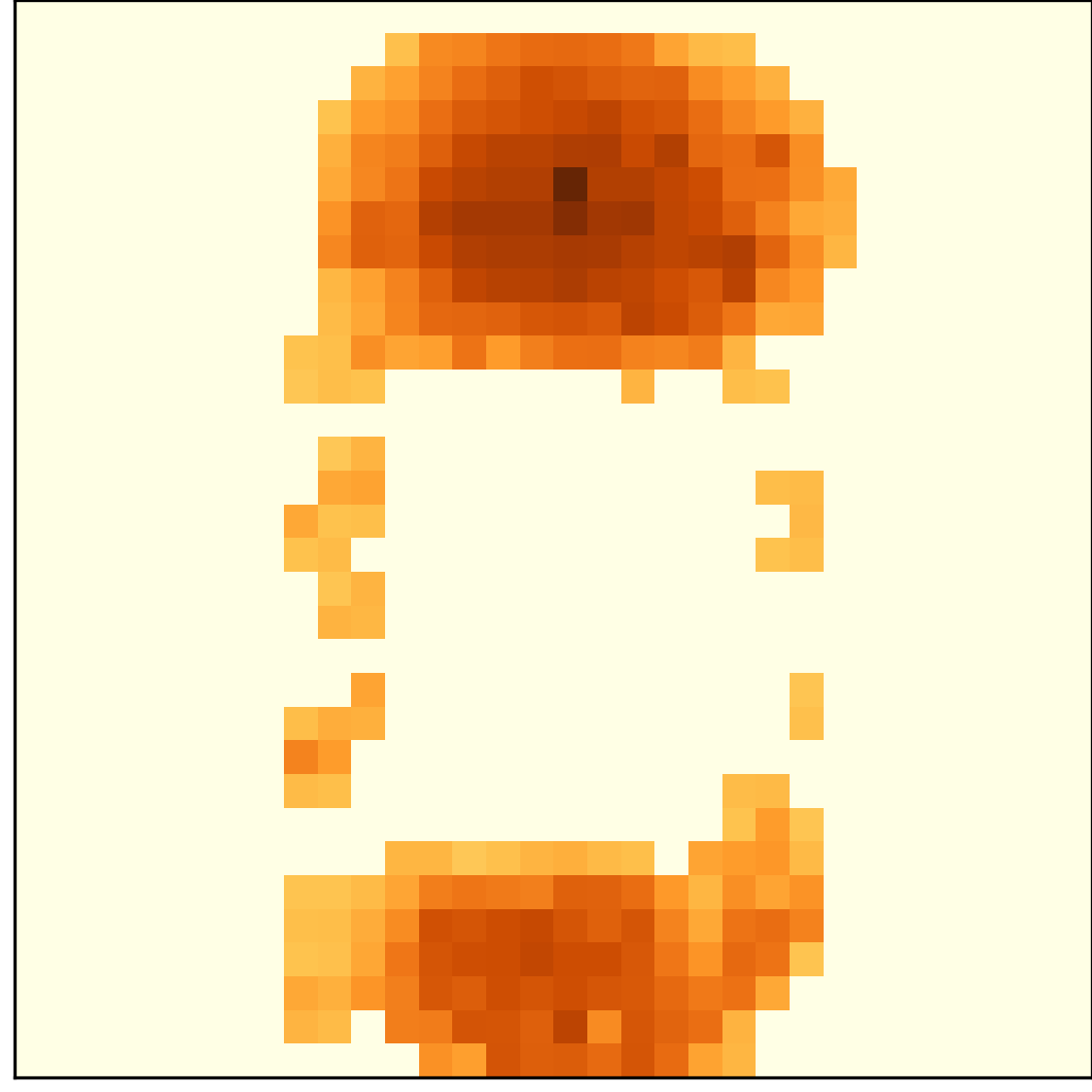}\\
\includegraphics[width=0.905\columnwidth]{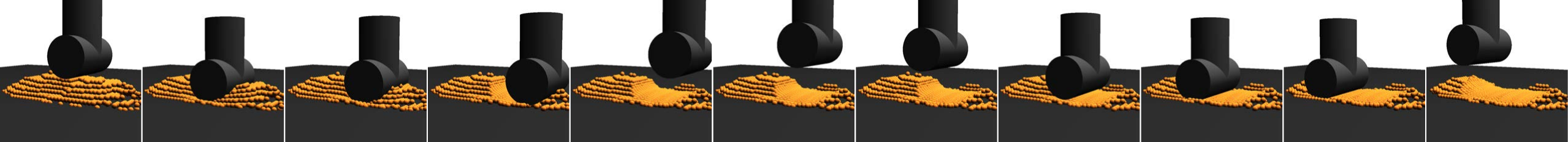}
\includegraphics[width=0.085\columnwidth]{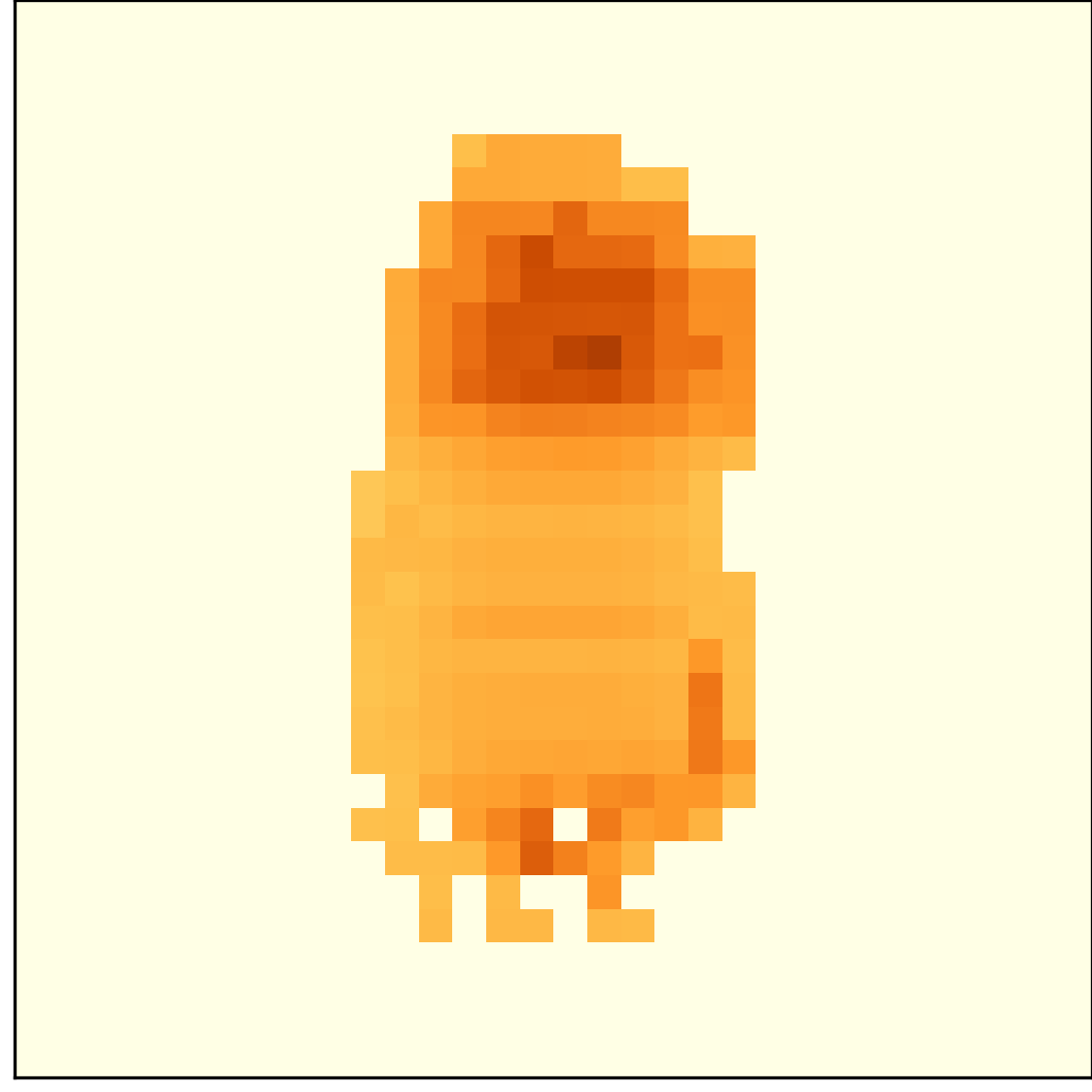}\\
\includegraphics[width=0.905\columnwidth]{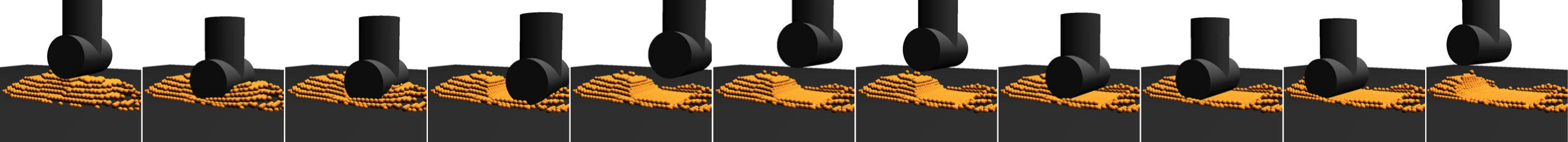}
\includegraphics[width=0.085\columnwidth]{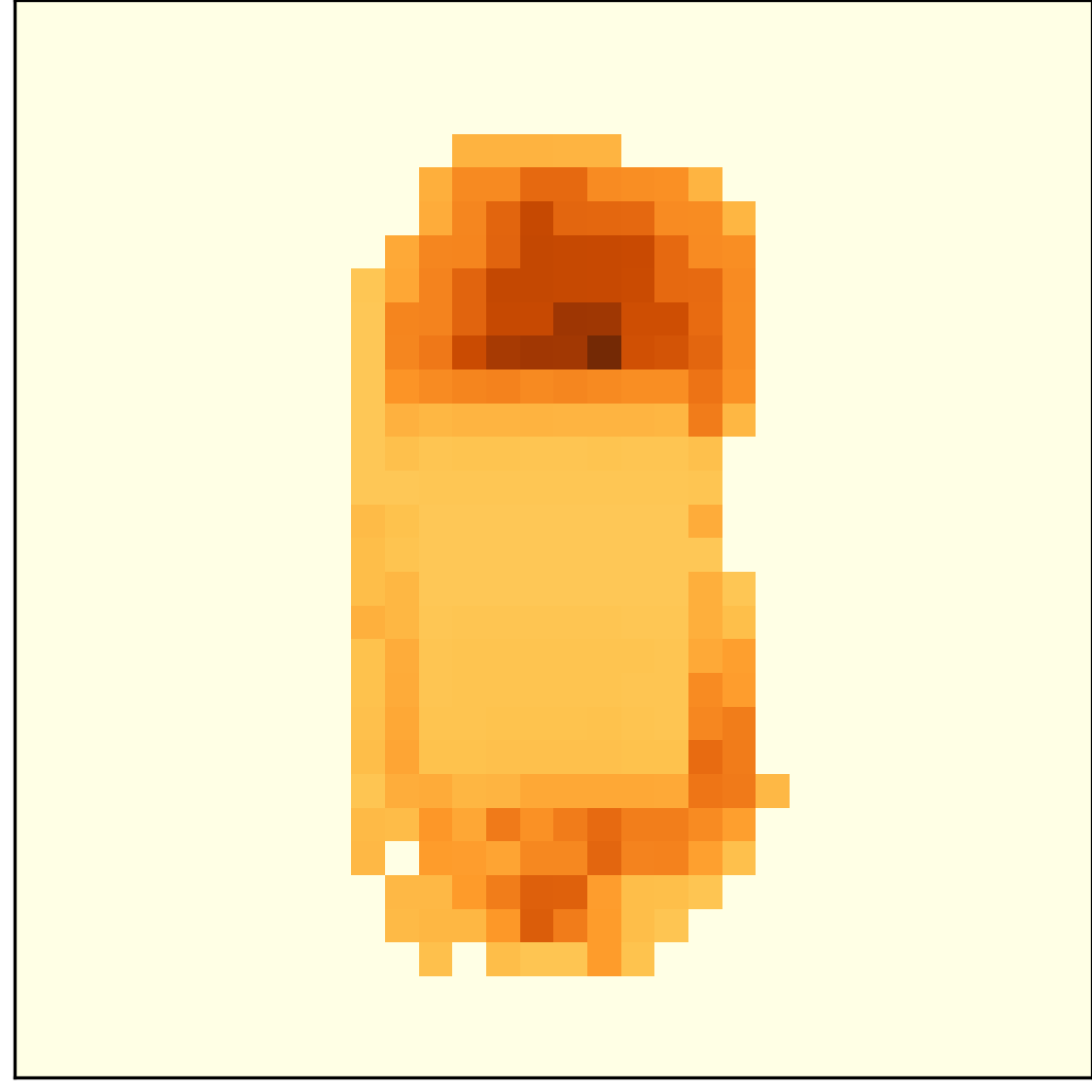}\\
\includegraphics[width=0.905\columnwidth]{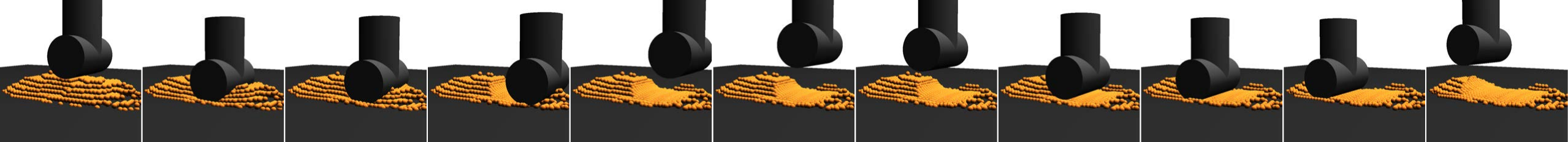}
\includegraphics[width=0.085\columnwidth]{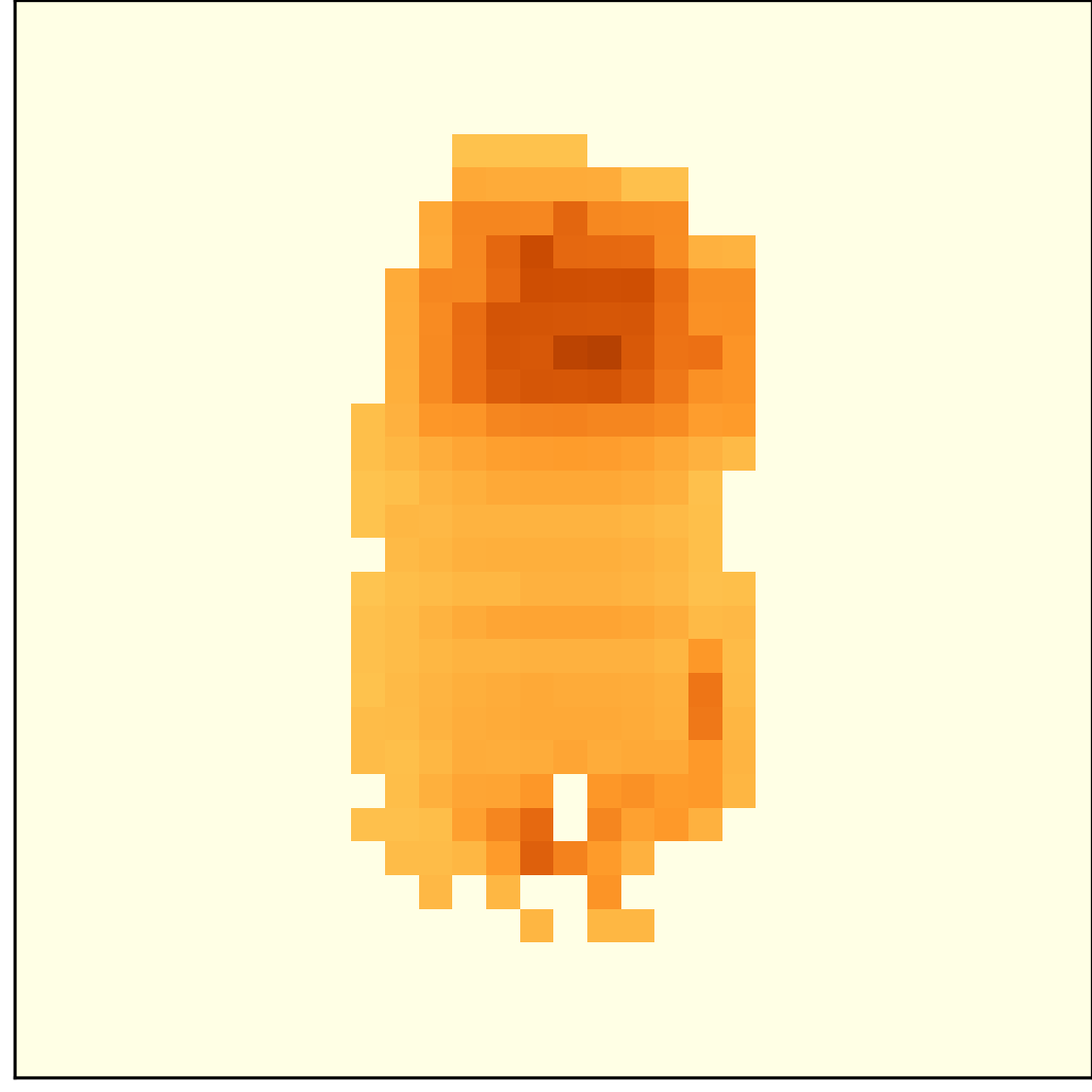}\\
\includegraphics[width=\columnwidth]{note_images/long_motion_comparison/Time_arrow.pdf}
\label{subfig:soil}
\end{subfigure}
\vspace{-10mm}
\caption{\xt{Comparison of the real manipulation trajectories and the resultant height maps for the cloud slime and soil materials and their simulation counterparts based on the recovered physics parameters. In each subfigure, the second to last rows correspond to the results with three random seeds.}}
\label{fig:slime-soil}
\end{figure}

\xt{This subsection reports the DPSI performances with cloud slime and soil materials. Note that we use the same elastoplastic physics models in simulation. To do so, we collect a single interaction data point for each material using the cylinder end-effector and the poking-shifting-2 motion, optimise the parameters with the PRT EMD loss from a randomised initialisation for $100$ iterations with three random seeds, and validate the results on another data collected with the flattening motion and the cylinder end-effector. The real manipulation trajectory of the flattening motion and its simulations based on the found parameters are shown in Figure~\ref{fig:slime-soil}. The recovered parameters are presented in Table~\ref{tab:slime-soil-param}.}

\xt{Figure~\ref{fig:slime-soil} shows that DPSI manages to recover parameters that simulate cloud slime and soil as closely as possible despite the mismatch of the physics models and the target materials. Especially in the case of soil, Table~\ref{tab:slime-soil-param} shows that DPSI finds large table friction coefficients and small end-effector friction coefficients to simulate the effect of the soil being flattened, although these are likely not close to the actual values. This suggests that DPSI will seek to find parameters that make the deformations of the simulated object as realistic as possible regardless of whether the physics models are a good match for the real materials. On the other hand, this also suggests that the manual selection of the physics models is important for DPSI to produce the best deformation simulation.}

\begin{table}[h]
\footnotesize
\centering
\begin{tabular}{lL|KKK}
\toprule
&Random seed                 & 1      & 2      & 3       \\
\cline{2-5}
&Young's modulus (kPa)       & 300000& 106971  & 117831  \\
&Poisson's ratio             & 0.4   & 0.33    & 0.03   \\
Cloud&Yield stress    (kPa)  & 13836 & 3907    & 10469 \\
slime&Material density (kg/m$^3$) & 1000  & 1000    & 1491    \\
&Table friction              & 1.18  & 0.13    & 0.54   \\
&Manipulator friction        & 2.0   & 0.01    & 0.58   \\
\midrule
&Random seed                 & 1      & 2      & 3       \\
\cline{2-5}
&Young's modulus (kPa)       & 10000  & 88963  & 10696  \\
&Poisson's ratio             & 0.32   & 0.33   & 0.08   \\
Soil&Yield stress    (kPa)   & 9333   & 1000   & 10037 \\
&Material density (kg/m$^3$) & 2000   & 1653   & 1823    \\
&Table friction              & 2.0    & 1.08   & 1.83   \\
&Manipulator friction        & 0.01   & 0.01   & 0.01   \\
\bottomrule
\end{tabular}
\caption{Physics parameters recovered by the DPSI method for Cloud slime and Soil materials with three random seeds.}
\label{tab:slime-soil-param}
\end{table}

\subsection{Cause and treatment to local minima}
\begin{figure}[t]
    \vspace{-2mm}
    \includegraphics[width=0.99\columnwidth, right]{note_images/legend.pdf}\\
    \includegraphics[width=0.44\columnwidth, right]{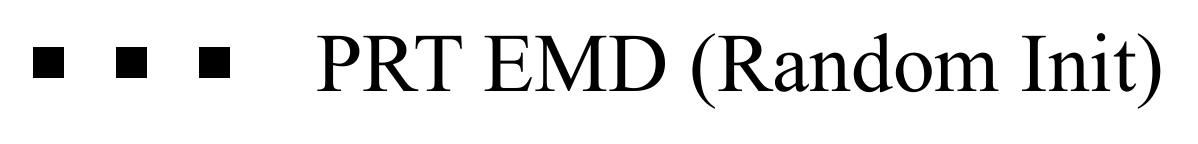}\\
    \includegraphics[width=0.99\columnwidth, right]{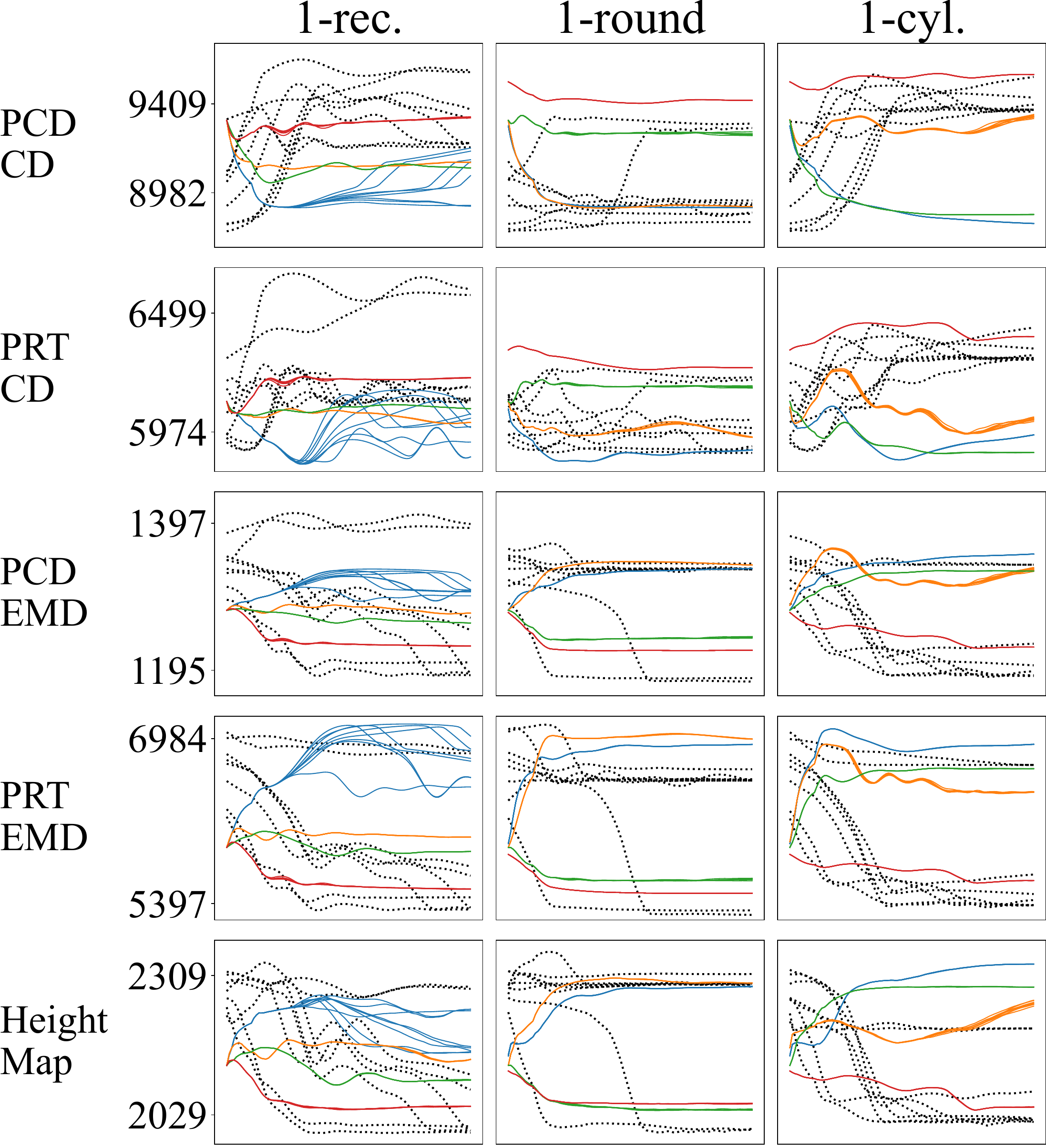}
    \includegraphics[width=0.99\columnwidth, right]{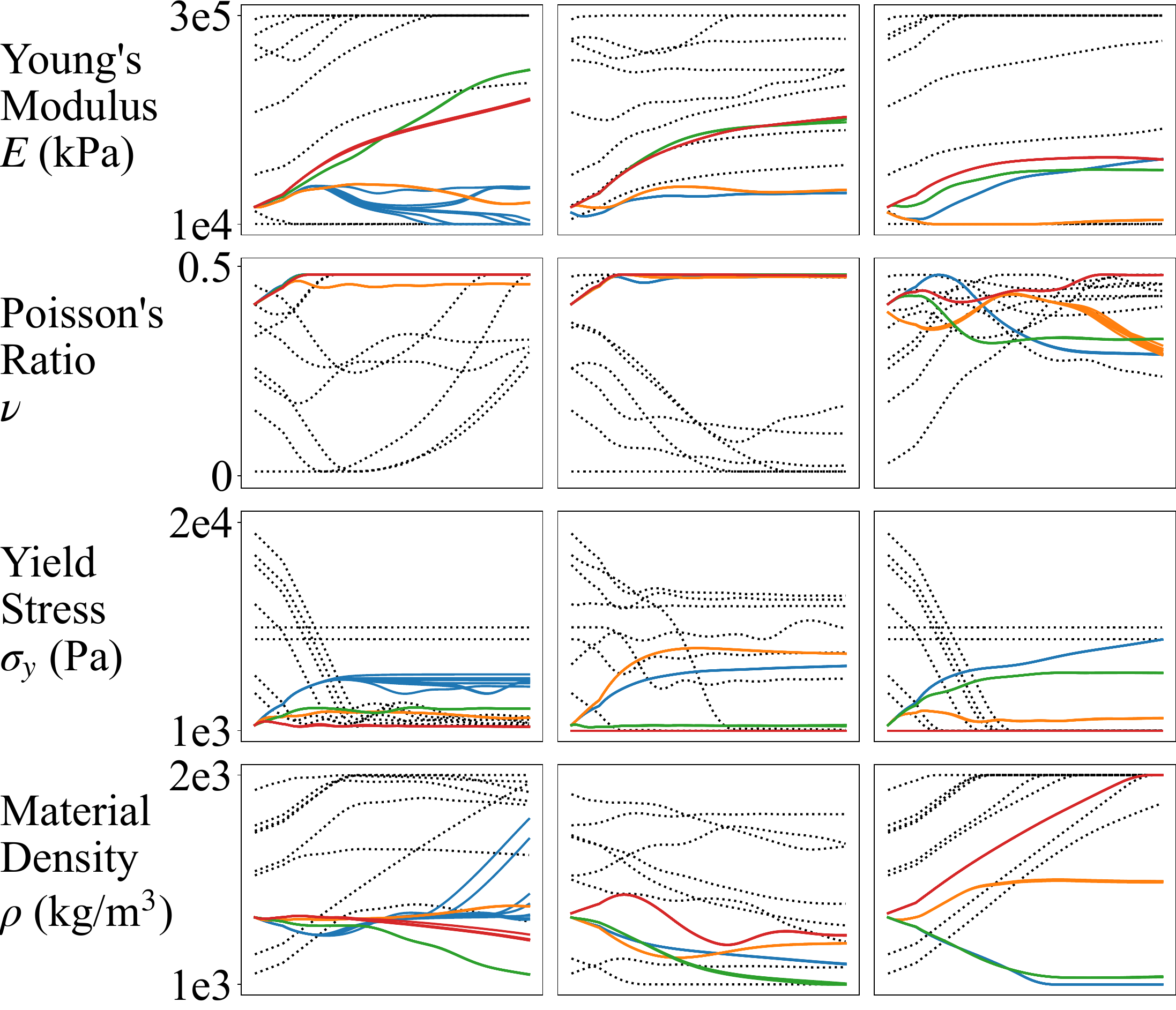}
    \caption{\small\xt{ Validation losses (top five rows) and parameter values (last four rows) over $100$ gradient updates at level-1 contact complexity from optimising the loss functions with an extra set of $8$ random seeds. Solid lines (highly overlapped) are results with the same initial solution while dotted lines show results of optimising the PRT EMD loss with randomised initial solutions. Each column presents the results of optimising with a different dataset. Each row shows the changes of an evaluation metric or a parameter, denoted on the left. In each figure, different colours indicate the results of optimising a different loss, as labelled by the legend on the top.}}
    \label{fig:extra-seeds}
\end{figure}

\xt{As shown by the loss landscapes and the optimisation results, although DPSI can find good solutions, it could get stuck in local minima. Therefore, we performed extra experiments to optimise the PRT EMD loss with the three 1-datapoint datasets at contact level 1 with eight more random seeds to examine the random seed sensitivity of DPSI. The results are shown by the black dotted lines in Figure~\ref{fig:extra-seeds}.}

\xt{First of all, these dotted lines show consistent results with the previous conclusions: they converge to stable (though different) loss and parameter values, still reveal the negative relationship between the CD and EMD losses, and can achieve comparable performances with results from larger datasets. On the other hand, it can be seen that the main cause of converging to different solutions is the different initial parameter values, which were determined by the pseudo-random process controlled by different random seed values. This is not surprising as gradient-descent algorithms with complex physics problems are known to be sensitive to the choice of the initial solution~\citep{hu2019difftaichi,Antonova2022RethinkingOW}.}

\xt{To further test the assumption, we perform the same experiments with the four loss functions using the same set of seeds but manually initialise the solutions using the hand-picked parameters in Table~\ref{tab:handpick-pa} selected according to the material/geotechnical engineering literature~\citep{rehman2018yield,waheed2023evaluation,StructX}. These hand-picked parameter values are less accurate in terms of validation loss as shown in Table~\ref{tab:best-param}. However, the (highly overlapped) solid lines in Figure~\ref{fig:extra-seeds} show that with this initial solution, DPSI converges with near-zero variances across eight random seeds, except for optimising the PCD CD loss with the 1-rec. dataset. This confirms that the differentiation process of the DPSI system is largely deterministic and the found solutions rely largely on the initial solutions, the optimisation algorithm and the loss functions. Moreover, with a good guess of the initial values, DPSI could converge very quickly (within $20$ iterations shown by the red/green solid lines).}

\begin{table}[t]
\centering
\footnotesize
\begin{tabular}{llllll}
\toprule
$E$ (kPa) & $\nu$ & $\rho$ (kg/m$^3$) & $\pmb{\sigma}_y$ (kPa) & $\eta_t$ & $\eta_m$\\
\midrule
$30000$ & $0.4$ & $1330$ & $1000$ & $0.7$ & $0.7$ \\
\bottomrule
\end{tabular}
\caption{\xt{Hand-picked physics parameter values based on the loss landscapes, published literature and open-source website in geo-technical engineering~\citep{rehman2018yield,waheed2023evaluation,StructX}. The yield stress value is slightly higher than the reported values because the reported values were found to cause the material collapse in simulation.}}
\label{tab:handpick-pa}
\end{table}

\begin{figure}[b]
\centering
\includegraphics[width=0.75\columnwidth]{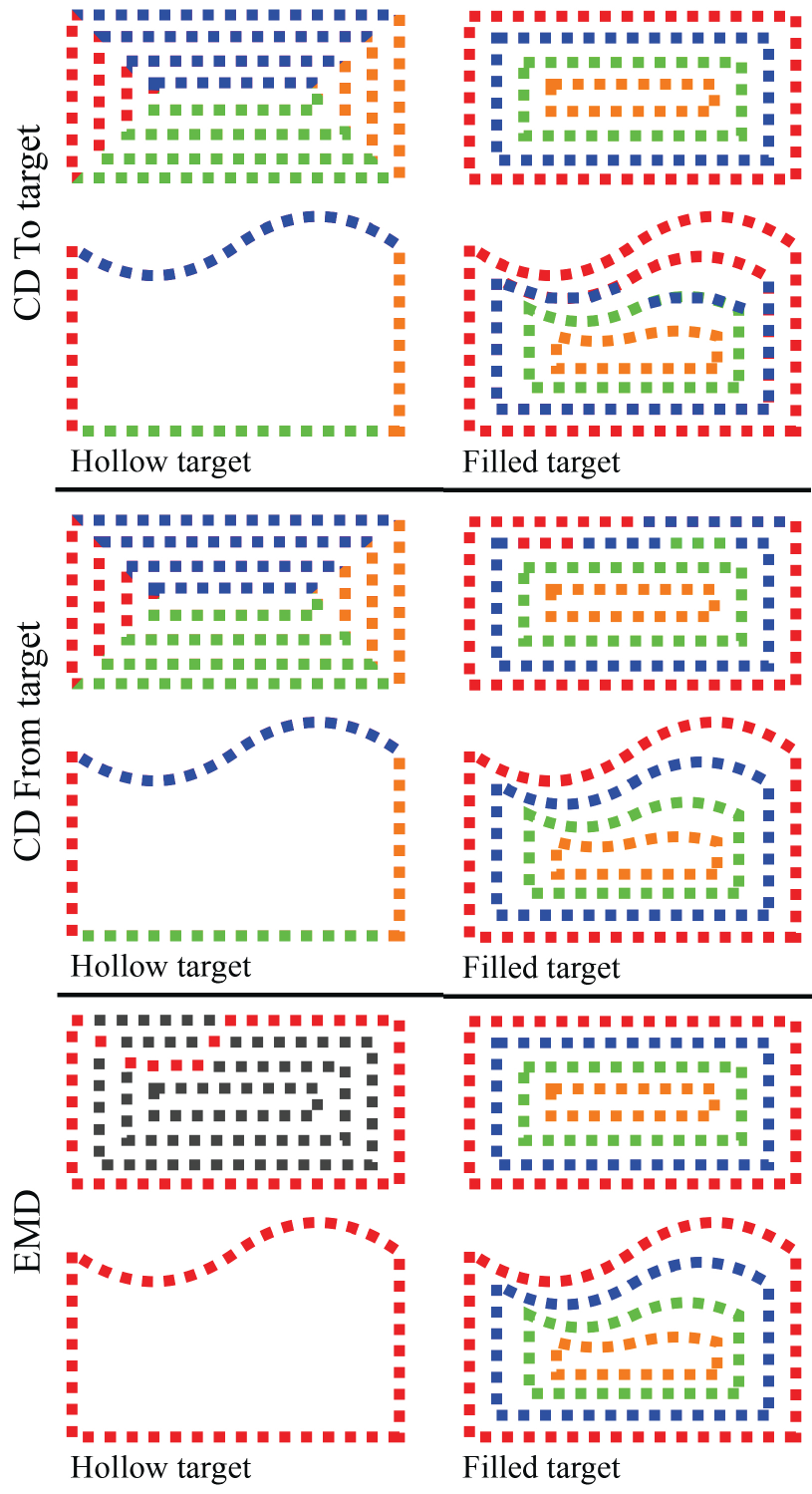}
\caption{Correspondence examples for the CD and EMD computations with hollow and filled point sets as targets. Colours indicate corresponding points. Note the CD loss is directional.}
\label{fig:cd-emd-points}
\end{figure}

\subsection{Remark on CD \& EMD losses}
\xt{As mentioned previously, the CD and EMD losses seem to focus on different spatial aspects of the object's geometry, leading to deformations that match either the horizontal or the vertical directions. With the results of optimising the four losses with the same initial solutions shown in Figure~\ref{fig:extra-seeds}, we can analyse the differences between CD and EMD losses with hollow or filled point sets.}

\xt{\textbf{Hollow or filled point sets}: For the CD losses, it can be seen by comparing the blue and orange solid lines in Figure~\ref{fig:extra-seeds}. The blue solids in the first two rows indicate that optimising the CD loss with a hollow point cloud target can achieve smaller losses in both CD loss metrics. However, the third to fifth rows in Figure~\ref{fig:extra-seeds} show that optimising CD with a filled particle system target (PRT CD loss) outperforms PCD CD in terms of both EMD loss metrics and the height map distance. On the other hand, optimising the EMD loss with a filled particle system target (red lines) exhibits the same pattern, consistently showing inferior performances in terms of the CD loss metrics while outperforming the PCD EMD losses (green lines) in terms of both EMD loss metrics and the height map distance. Therefore, the results indicate that optimising with a filled particle system target generated from a reconstructed watertight mesh is better than optimising with a hollow point cloud target.}

\xt{The cause behind it is the inaccurate point-to-point correspondences generated with a hollow point set. As shown by Figure~\ref{fig:cd-emd-points}, a hollow point set either averages the spatial information of an area of particles over a thin layer of target points in computing the CD loss or ignores the information of a large body of particles within the particle body in computing the EMD loss. On the contrary, with a non-hollow target, the point correspondences connect spatially closer points and do not disregard information within the object body.}

\xt{\textbf{CD or EMD:} Figure~\ref{fig:extra-seeds} shows that either with a hollow or non-hollow target, the EMD losses (green and red lines) consistently outperform the CD losses (blue and orange lines) in terms of the EMD and height map distance metrics while being inferior in terms of the CD metrics. Again, we believe that this is another consequence of the different ways CD and EMD compute the point correspondences.} 

\xt{Figure~\ref{fig:cd-emd-points} shows that, with a hollow target, CD distributes the spatial information of the points within over the surface points, which is problematic. Take the shape in Figure~\ref{fig:cd-emd-points} as an example, the particles above the target surface will be forced down, while the particles below the target surface will be pushed up, which is not supposed to happen. On the other hand, the EMD loss does not suffer from this effect because it corresponds the particles to the closest points from the target surface only. With a non-hollow target point set, the CD loss produces different correspondences in the two computation directions, which results in a mixture of correct and wrong spatial relations, while the EMD loss is more likely to find the best match spatially due to the one-to-one correspondence generated by a linear assignment algorithm.}

\xt{In short, this analysis leads to the conclusion that the PRT EMD loss, which computes point correspondences between the filled simulation particle system and a non-hollow target point cloud, is the most ideal choice for capturing the correct spatial differences between point sets.}

\subsection{Remark on computation cost}
\xt{In this subsection, we report the running time and GPU memory needed for forward and backward computations for the DPSI framework with different particle densities and motions. The statistics are summarised in Table~\ref{tab:com-time}, which were obtained using a Nvidia GeForce RTX 4090 GPU.}

\xt{It can be seen that the computation time and memory increase as the number of simulation particles and the length of motion increase. Because of the pre-compilation of kernels when they are first called in a TaiChi programme, the first forward and backward passes always take much longer and the later repetitive calls are much faster, as denoted by "CP. Foward time (s)".}

\xt{With particle numbers around $2000$ to $4000$, a DPSI optimisation run that consists of $100$ iterations of forward-backward computations using $1$ datapoint can be done in about $10$ minutes. As shown by Figures~\ref{fig:loss-param-curves-lv1} and~\ref{fig:loss-param-curves-lv2}, most runs converge within $50$ iterations, which means that a plausible solution can be found by DPSI in about $5$ minutes. With a good initial guess of the parameters, as shown by the dotted lines in Figure~\ref{fig:extra-seeds}, this runtime can be further reduced to around $2$ minutes. In addition, the memory consumption for these simulations is within $3$ GB, which permits implementations on portable and small GPUs. With no more than $2$ minutes for the 3D reconstruction pipeline, a parameter identification process could take less than $5$ minutes to complete. These statistics confirm that DPSI meets the criteria in computational costs for practical deployment. It also suggests again that it is practical to perform parameter identification with DPSI at any time without the need to rely on previously found parameters or slow manual measuring techniques.}

\begin{table}[b]
\footnotesize
\centering
\begin{tabular}{Gl|III}
\toprule
Motion & Particle density & $4\text{x}10^7$ & $6\text{x}10^7$ & $8\text{x}10^7$\\
\midrule
&Number of particles      & 1946            & 2808            & 3926\\
&CP. Forward time (s)     & 9.56            & 9.83            & 9.23\\
&Forward time (s)         & 1.45            & 1.61            & 2.17\\
Poking-1&Forward GPU (GB) & $\sim$1.6       & $\sim$1.8       & $\sim$1.9\\ 
&CP. Backward time (s)    & 24.69           & 24.71           & 24.42\\
&Backward time (s)        & 2.27            & 2.39            & 2.56\\
&Backward GPU (GB)        & $\sim$1.7       & $\sim$1.8       & $\sim$1.9\\
&$100$ FB-Iterations (min)& 6.71          & 7.18            & 8.37\\
\midrule
&Number of particles      & 3880            & 5836            & 7738\\
&CP. Forward time (s)     & 11.09           & 14.63           & 23.97\\
Poking-&Forward time (s)  & 2.87            & 7.06            & 16.78\\
shifting-1&Forward GPU (GB)& $\sim$1.9      & $\sim$2.2       & $\sim$2.5\\
&CP. Backward time (s)    & 26.15           & 27.67           & 27.34\\
&Backward time (s)        & 4.31            & 5.42            & 5.79\\
&Backward GPU (GB)        & $\sim$1.9       & $\sim$2.2       & $\sim$2.5\\
&$100$ FB-iterations (min)& 12.47           & 21.30           & 38.10\\
\midrule
&Number of particles      & 3059            & 4688            & 6271\\
&CP. Forward time (s)     & 14.02           & 15.19           & 15.94\\
&Forward time (s)         & 6.08            & 6.54            & 7.88\\
Flattening&Forward GPU (GB) & $\sim$1.8     & $\sim$1.9       & $\sim$2.1\\ 
&CP. Backward time (s)    & 31.95           & 34.83           & 34.93\\
&Backward time (s)        & 10.14           & 12.29           & 13.03\\
&Backward GPU (GB)        & $\sim$1.8       & $\sim$1.9       & $\sim$2.1\\
&$100$ FB-iterations (min)& 27.53           & 31.90           & 35.35\\
\bottomrule
\end{tabular}
\caption{\xt{Computational costs (time and GPU memory consumption) of the TaiChi-based differentiable simulator used in the DPSI framework with a Nvidia GeForce RX 4090 GPU. The costs of simulations with three motions and particle densities are reported. CP.: compile-time costs. The first time a program is run takes extra time to compile TaiChi kernels. $100$ FB-iterations (min): the runtime of $100$ forward and backward computations in minutes.}}
\label{tab:com-time}
\end{table}

\section{Conclusion}
This work addresses the important problem of closing the gap between simulated and real-world manipulations of elastoplastic objects. In particular, we propose a differentiable physics-based system identification framework (DPSI) that can identify physics parameters through gradient descent algorithms. Substantial experiments demonstrate that the proposed framework can identify parameters that reproduce quantitatively and qualitatively realistic elastoplastic object manipulation dynamics in the presence of real-world perceptual and data collection challenges. These challenges include 1) simple and short motions, 2) incomplete trajectories, 3) object occlusions, 4) point cloud noises, and 5) small data. The proposed framework is the first example of system identification with differentiable physics-based, particle-based simulation for robotic volumetric elastoplastic object manipulation. It serves as the foundation for faster and more accurate real-world deployments of deformable object manipulation.

In addition, with the use of physics-based dynamics models, the identified parameters are physically meaningful. The experiments reveal that different parameter values found through optimisation can be interpreted in a way that is aligned with our understanding and intuitions about real-world physics. Therefore, the DPSI framework not only gives users confidence in the simulation controlled by the physics models and these parameters but also provides users with intuitive angles to identify the limitations of the reconstructed manipulation dynamics.

\subsection{Limitations and future research}
Several limitations can be observed from the experiments conducted in this work. The most obvious would be the under-representativeness of the physics models that describe the underlying manipulation dynamics. As discussed in the Results section, there is a lack of elastic returning effect at the areas of contact with sharp edges of objects, the insufficient spread of force impact that causes the deformation to stay near the local contact area, and the artefacts of floating displaced particles. Though these inaccuracies may be negligible for coarse manipulations that appear more often in our daily lives, they would lead to unacceptable solutions for high-precision manipulation tasks such as surgery, soft object assembly, etc. In addition, various approximations in the MLS-MPM and programming implementations also contribute to the simulation inaccuracy. As such, future research is needed to develop more accurate physics models, numerical approximations, and coding implementations to simulate object deformations. 

\xt{Secondly, an important assumption of this work is that the deformation behaviour of the target material in need of parameter identification can be simulated by the selected elastoplasticity model. Thus, it should be noted that when the target material's behaviour is largely underrepresented by the selected physics model, DPSI would not be able to produce very realistic simulations. In other words, the manual selection of the elastoplasticity model is of high importance for the proposed DPSI framework. For example, our results with soil reveal that DPSI found parameters to make simulations appear as close to the real materials as possible, but the use of the von Mises plasticity model will make it difficult to recover parameter values that accurately reflect realistic soil or foam behaviours, which are better captured by the Drucker–Prager model.}

\xt{Thirdly, an important lesson learnt from the results is that increased complexity of motion and contact mode leads to fewer local minima and better system identification accuracy. The number of local minima indicates that the collected data is insufficient to fully induce the correct values of the concerned parameters. Thus, in the future, we envision a better framework that incorporates learning-based approaches to allow the automatic selection of diverse tools and interaction motions to achieve more efficient and accurate parameter identification.}

Another limitation comes from the means of capturing real-world observations. In this work, we employ the multi-view fusion point clouds as the observation space, which suffers from noise and the inaccurate estimate of the camera extrinsic matrix. In the future, new methods should be developed to reduce point cloud noises and improve the precision of camera calibration. Lastly, the simulated scene in this work is limited to an end-effector, a table surface, and a target elastoplastic object. Efforts are needed to extend the framework to support more realistic and complex contact dynamics. More experiments are also needed to examine the feasibility of large-scale simulation in terms of efficiency and accuracy.

\section*{ACKNOWLEDGMENT}
This work was supported by the Engineering and Physical Sciences Research Council (grant No. EP/X018962/1). The authors would like to thank Samuel Moeller (Engineer support staff at the School of Engineering, Cardiff University) and Dr Prasad Rayamane for their help with 3D printing The end-effectors.

\bibliographystyle{SageH}
\bibliography{ref}

\clearpage
\section{Appendix}
\section{A Optimisation results with height map loss}

\begin{figure}[t]
    \vspace{-3mm}
    \centering
    \includegraphics[width=0.95\columnwidth]{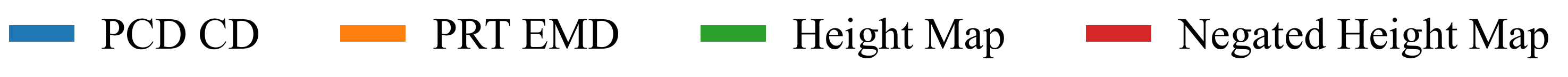}
    \begin{subfigure}{\columnwidth}
        \vspace{-2mm}
        \caption{Optimisation losses}
        \vspace{-1mm}
        \includegraphics[width=0.9\columnwidth]{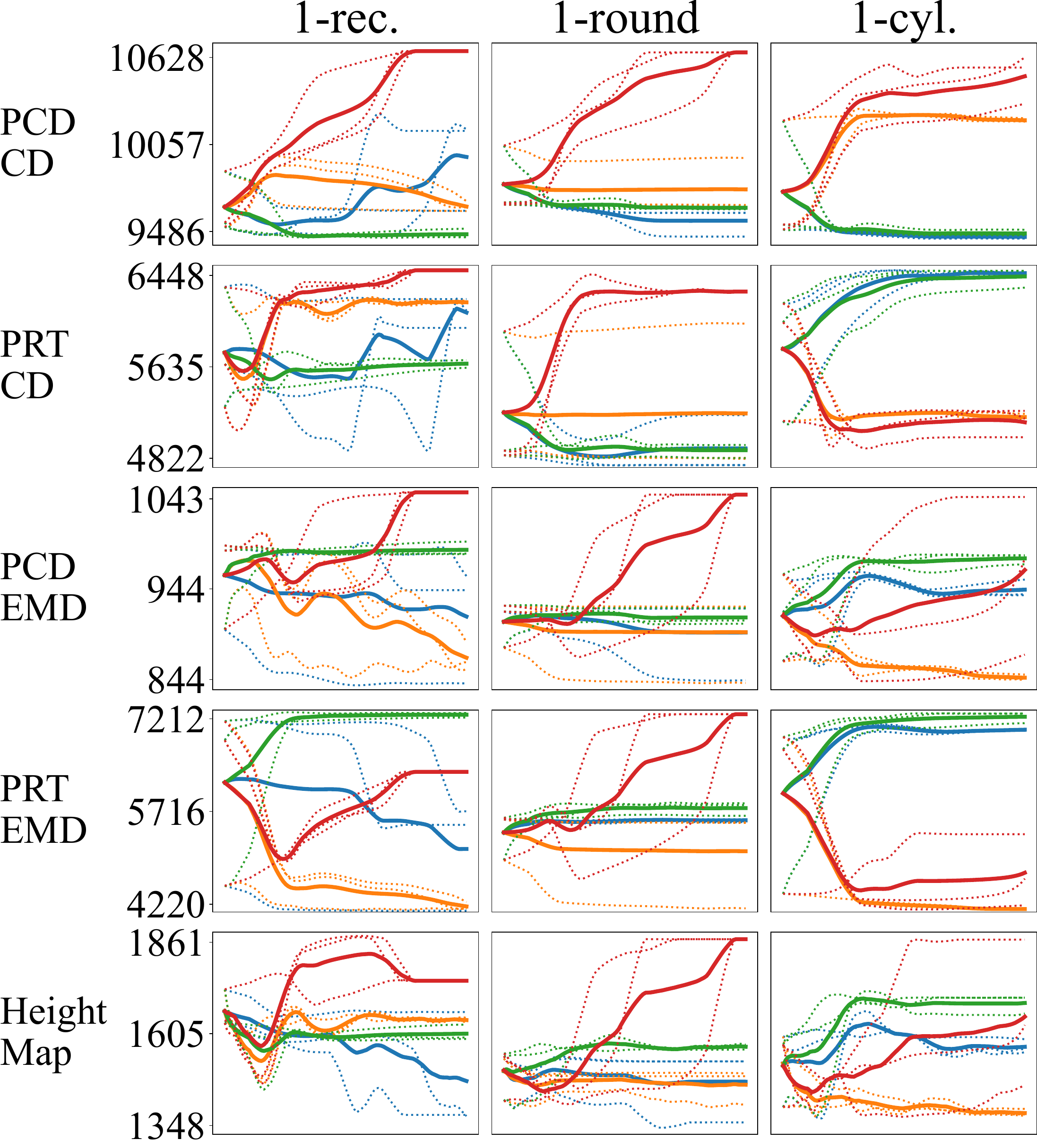}
    \end{subfigure}
    \begin{subfigure}{\columnwidth}
        \vspace{-2mm}
        \caption{Validation losses}
        \vspace{-1mm}
        \includegraphics[width=0.9\columnwidth]{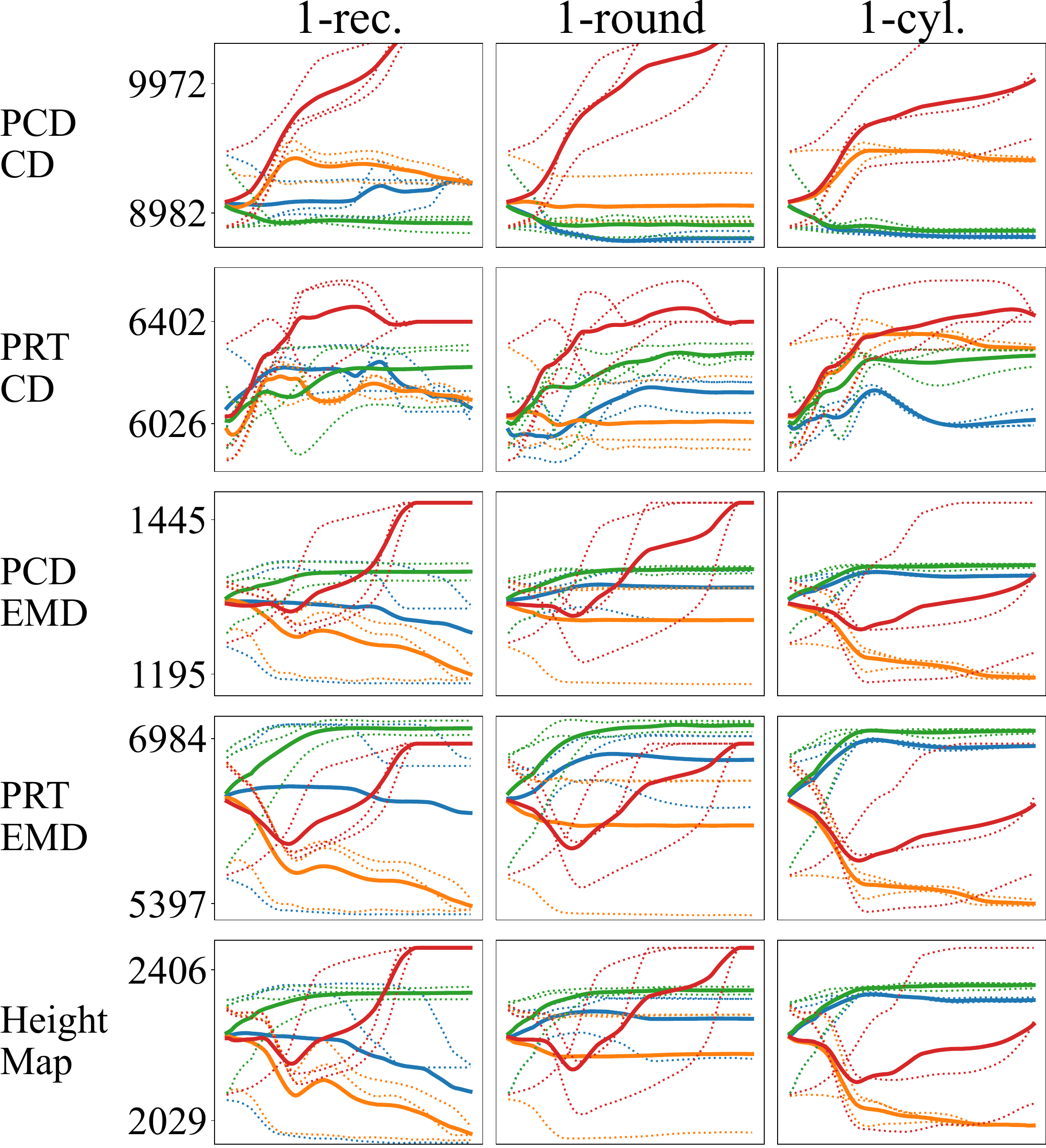}
    \end{subfigure}
    \vspace{-3mm}
    \caption{\small Optimisation and validation losses over $100$ gradient updates at level-1 contact complexity from optimising the PRT CD, PRT EMD, Height Map and Negated Height Map loss functions. Each row shows the changes of an evaluation metric or a parameter, denoted on the left. In each figure, different colours indicate the results of minimising a different loss function, as labelled by the legend on the top. Each colour has three dotted lines corresponding to the results of three random seeds and a solid line corresponding to their means.}
    \label{fig:hm-loss}
\end{figure}

\xt{As mentioned in the main text, optimising the parameters using the height map loss results in unsatisfactory performances. Here we compare the optimisation and validation losses of minimising the height map loss with those of minimising the PCD CD and PRT EMD losses in Figure~\ref{fig:hm-loss}.}

\xt{The green curves show that minimising the height map loss results in an increase of the height map loss with both optimisation and validation datasets, which is counterintuitive. To verify that the counter effect is not due to mistaken loss computations, we further optimise the negated height map loss (by effectively using the negated gradients from the height map loss), which resulted in worse performances (red curves). In addition, the height map loss is found to perform worse than the PCD CD loss, which was found to be the worst in all point-based loss functions (see the blue curves in Figures~\ref{fig:loss-param-curves-lv1} and~\ref{fig:loss-param-curves-lv2}. Therefore, we discarded the results from the main article to reduce complications during the result analysis.}

\xt{The possible cause of this phenomenon could be the loss of spatial information and geometric precision during height map computations. While height maps appear to be visually more intuitive to humans, it is computed by collecting the highest $z$ values of the points within an area defined by the height map resolution. As such, the gradients from the height map loss are only relevant to the $z$ values of the highest points in each cell, having no connections with the points that are lower in each cell nor with the horizontal aspect of the selected points.}

\section{B Skill planning with the MPM simulator}

\xt{To demonstrate the feasibility of planning with the MPM simulator, we create a simple greedy exhaustive search algorithm that selects $10$ actions to manipulate a piece of clay to match a target height map using the cylinder end-effector. The process is summarised in Algorithm 1.} 

\begin{figure}[t]
\vspace{-3mm}
\centering
\includegraphics[width=\columnwidth]{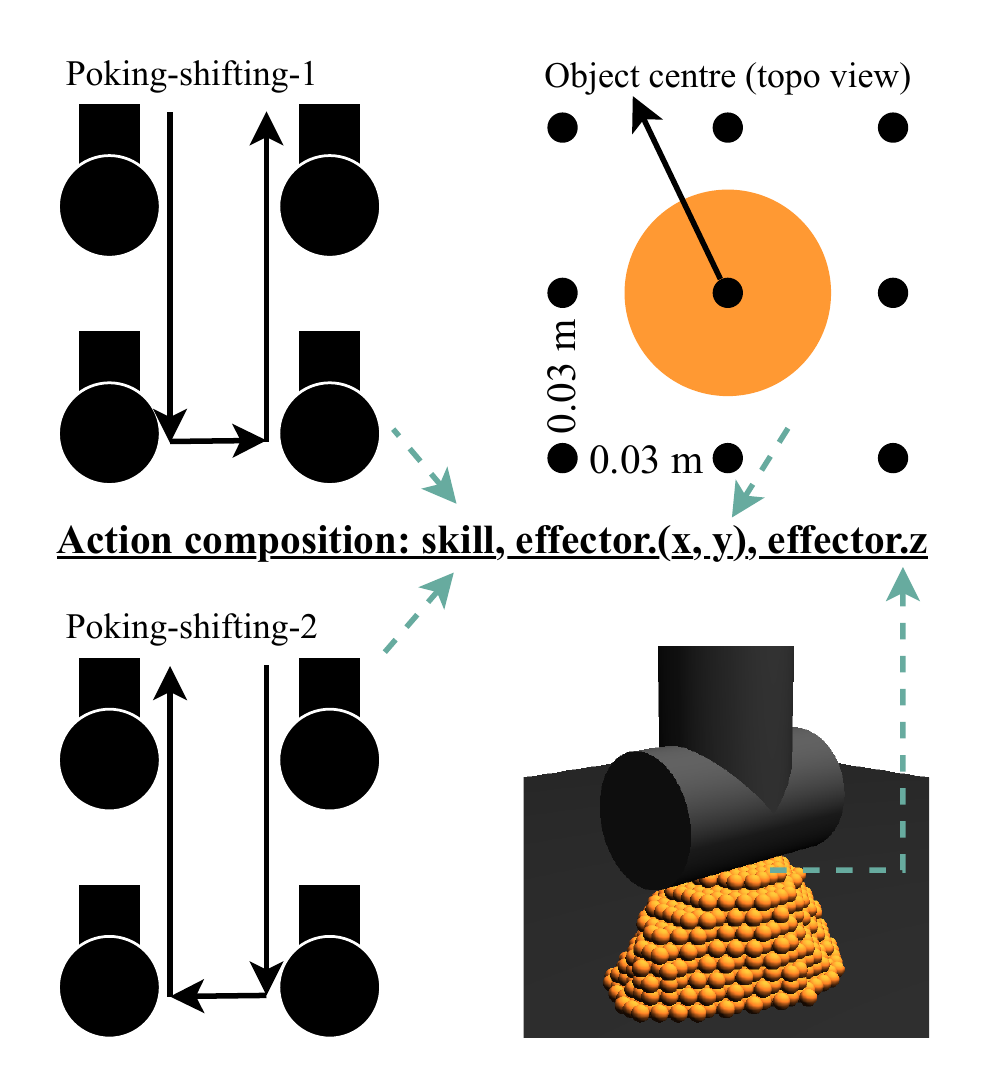}
\vspace{-6mm}
\caption{Action composition for the skill planning algorithm with the cylinder end-effector. An action applies one of the two poking-shifting motions at one of the nine neighbouring locations surrounding the object from the highest $z$ value of the object.}
\label{fig:a-com}
\end{figure}

\xt{In short, the algorithm starts with an initial object configuration in the simulator and performs exhaustive searches to find $8$ consecutive actions that result in the lowest height map distance from the target height map. An action effectively applies one of the two poking-shifting motions at one of the nine locations surrounding the object. As shown by Figure~\ref{fig:a-com}, the neighbouring locations are $0.03$ meters away from each other in the $x$ and $y$ directions. The height from which the skill motion starts is set to the highest $z$ value of the object before each action. For each action, the algorithm queries the simulator for the height map loss values of the manipulation results of applying the two poking-shifting motions at the nine locations (hence $18$ forward simulations). An empty action that does not move the end-effector is also considered in each search. The action with the lowest loss is selected and the algorithm proceeds to the next action search with a new object state from the last action. The algorithm returns a plan of $8$ actions.}

\begin{figure}[t]
\centering
\includegraphics[width=0.835\columnwidth]{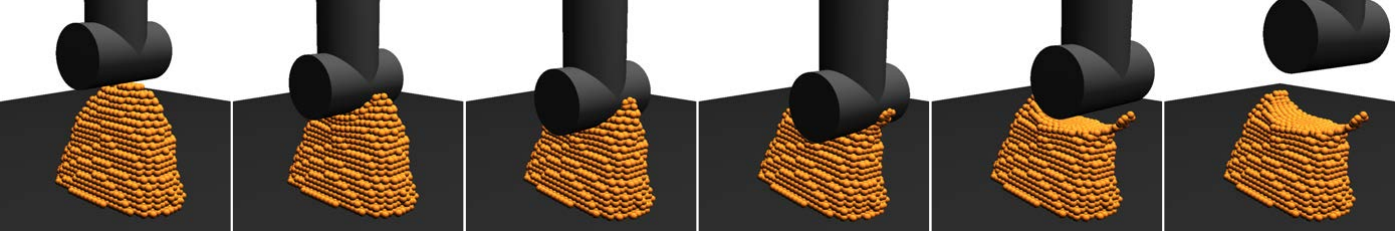}
\includegraphics[width=0.155\columnwidth]{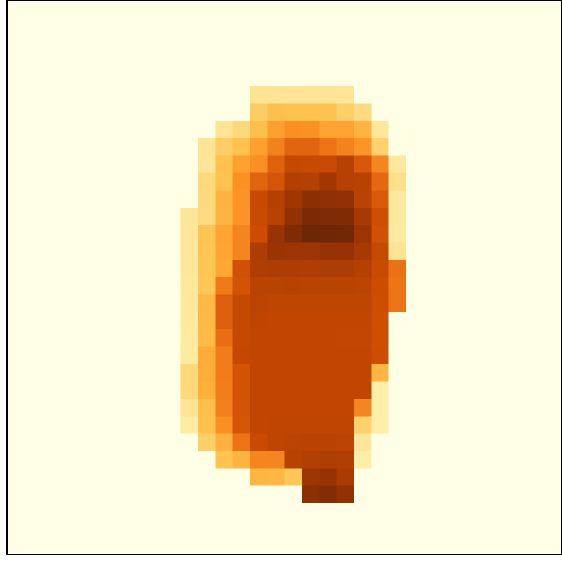}\\
\includegraphics[width=0.835\columnwidth]{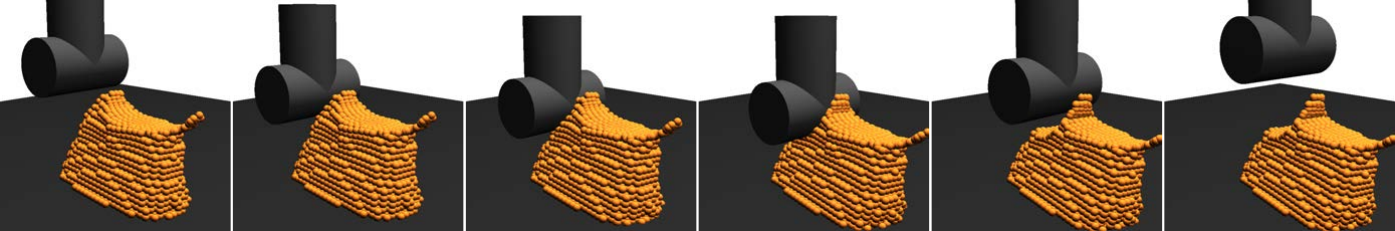}
\includegraphics[width=0.155\columnwidth]{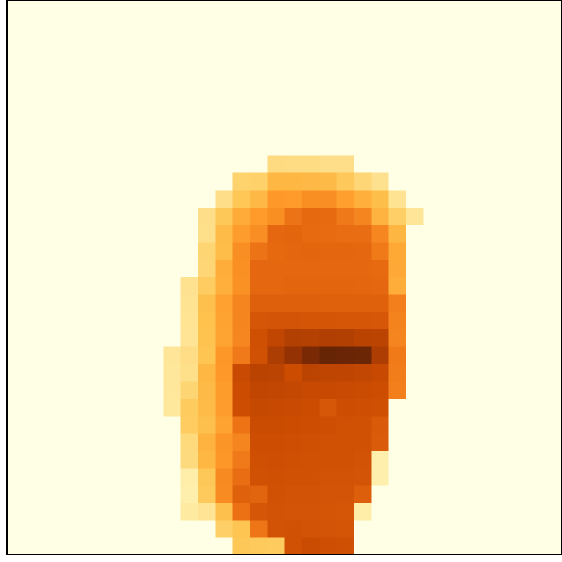}\\
\includegraphics[width=0.835\columnwidth]{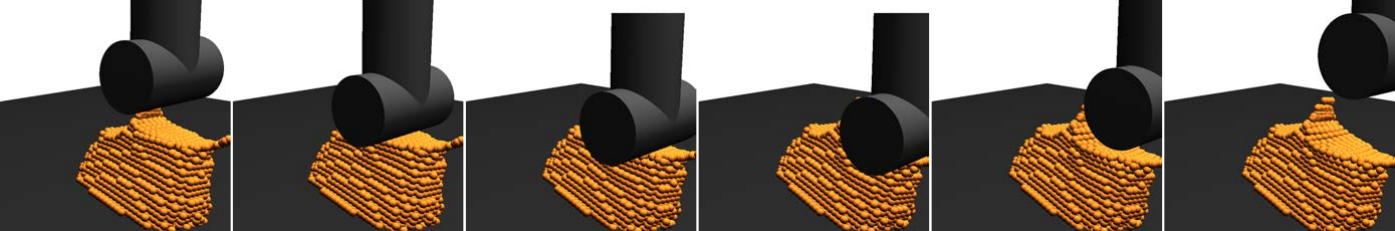}
\includegraphics[width=0.155\columnwidth]{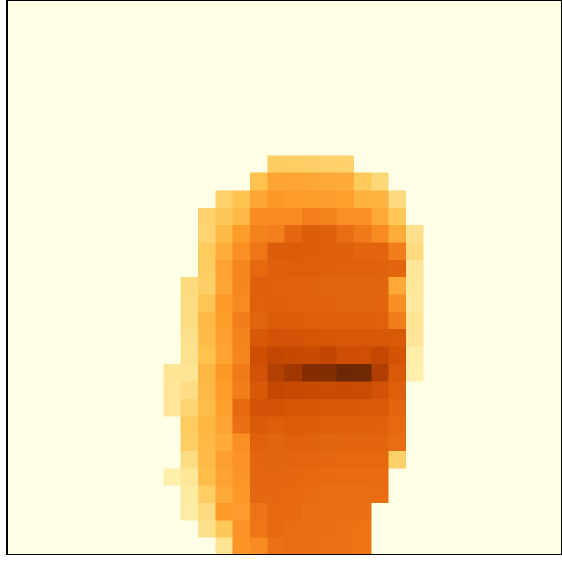}\\
\includegraphics[width=0.835\columnwidth]{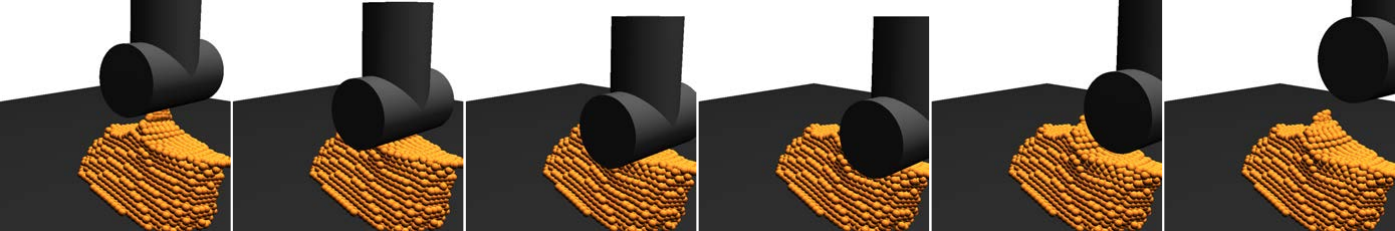}
\includegraphics[width=0.155\columnwidth]{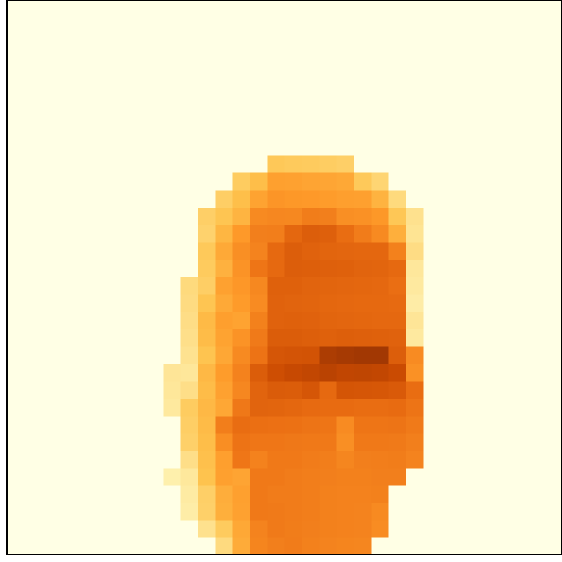}\\
\includegraphics[width=0.835\columnwidth]{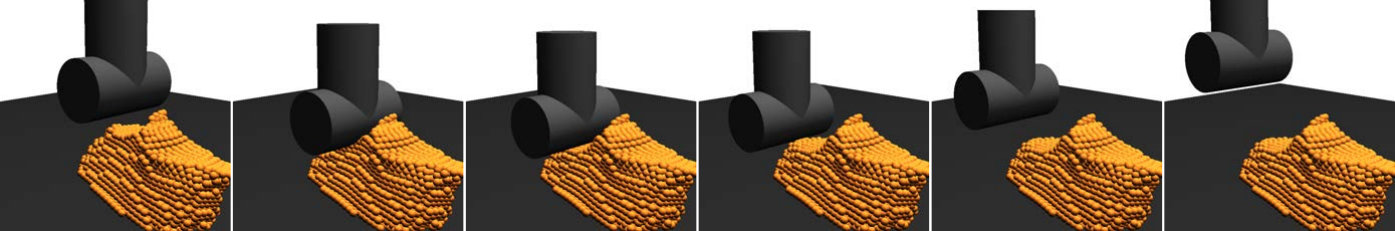}
\includegraphics[width=0.155\columnwidth]{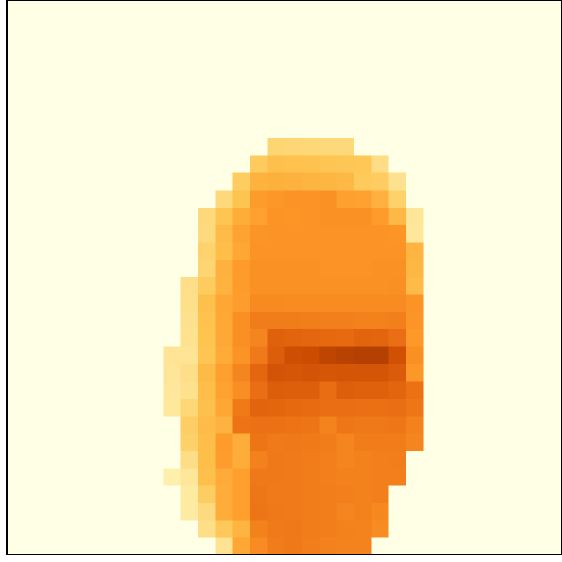}\\
\includegraphics[width=0.835\columnwidth]{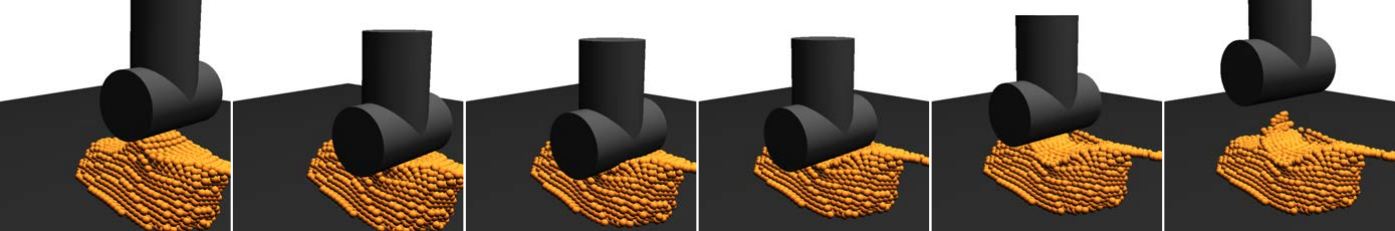}
\includegraphics[width=0.155\columnwidth]{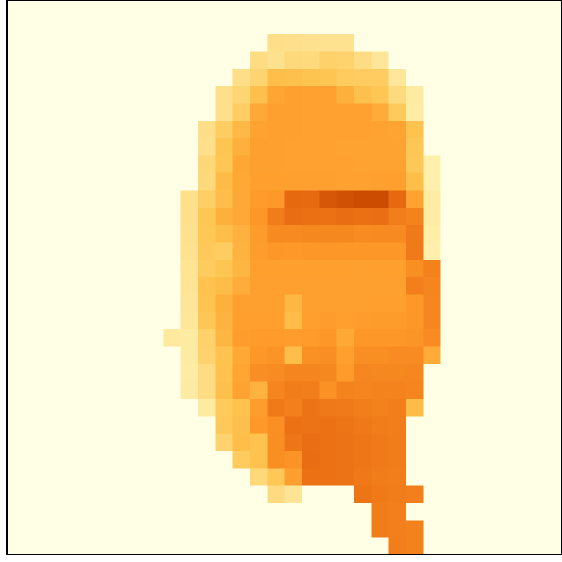}\\
\includegraphics[width=0.835\columnwidth]{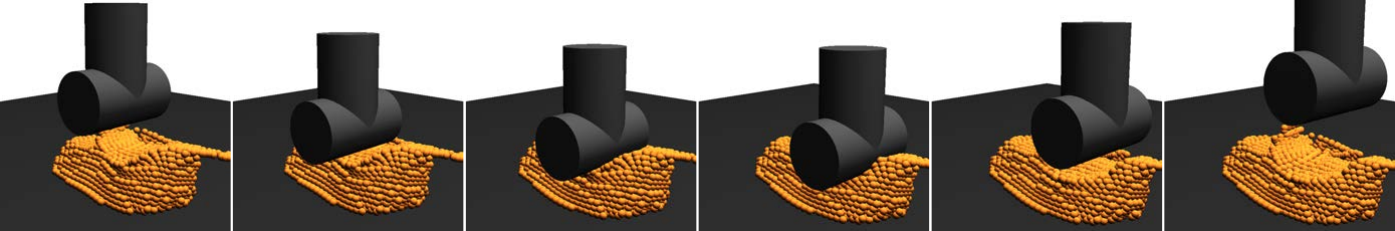}
\includegraphics[width=0.155\columnwidth]{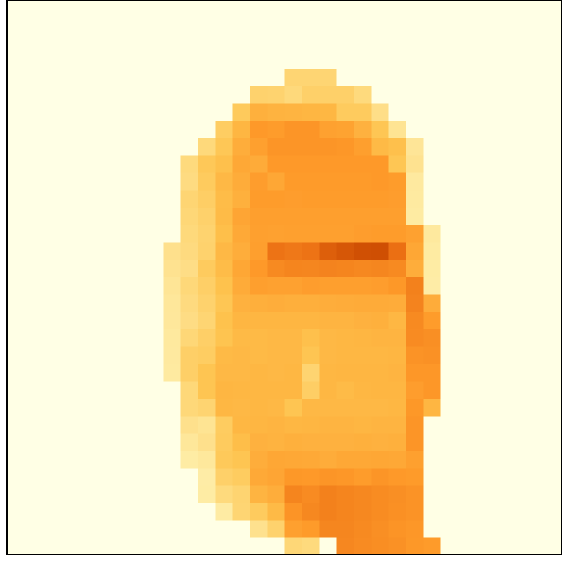}\\
\includegraphics[width=0.835\columnwidth]{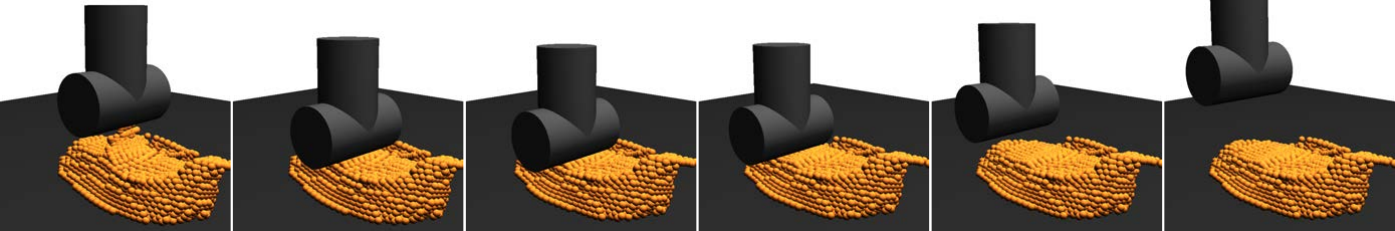}
\includegraphics[width=0.155\columnwidth]{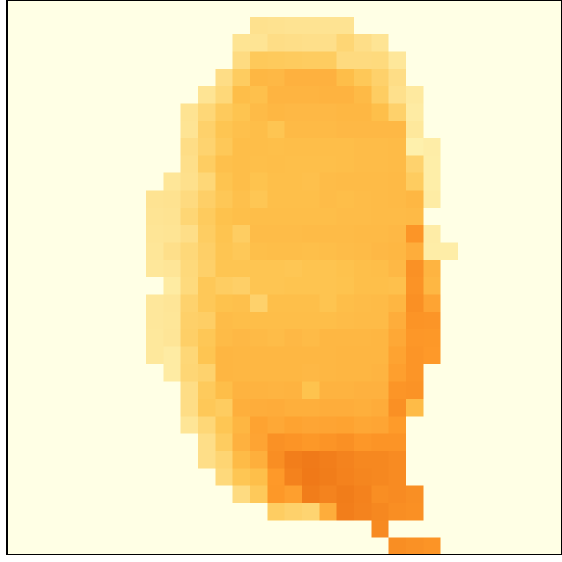}\\
\includegraphics[width=0.67\columnwidth]{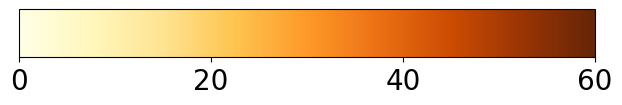}
\includegraphics[width=0.155\columnwidth]{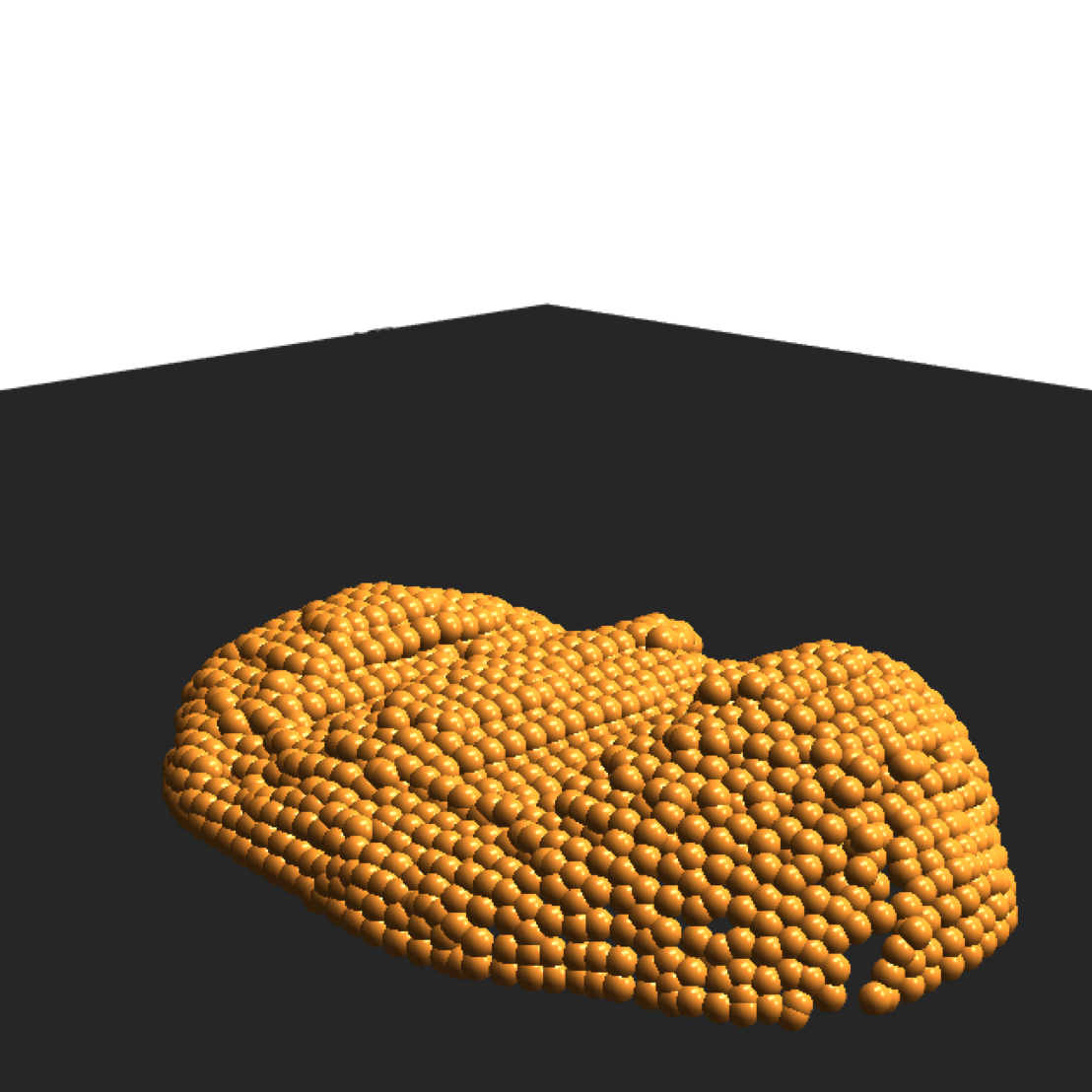}
\includegraphics[width=0.155\columnwidth]{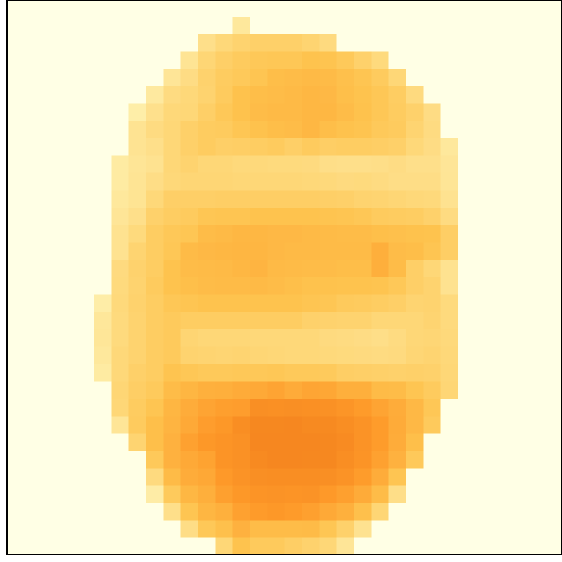}
\caption{Visualising a plan returned by the exhaustive search algorithm that takes $8$ actions to flatten a piece of clay to match a target height map (bottom right). All height maps are plotted under the same colour/value distribution shown by the colour bar.}
\label{fig:plan-0}
\end{figure}

\begin{figure}[t]
\centering
\includegraphics[width=0.835\columnwidth]{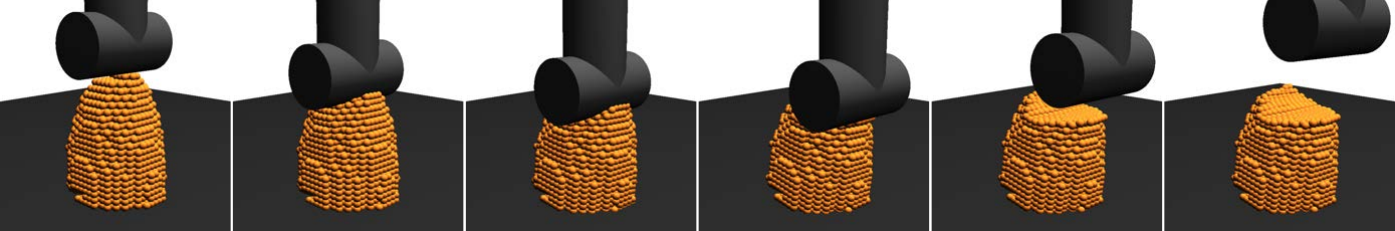}
\includegraphics[width=0.155\columnwidth]{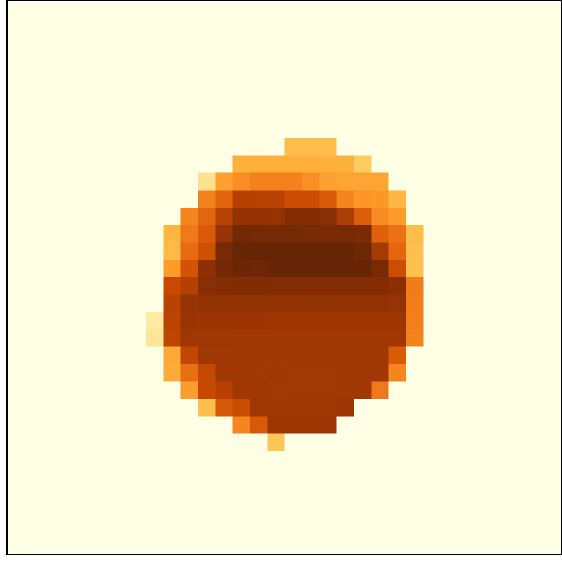}\\
\includegraphics[width=0.835\columnwidth]{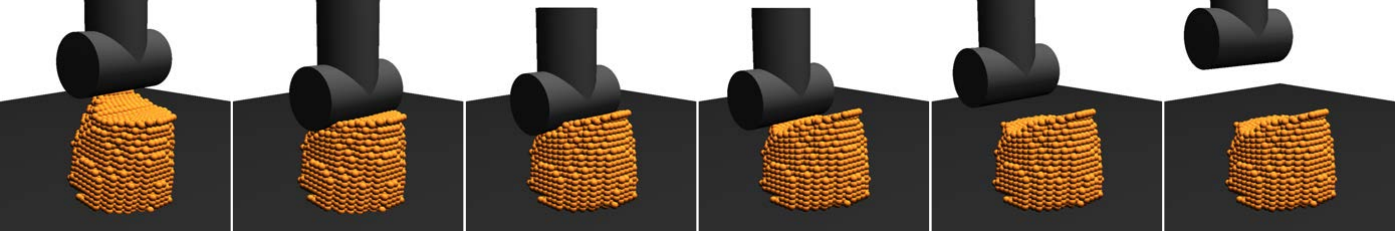}
\includegraphics[width=0.155\columnwidth]{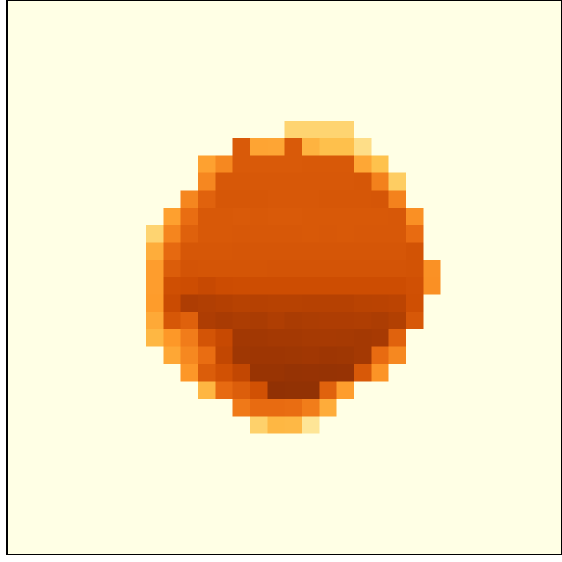}\\
\includegraphics[width=0.835\columnwidth]{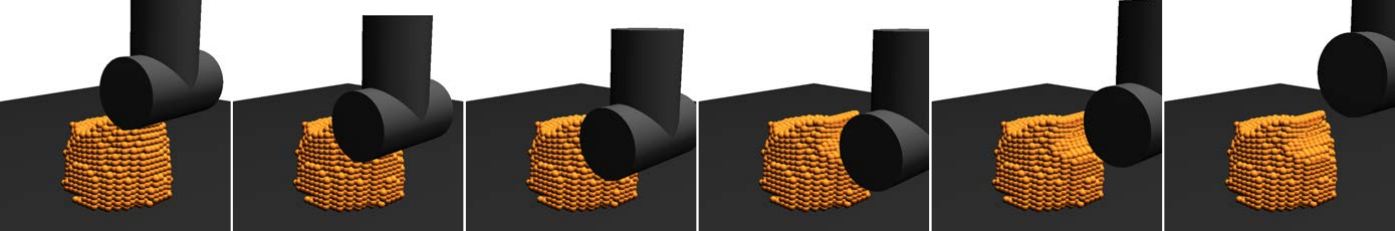}
\includegraphics[width=0.155\columnwidth]{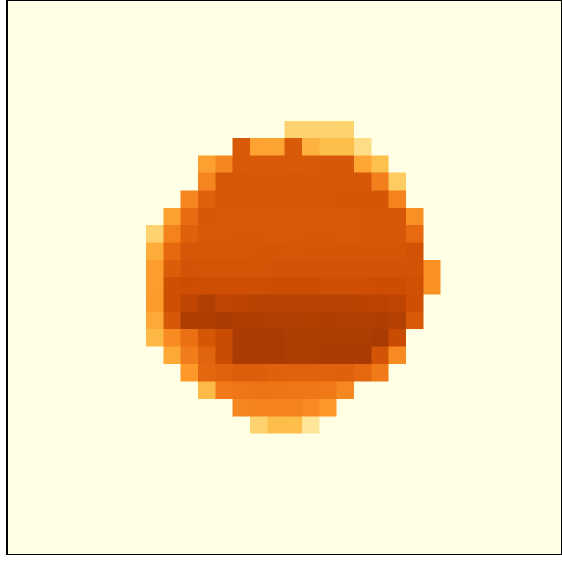}\\
\includegraphics[width=0.835\columnwidth]{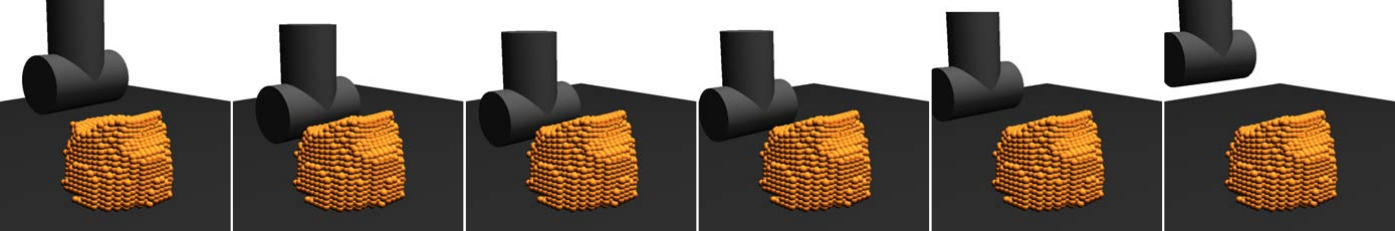}
\includegraphics[width=0.155\columnwidth]{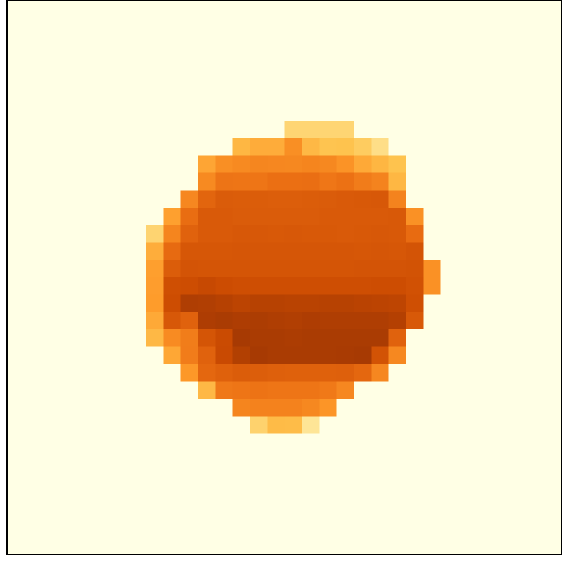}\\
\includegraphics[width=0.835\columnwidth]{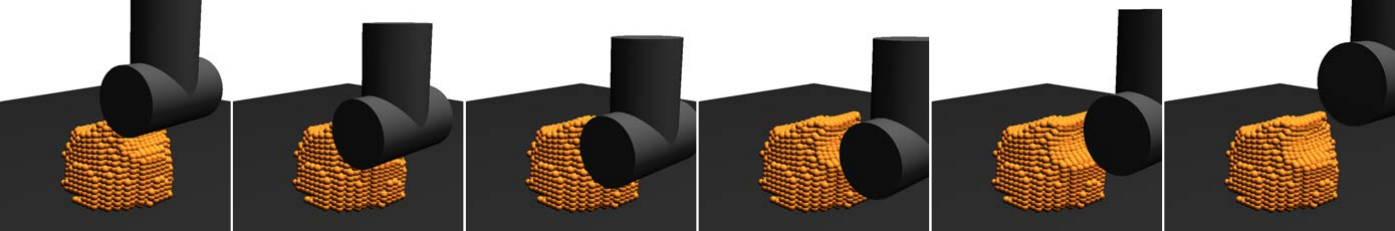}
\includegraphics[width=0.155\columnwidth]{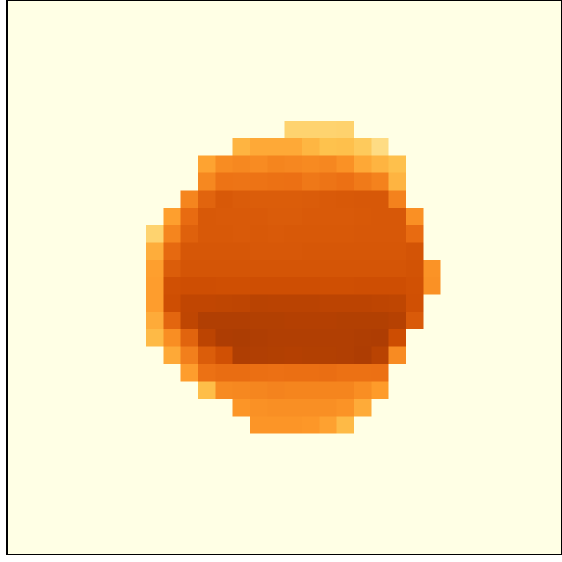}\\
\includegraphics[width=0.835\columnwidth]{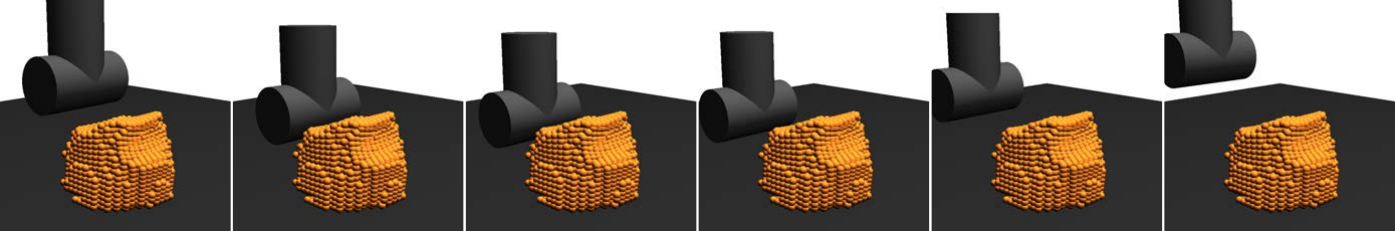}
\includegraphics[width=0.155\columnwidth]{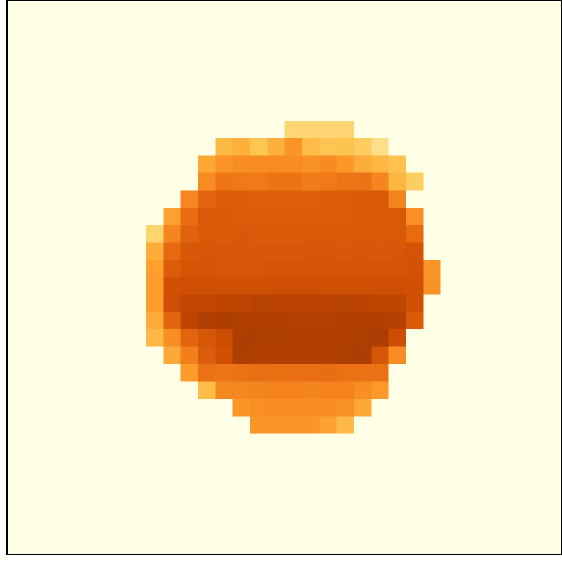}\\
\includegraphics[width=0.835\columnwidth]{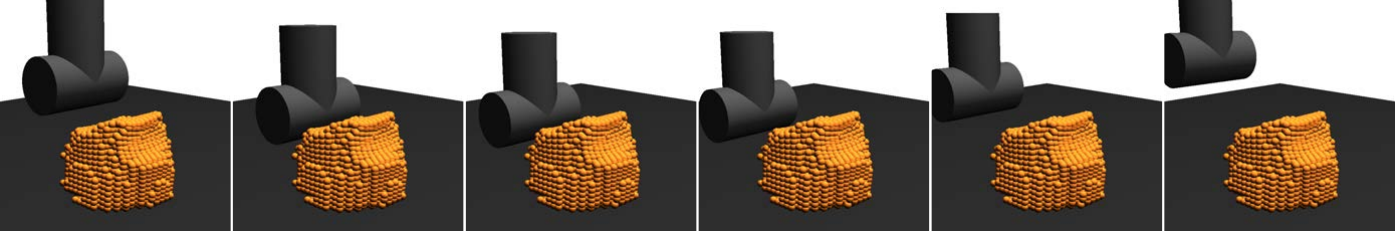}
\includegraphics[width=0.155\columnwidth]{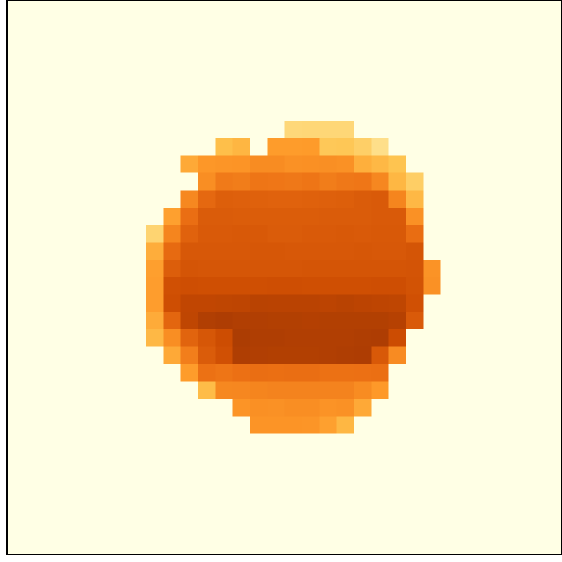}\\
\includegraphics[width=0.835\columnwidth]{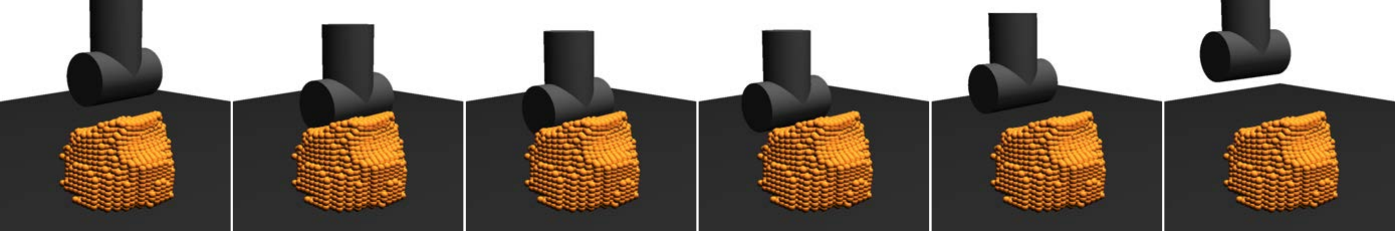}
\includegraphics[width=0.155\columnwidth]{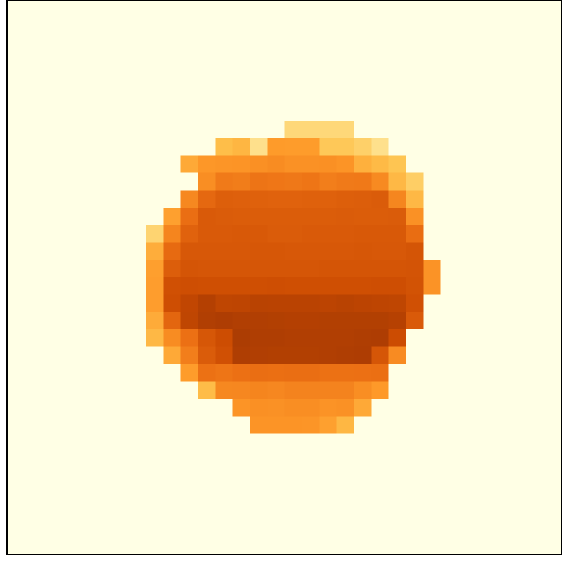}\\
\includegraphics[width=0.67\columnwidth]{note_images/planning/just_colorbar.png}
\includegraphics[width=0.155\columnwidth]{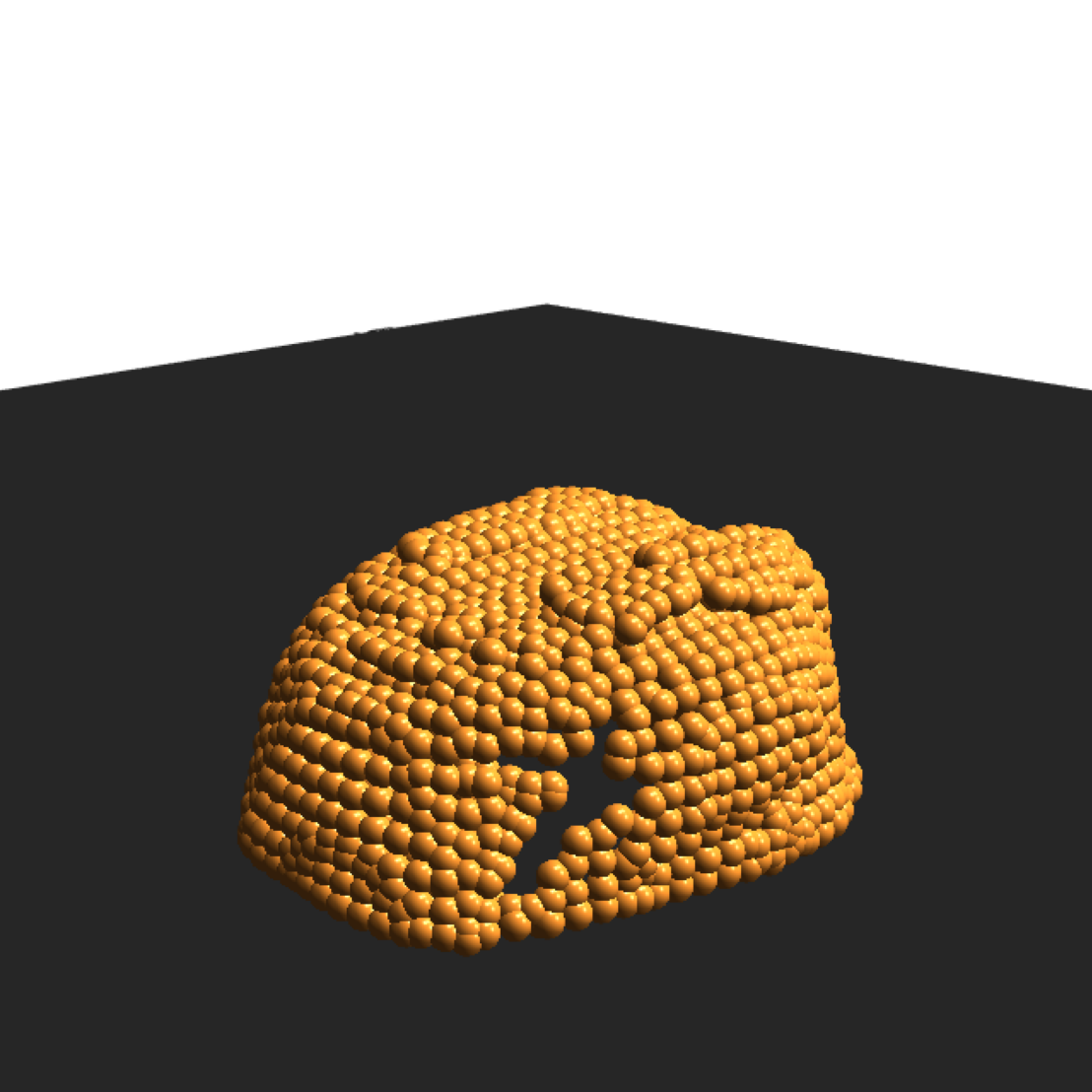}
\includegraphics[width=0.155\columnwidth]{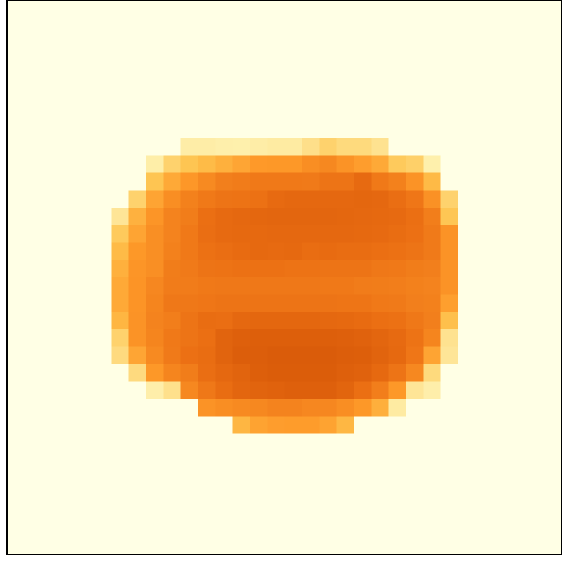}
\caption{Visualising a plan returned by the exhaustive search algorithm that takes $8$ actions to flatten a piece of clay to match a target height map (bottom right). All height maps are plotted under the same colour/value distribution shown by the colour bar.}
\label{fig:plan-1}
\end{figure}

\xt{Figures~\ref{fig:plan-0} and~\ref{fig:plan-1} visualise two plans returned by the algorithm that successfully manipulate the clays to a shape close to their respective target height maps. These examples showcase the feasibility of planning with the MPM simulator. Each plan takes roughly $7.5$ minutes to finish on a 4090 GPU, where each iteration (search over $19$ actions) takes roughly $0.9$ minute. This is not yet real-time but promising considering the high dimensionality of deformable object states.}

\begin{table}[t]
\small
\centering
\begin{tabular}{M}
\toprule
\textbf{Algorithm 1}: Greedy exhaustive search\\
\midrule
\texttt{Initialise} object particle system\\
\texttt{Initialise} MPM simulator: \textit{ENV}\\
\texttt{Load} target height map\\
\texttt{Load} DPSI-found parameters: \textit{ENV}.set\_parameters($\pmb{\theta}^{sim}$)\\
Effector\_init\_pos $\leftarrow$ \textit{ENV}.get\_object\_centre()\\ 
New\_state $\leftarrow$ \textit{ENV}.get\_state()\\
\textit{Plan} = []\\
\texttt{For} $n$ = $0$ to \textit{Number\_of\_ actions}:\\
\ $|$\ \ $\rhd$\textit{Evaluate all skills and start locations}\\
\ $|$\ \ Min\_loss $\leftarrow \infty$\\
\ $|$\ \ \texttt{For} $i$ = $1$ to $2$:\\
\ $|$\ \ $|$\ \ \texttt{For} $j$ = $-1$ to $1$:\\
\ $|$\ \ $|$\ \ $|$\ \ \texttt{For} $k$ = $-1$ to $1$:\\
\ $|$\ \ $|$\ \ $|$\ \ $|$\ \ Load skill trajectory: skill$_i$\\
\ $|$\ \ $|$\ \ $|$\ \ $|$\ \ Effector\_init\_pos' = Effector\_init\_pos.copy()\\
\ $|$\ \ $|$\ \ $|$\ \ $|$\ \ Effector\_init\_pos'.x $+= 0.3\times j$\\
\ $|$\ \ $|$\ \ $|$\ \ $|$\ \ Effector\_init\_pos'.y $+= 0.3\times k$\\
\ $|$\ \ $|$\ \ $|$\ \ $|$\ \ Effector\_init\_pos'.z $=$ \textit{ENV}.get\_object\_height()\\
\ $|$\ \ $|$\ \ $|$\ \ $|$\ \ \textit{ENV}.set\_state(New\_state)\\
\ $|$\ \ $|$\ \ $|$\ \ $|$\ \ \textit{ENV}.set\_agent\_pos(Effector\_init\_pos')\\
\ $|$\ \ $|$\ \ $|$\ \ $|$\ \ \textit{ENV}.simulate(skill$_i$)\\
\ $|$\ \ $|$\ \ $|$\ \ $|$\ \ Loss = \textit{ENV}.get\_loss()\\
\ $|$\ \ $|$\ \ $|$\ \ $|$\ \ \texttt{If} Loss $<$ Min\_loss: \\
\ $|$\ \ $|$\ \ $|$\ \ $|$\ \ $|$\ \ Min\_loss $\leftarrow$ Loss\\
\ $|$\ \ $|$\ \ $|$\ \ $|$\ \ $|$\ \ Best\_action$_n \leftarrow$ (i, j, k)\\
\ $|$\ \ $|$\ \ $|$\ \ $|$\ \ $|$\ \ Final\_state $\leftarrow$ \textit{ENV}.get\_state()\\
\ $|$\ \ $\rhd$\textit{Evaluate empty action}\\
\ $|$\ \ \textit{ENV}.set\_state(New\_state)\\
\ $|$\ \ \textit{ENV}.set\_agent\_pos(Effector\_init\_pos')\\
\ $|$\ \ \textit{ENV}.simulate(empty\_action)\\
\ $|$\ \ Loss = \textit{ENV}.get\_loss()\\
\ $|$\ \ \texttt{If} Loss $<$ Min\_loss: \\
\ $|$\ \ $|$\ \ Min\_loss $\leftarrow$ Loss\\
\ $|$\ \ $|$\ \ Best\_action$_n \leftarrow$ (empty\_action)\\
\ $|$\ \ $|$\ \ Final\_state $\leftarrow$ \textit{ENV}.get\_state()\\
\ $|$\ \ \textit{Plan}.append(Best\_action$_n$)\\
\ $|$\ \ New\_state $\leftarrow$ Final\_state\\
\texttt{Return} \textit{Plan}\\
\bottomrule
\end{tabular}
\end{table}

\end{document}